\documentclass[letterpaper, 10 pt, conference]{ieeeconf}  

\IEEEoverridecommandlockouts                              

\usepackage{epsfig} 
\usepackage{amsmath} 
\usepackage{amsfonts}
\usepackage{listings}
\usepackage{url}
\usepackage{hyperref}
\usepackage{color}
\usepackage{cancel}
\usepackage{amsmath}
\usepackage{enumitem}
\usepackage{caption}
\usepackage{subcaption}
\usepackage{makecell}
\usepackage{bm}

\title{\LARGE \bf
Modeling and Numerical Analysis of Kangaroo Lower Body based on Constrained Dynamics of Hybrid Serial-Parallel Floating-Base Systems
}

\author{Enrico Mingo Hoffman, Andrea Curti, Narcis Miguel, \\ Sai Kishor Kothakota, Adria Roig, and Luca Marchionni
\thanks{All the authors are with PAL Robotics, Barcelona, Spain \newline
        email: {\tt\small \{name.surname\}@pal-robotics.com}
        }
}

\makeatletter
\def\namedlabel#1#2{\begingroup
    #2%
    \def\@currentlabel{#2}%
    \phantomsection\label{#1}\endgroup
}
\makeatother

\begin{document}

\maketitle
\thispagestyle{empty}
\pagestyle{empty}

\begin{abstract}
This paper presents the modeling and numerical analysis of the Kangaroo lower body prototype, a novel bipedal humanoid robot developed and manufactured by PAL Robotics. 
Kangaroo features high-power linear electric actuators combined with unique serial-parallel hybrid chains, which allow for the positioning of all the leg actuators near the base of the robot in order to improve the overall mass distribution.   
To model and analyze such complex nonlinear mechanisms, we employ a constrained formulation that is extended to account for floating-base systems in contact with the environment.
A comparison is made to demonstrate the significant improvements achieved with TALOS, another humanoid bipedal robot designed by PAL Robotics, in terms of equivalent Cartesian inertia at the feet and centroidal angular momentum.
Finally, the paper includes numerical experiments conducted through simulation and preliminary tests performed on the actual Kangaroo platform.
\end{abstract}

\section{Introduction}\label{sec:introduction}
In recent years, significant advancements in mechanics and control influenced the mechanical design of humanoid biped robots, especially the incorporation of lightweight and impact-resilient mechanics, coupled with powerful actuation systems in the lower body. 
As a result, advanced control techniques applied to locomotion have emerged, accompanied by the capability of traversing challenging terrains with the ultimate goal of bridging the gap between laboratory-based research and the practical applicability of humanoid biped robots in less structured real-world scenarios.
\par
In the past, the design of biped humanoid robots has predominantly relied on well-established systems, such as the HRP series,~\cite{kaneko2002design, kaneko2008humanoid}, or ASIMO,~\cite{hirose2007honda}, characterized by stiff position control, serial kinematics, electrical rotary actuators, and high reductions gearboxes, (see also~\cite{tellez2008reem, park2007mechanical}), with very few exceptions, e.g. LOLA~\cite{lohmeier2009humanoid} and TULIP~\cite{hobbelen2008}. 
While these robots demonstrated the ability to perform complex motions, their interaction with the environment and agility were considerably limited.
\begin{figure}[htb!]
    \centering
    \includegraphics[width=1.\columnwidth, trim={2cm 2cm 2cm 2cm}, clip=true]{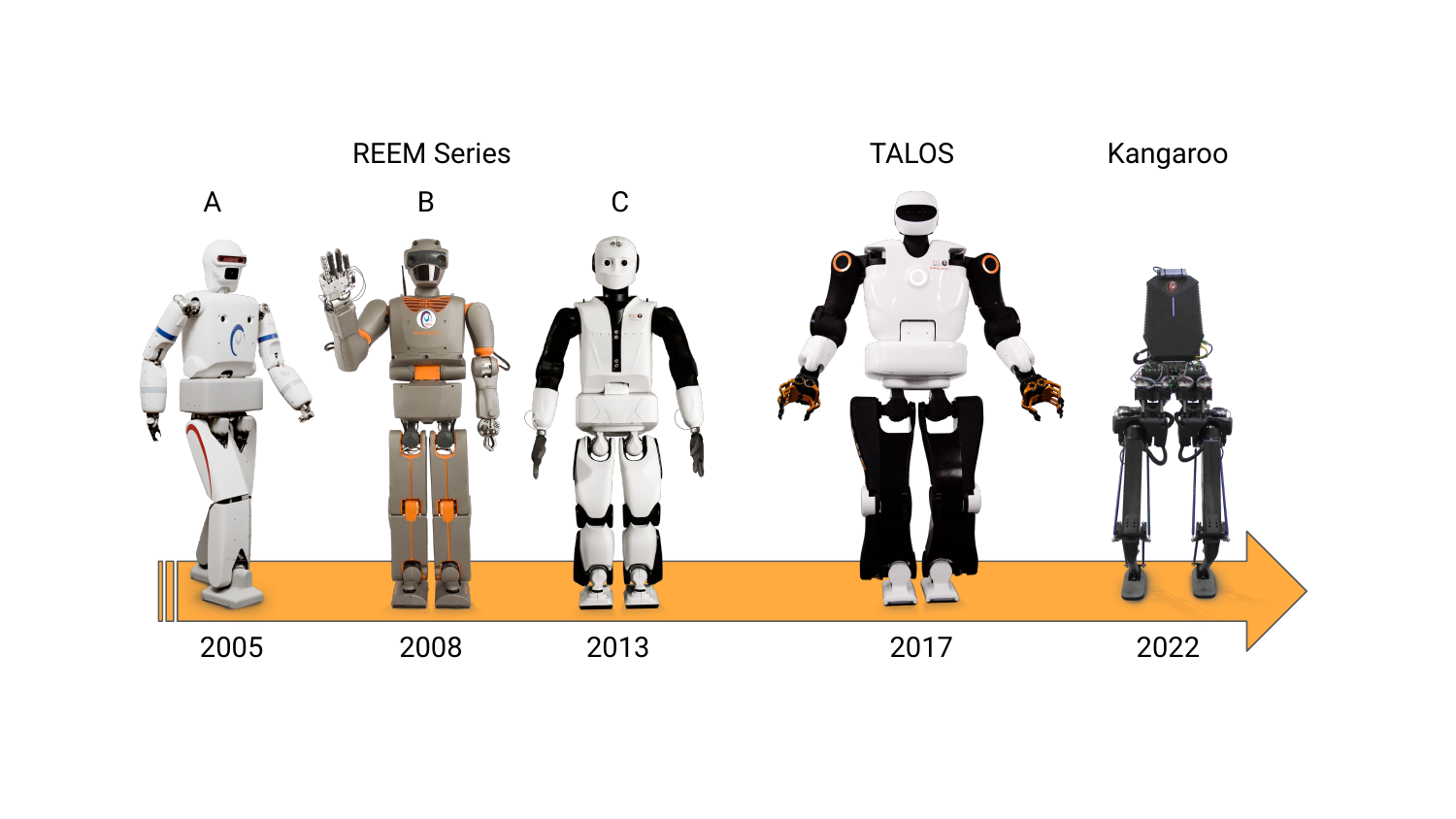}
    \caption{Humanoid bipedal robots designed and produced by PAL Robotics since 2004.}
    \label{fig:pal_humanoids}
\end{figure}
\par
During the DARPA Robotics Challenge (DRC), humanoid torque-controlled robots with high-power capabilities were specifically designed for intervention in disaster scenarios~\cite{spenko2018darpa}, showcasing significant advancements, particularly in terms of interaction capabilities.  
However, these systems exhibited certain limitations, including slow movements and low-impact resilience ~\cite{Tsagarakis:2017, radford2015valkyrie, kojima2015development, stasse2017talos}.
The sluggishness of these robots can be attributed to the heavy weight of the platforms and their high inertia, which stemmed from the utilization of actuators positioned near the moving joints. 
Furthermore, the incorporation of gearboxes with high reduction ratios, such as harmonic drives, contributed to their low-impact resilience. 
Another common characteristic of these systems was the use of torsional elements or strain gauges for joint torque measurement, rather than relying on current measurements. 
This approach was primarily necessitated by the high-friction nature of the high-reduction gearboxes, which made it challenging to estimate and compensate.
\par
After the DRC, novel humanoid platforms have been developed following different design paradigms, in particular: relocating actuators to improve mass distribution, especially on the legs, and employing different actuation units than classical electrical actuators with harmonic drivers, to enhance impact robustness and torque/force control.
Remarkable examples of this new generation of biped robots are the new Atlas from Boston Dynamics, see~\cite{guizzo2019leaps}, Cassie/Digit from Agility Robotics~\cite{hurst2019walk}, and UCLA's Artemis~\cite{zhu2023design}. 
While Atlas relies on hydraulic actuation to achieve robustness and high-power impulsive motions, Cassie/Digit and Artemis use high-torque motors paired with high-efficiency low-reduction-ratio gearboxes located near the main body of the robot, with the motion transmitted through a series of closed linkages to reduce foot inertia.
\par
Following this new trend in humanoid bipedal robotics, we present the modeling and analysis, of the lower body prototype of the Kangaroo robot, a bipedal platform recently developed in PAL Robotics (see Fig.~\ref{fig:pal_humanoids}), 
The mechanical design of the Kangaroo's anthropomorphic legs is based on novel high-power and robust linear actuation units located near the pelvis area with the actuation transferred to joints through a complex system of serial-parallel hybrid chains.
Robotics systems designed with serial-parallel hybrid chains entail multiple advantages such as robustness during impacts, low inertia at the end-effectors, and lightweight leg structure.
\par
The main contribution of this research paper is twofold.
Firstly, we present the analysis of the lower body of Kangaroo including all the Degrees of Freedom (DOFs) (\emph{a.k.a. full-model}), introducing a comprehensive study on floating-base systems presenting serial-parallel hybrid chains in contact with the environment, based on constrained multi-body modeling.
Secondly, we carry on a comparison with the TALOS robot, focusing on dynamic quantities that are relevant in humanoid bipedal platforms, i.e. equivalent Cartesian inertia at the feet and Centroidal Angular Momentum.
\par
The paper is organized as follows: Section~\ref{sec:soa} presents related works on humanoid bipedal platforms using closed linkages, their analysis, and control methods.
Section~\ref{sec:modeling} introduces the background on kinematic and dynamic modeling of floating-base robots presenting serial-parallel hybrid linkages.
Section~\ref{sec:Kangaroo} presents the Kangaroo robot and its lower-body kinematic structure, together with an analysis of the closed sub-mechanisms present in the legs, including joint limits and non-linear transmission effects.
In Section~\ref{sec:comparison}, we carry out the comparison against the TALOS biped robot.
The full-model validation and preliminary experiments with a prototype of the Kangaroo robot are reported in Section~\ref{sec:experiments}.
Finally, Section~\ref{sec:conclusion} closes the paper with final remarks and future development.

\section{Related Works}\label{sec:soa}
The main reason to adopt serial mechanisms relies on their straightforward structure, augmented workspace, and simple manufacturing and maintenance processes, therefore representing most of the time the state-of-the-art in robotics systems.
For these reasons, most bipedal robots' legs consist of 6 DOFs serial mechanisms with actuators distributed along the whole kinematic chain, from the hip to the ankle, see for example~\cite{kaneko2002design, hirose2007honda, tellez2008reem, park2007mechanical}.
However, despite the aforementioned advantages, these types of robot architectures generally present only limited precision, low structural stiffness, and poor dynamic characteristics, related to mass and inertia distribution. 
In particular, the latter plays a fundamental role in agile locomotion and impact handling and mitigation.
On the other hand, a parallel robot can provide higher stiffness, speed, accuracy, and payload capacity, at the cost of a reduced workspace and complex geometry, requiring careful analysis and control.
\cite{kumar2020survey} defines a series-parallel hybrid robot as a robot constituted by a serial or tree-type combination of serial and parallel mechanisms, combining the advantages of both worlds, but also inheriting their kinematic complexities.
\par
The last few decades of research in humanoid bipedal robotics have shown that achieving high dynamic performance requires a stiff structure and good mass distribution~\cite{stasse2019overview}. 
These characteristics can be achieved using series-parallel hybrid structures.
In fact, more recently, several research works started to introduce mechanical leg designs that are based on series-parallel hybrid kinematics.
In some cases, sub-mechanisms are based on parallel kinematics to achieve the lightweight, modular, and compact design of sub-parts.
A typical case is the ankle, where motors are relocated closer to the knee by employing a four-bar linkage mechanism with a reduction of inertia at the leg end-effector, as in~\cite{Tsagarakis:2017, radford2015valkyrie, Ruscelli18}.
A first example of a humanoid bipedal robot strongly based on series-parallel hybrid kinematics chains is LOLA, from~\cite{lohmeier2009humanoid}, which legs were designed mixing rotational, linear actuators, and parallel/differential transmissions.
Other examples of humanoid bipedal systems where series-parallel hybrid chains were extensively used for the upper and lower body are HYDRA,~\cite{Kaminaga16}, SAFFIR,~\cite{lahr2016biologically}, CARL,~\cite{schutz2017carl}, and Disney's HYBRID LEG,~\cite{gim2018design}, to name a few.
\par
From the software and control point of view, the parallel mechanism is often handled by relying on an abstraction level that separates the parallel-closed chain computation into specialized functions (e.g. \emph{transmissions} in \texttt{ros\_control}~\cite{chitta2017ros_control}), or approximates the system as a serial chain.
\par
In~\cite{mronga2021whole}, is presented a Quadratic Programming (QP)-based framework for whole-body control taking into account closed kinematic loops. 
The authors show how approaches that are based on serial-chain abstraction and separated into specialized functions lead to theoretical and practical disadvantages, in particular:
\begin{itemize}
    \item the real physical limits induced by the mechanism can not be modeled using a simple box constraint on the serial model,
    \item the whole-body control solution may be less accurate as it does not consider the correct dynamics of the system,
\end{itemize}
The proposed framework is applied to a full-size humanoid robot with multiple series-parallel hybrid chains, named RH5, see~\cite{Kumar20, ebetaer2021design}.  

\par
Recently, agile and dynamic walking has been achieved with platforms based on parallel mechanism legs, e.g. ATRIAS,~\cite{grimes2012design}, and series-parallel hybrid mechanism, e.g. Cassie and Digit~\cite{reher2019dynamic}.
ATRIAS leg design is based on a four-bar linkage driven by 2 motors for leg extension/retraction in common mode, and swing in differential mode.  
The design of ATRIAS has evolved in Cassie/Digit, employing a series-parallel hybrid mechanism modeled by cutting the loops and adding consistency constraints, see~\cite{apgar2018fast}.
\par
In~\cite{gim2018design} is proposed an IK approach to retarget motions computed from a serial model onto a hybrid one to control the Disney's HYBRID LEG, see also~\cite{schumacher2021versatile}.
However, the proposed approach is only kinematics-based.
\par
The work in~\cite{kumar2020survey} reports (in Table 1) a comparison among different series-parallel hybrid humanoid robots, considering the number of parallel sub-mechanism modules and the provided free DOFs w.r.t. the total number of free DOFs.
The same table is reported in Table~\ref{tab:comparison}, considering only humanoid robots and augmented with the data for the  Kangaroo robot.
It is worth noticing that among all the considered humanoid bipedal systems, Kangaroo is the only one relocating the ankle's actuators at the back of the top rear femur, hence not presenting any actuator nor electronics under the knee. 
This unique feature makes Kangaroo a platform particularly resilient against impacts and potentially ideal to perform agile and dynamic locomotion.

\begin{table*}[h]
\caption{Overview of parallel sub-mechanism based modules with different complexities in serial-parallel hybrid robots} 
\centering 
\begin{tabular}{l c c c c c c c c}
Robot                       & \multicolumn{5}{c}{Number of Parallel mechanisms (Free DOFs)} & & \multicolumn{2}{c}{Composition Free DOFs} \\ 
                            \cline{2-6}  
                            \cline{8-9}
Name(year)                  & 1-DOF & 2-DOF & 3-DOF & 4-DOF & 6-DOF & & Total & Parallel \\
\hline 
LOLA,~\cite{lohmeier2009humanoid}                  & 2 & 2 & – & – & – & & 25 & 6 \\
AILA,~\cite{lemburg2011aila}                  & 2 & 2 & – & – & – & & 20 & 6 \\
Valkyrie,~\cite{radford2015valkyrie}              & – & 5 & – & – & – & & 35 & 10 \\
TORO,~\cite{englsberger2014overview}                  & 2 & – & – & – & – & & 27 & 2 \\
THOR,~\cite{lee2014thesis}                  & 4 & 4 & – & – & – & & 30 & 12 \\
SAFFIR,~\cite{lahr2014thesis}                & 4 & 4 & – & – & – & & 30 & 12 \\
LARMBOT,~\cite{cafolla2016larmbot}               & – & – & 2 & 1 & – & & 22 & 10 \\
TALOS,~\cite{stasse2017talos}                 & 2 & – & – & – & – & & 27 & 2 \\
RH5,~\cite{ebetaer2021design}                   & 5 & 5 & – & – & – & & 32 & 15 \\
Disney Biped,~\cite{gim2018design}          & – & – & – & – & 2 & & 12 & 12 \\
\textbf{Kangaroo (2022)}    & \textbf{4} & \textbf{4} & \textbf{-} & \textbf{-} & \textbf{-} & & \textbf{12} & \textbf{12} 
\end{tabular}
\label{tab:comparison}
\end{table*}


\section{Background on Modeling Serial-Parallel Hybrid Chains in Floating-Base Systems}\label{sec:modeling}
Classical approaches in modeling parallel linkages are based on geometrical analysis, leveraging case-specific analytical expressions for the linkage kinematics, as in~\cite{kumar2019kinematic} where the closed-form solution of a 2SPRR-U humanoid ankle is presented.
This solution permits embedding the closed loop constraints directly in the equation of motions eventually saving computational cost and improving numerical accuracy.
However, such closed-form solutions are not always available and do not generalize to arbitrary serial-parallel hybrid mechanisms.
\par
According to \cite{laulusa2008review}, a more comprehensive and promising method entails modeling these linkages as constrained multi-body systems. 
This approach allows for the consideration of intricate geometries and dynamic effects associated with multiple serial-parallel hybrid chains.
While the direct map between the linkage parameters and the resulting motion is no longer expressed analytically, it can still be determined through numerical computation.
In this section, we recall this approach, including floating-base systems incorporating generic series-parallel hybrid chains, with the aim of using it for the modeling and analysis of the Kangaroo robot.
\par
From now on, as we did in the previous sections, we will call \emph{actuated} DOFs the ones that can provide force/torque, and \emph{passive} DOFs the ones that can not.
We will consider mechanisms with the following assumptions:
\begin{description}
    \item[\namedlabel{itm:assumption1}{Assumption 1}] The motion of the passive DOFs is totally constrained by the actuated DOFs;
    \item[\namedlabel{itm:assumption2}{Assumption 2}] The number of \emph{free} DOFs coincides with the number of actuated DOFs. 
\end{description}

\subsection{Modeling of Serial-Parallel Hybrid Chains}\label{subsec:sphc}
\begin{figure}[htb!]
    \centering
    \includegraphics[width=1.\columnwidth, trim={7.5cm 6cm 9cm 3.5cm}, clip=true]{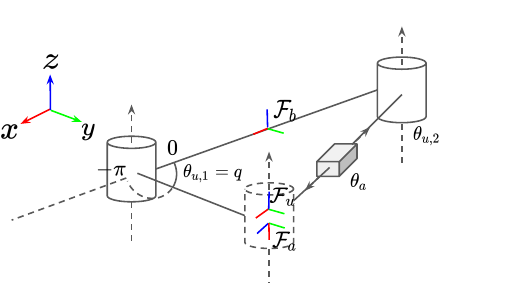}
    \caption{1DOF RR\underline{P}R closed mechanism.}
    \label{fig:crank_slider}
\end{figure}

Let's consider a simple 1DOF RR\underline{P}R closed mechanism as the one depicted in Fig.~\ref{fig:crank_slider}. 
To model this closed linkage as a constrained multi-body system, the mechanism is first \emph{opened} at one of the passive DOFs, represented by a dashed line in Fig.~\ref{fig:crank_slider}, forming two open kinematic chains.
In this case, we have:
\begin{equation}
    \boldsymbol{\theta} = 
    \begin{bmatrix} 
    \boldsymbol{\theta}_u
    \\ 
    \boldsymbol{\theta}_a
    \end{bmatrix} \in \mathbb{R}^n,
    \label{eq:generalized_coordinates_hsp}
\end{equation}
with $\boldsymbol{\theta}_u \in \mathbb{R}^m$ the passive DOFs and $\boldsymbol{\theta}_a \in \mathbb{R}^{n-m}$ the actuated DOF, $m < n$.
For the 1DOF RR\underline{P}R mechanism considered, we have in particular $n = 3$ and with $m = 2$.
\par
We denote with $\mathcal{F}_a$ and $\mathcal{F}_u$ the frames associated with each open chain, placed at the removed DOF, as shown in Fig.~\ref{fig:crank_slider}.
Considering~\ref{itm:assumption1} and~\ref{itm:assumption2}, we can enforce a constraint in the form:
\begin{equation}
    {^a\mathbf{f}_u}\left(\boldsymbol{\theta}\right) = \mathbf{0} \in \mathbb{R}^m.
    \label{eq:crank_slider_constraint}
\end{equation}
along the constrained directions to keep the two frames overlapped to close the linkage. 
With $\boldsymbol{\theta}_a$ as known quantities, the equation~\eqref{eq:crank_slider_constraint} consists in a non-linear system of $m$ equations in $m$ unknowns $\boldsymbol{\theta}_u$.  
For example, for the considered closed linkage, the motion is constrained on the local plane $yz$ of frame $\mathcal{F}_a$, as shown in Fig.~\ref{fig:crank_slider}, resulting in the relative position of frame $\mathcal{F}_u$ w.r.t. $\mathcal{F}_a$, along the local $yz$ directions, being $\mathbf{0}$.
We can compute the constraint Jacobian from~\eqref{eq:crank_slider_constraint}:
\begin{equation}
    {\mathbf{J}_l}\left(\boldsymbol{\theta}\right)\boldsymbol{\dot{\theta}} = \mathbf{0},
    \label{eq:crank_slider_constraint_differential}
\end{equation}
with ${\mathbf{J}_l}\left(\boldsymbol{\theta}\right) \in \mathbb{R}^{m \times n}$ being the \emph{relative Jacobian} between the frames $\mathcal{F}_u$ and $\mathcal{F}_a$ expressed in $\mathcal{F}_a$, such that:
\begin{equation}
    {\mathbf{J}_l}\left(\boldsymbol{\theta}\right)\boldsymbol{\dot{\theta}} = {^a\mathbf{J}_{a,u}}\left(\boldsymbol{\theta}\right)\boldsymbol{\dot{\theta}} = {^a\mathbf{v}_{a,u}},
\end{equation}
where ${^a\mathbf{v}_{a,u}}$ is the velocity along the constrained directions of the frame $\mathcal{F}_u$ w.r.t. the frame $\mathcal{F}_a$ expressed in frame $\mathcal{F}_a$ (please refer to Appendix~\ref{app:A} on how to compute relative kinematics quantities).
It is possible to divide the constraint Jacobian into its actuated and passive parts:
\begin{equation}
    \mathbf{J}_l\left(\boldsymbol{\theta}\right)\boldsymbol{\dot{\theta}} = 
    \begin{bmatrix}
\mathbf{J}_{l,u}\left(\boldsymbol{\theta}\right) &
\mathbf{J}_{l,a}\left(\boldsymbol{\theta}\right)
\end{bmatrix}
\begin{bmatrix} 
\boldsymbol{\dot{\theta}}_u
\\
\boldsymbol{\dot{\theta}}_a
\end{bmatrix} = \mathbf{0},
\end{equation}
with $\mathbf{J}_{l,u}\left(\boldsymbol{\theta}\right) \in \mathbb{R}^{m \times m}$ the passive part, and $\mathbf{J}_{l,a}\left(\boldsymbol{\theta}\right) \in \mathbb{R}^{m \times n-m}$ the actuated part of the constraint Jacobian $\mathbf{J}_l\left(\boldsymbol{\theta}\right)$.
The constraint~\eqref{eq:crank_slider_constraint_differential} can be written as:
\begin{equation}
    \mathbf{J}_{l,u}\left(\boldsymbol{\theta}\right)\boldsymbol{\dot{\theta}}_u + \mathbf{J}_{l,a}\left(\boldsymbol{\theta}\right)\boldsymbol{\dot{\theta}}_a = \mathbf{0},
    \label{eq:separate_jacobians_crank_slider}
\end{equation}
that permits to compute the passive velocities from the actuated ones, a.k.a. Differential Forward Kinematics (DFK):
\begin{subequations}
\label{eq:closed_linkage_fk}
\begin{align}
    \boldsymbol{\dot{\theta}}_u & = -\mathbf{J}_{l,u}\left(\boldsymbol{\theta}\right)^{-1}\mathbf{J}_{l,a}\left(\boldsymbol{\theta}\right)\boldsymbol{\dot{\theta}}_a 
     = \mathbf{J}_m\left(\boldsymbol{\theta}\right)\boldsymbol{\dot{\theta}}_a \tag{\ref{eq:closed_linkage_fk}},
\end{align}
\end{subequations}
with $\mathbf{J}_m\left(\boldsymbol{\theta}\right) \in \mathbb{R}^{m \times n-m}$ named the \emph{mapping Jacobian}, and considering $\mathbf{J}_{l,u}\left(\boldsymbol{\theta}\right)$ invertible, thus not in singularity.
\par
The passive accelerations $\boldsymbol{\ddot{\theta}}_u$ can be computed from the (desired/measured) actuated accelerations $\boldsymbol{\ddot{\theta}}_a$, positions $\boldsymbol{\theta}$ and velocities $\boldsymbol{\dot{\theta}}$.
In fact, the constraint~\eqref{eq:crank_slider_constraint} can be easily expressed at the acceleration level:
\begin{equation}
    \mathbf{J}_l\left(\boldsymbol{\theta}\right)\boldsymbol{\ddot{\theta}} + \mathbf{\dot{J}}_l(\boldsymbol{\theta}, \boldsymbol{\dot{\theta}})\boldsymbol{\dot{\theta}} = \mathbf{0},
    \label{eq:cranck_slider_acc}
\end{equation}
please refer to Appendix~\ref{app:A2} for the computation of the relative bias term $\mathbf{\dot{J}}_l(\boldsymbol{\theta}, \boldsymbol{\dot{\theta}})\boldsymbol{\dot{\theta}}$.
Splitting \eqref{eq:cranck_slider_acc} into its actuated and passive parts leads to:
\begin{equation}
    \begin{bmatrix}
\mathbf{J}_{l,u}\left(\boldsymbol{\theta}\right) &
\mathbf{J}_{l,a}\left(\boldsymbol{\theta}\right)
\end{bmatrix}
\begin{bmatrix} 
\boldsymbol{\ddot{\theta}}_u
\\
\boldsymbol{\ddot{\theta}}_a
\end{bmatrix}
+
\mathbf{\dot{J}}_l(\boldsymbol{\theta}, \boldsymbol{\dot{\theta}})\boldsymbol{\dot{\theta}} = \mathbf{0},
\end{equation}
therefore:
\begin{subequations}
\label{eq:passive_accelerations_crank_slider}
\begin{align}
 \boldsymbol{\ddot{\theta}}_u & = -\mathbf{J}_{l,u}\left(\boldsymbol{\theta}\right)^{-1}\left(\mathbf{J}_{l,a}\left(\boldsymbol{\theta}\right)\boldsymbol{\ddot{\theta}}_a +  \mathbf{\dot{J}}_l(\boldsymbol{\theta}, \boldsymbol{\dot{\theta}})\boldsymbol{\dot{\theta}}\right) = \nonumber \\
 & = \mathbf{J}_m\left(\boldsymbol{\theta}\right)\boldsymbol{\ddot{\theta}}_a -\mathbf{J}_{l,u}\left(\boldsymbol{\theta}\right)^{-1}\mathbf{\dot{J}}_l(\boldsymbol{\theta}, \boldsymbol{\dot{\theta}})\boldsymbol{\dot{\theta}}. \tag{\ref{eq:passive_accelerations_crank_slider}}
\end{align}
\end{subequations}

\par
The equations of motions consist of a constrained dynamic, see~\cite{carpentier_hal-03271811}, and it can be written as:
\begin{equation}
    \mathbf{M}(\boldsymbol{\theta})\boldsymbol{\ddot{\theta}} + \mathbf{h}(\boldsymbol{\theta}, \boldsymbol{\dot{\theta}}) = \mathbf{S}\boldsymbol{\tau} + \mathbf{J}_l\left(\boldsymbol{\theta}\right)^T\boldsymbol{\lambda},
    \label{eq:crank_slider_dynamics}
\end{equation}
with $\mathbf{M}(\boldsymbol{\theta}) \in \mathbb{R}^{n \times n}$ the inertia matrix, $\mathbf{h}(\boldsymbol{\theta}, \boldsymbol{\dot{\theta}}) \in \mathbb{R}^n$ non-linear terms consisting in Coriolis/Centrifugal and gravitational forces, $\boldsymbol{\tau} \in \mathbb{R}^{n-m}$ the actuated torques/forces, $\boldsymbol{\lambda} \in \mathbb{R}^{m}$ the constrained forces, and $\mathbf{S} = \left[\mathbf{0}_{m \times n-m} \ \mathbf{I}_{n-m \times n-m}\right]^T$ the passive selection matrix such that $\mathbf{S}\boldsymbol{\tau} \in \mathbb{R}^n$.
The dynamics in~\eqref{eq:crank_slider_dynamics} can be as well divided into its actuated and passive parts:
\begin{subequations}
\begin{align}
    \mathbf{M}_u\left(\boldsymbol{\theta}\right)\boldsymbol{\ddot{\theta}} + \mathbf{h}_u(\boldsymbol{\theta}, \boldsymbol{\dot{\theta}}) &= \mathbf{J}_{l,u}\left(\boldsymbol{\theta}\right)^T\boldsymbol{\lambda}, \label{eq:crank_slider_dynamics_passive}\\
    \mathbf{M}_a\left(\boldsymbol{\theta}\right)\boldsymbol{\ddot{\theta}} + \mathbf{h}_a(\boldsymbol{\theta}, \boldsymbol{\dot{\theta}}) &= \boldsymbol{\tau} + \mathbf{J}_{l,a}\left(\boldsymbol{\theta}\right)^T\boldsymbol{\lambda}, \label{eq:crank_slider_dynamics_actuated}
\end{align}
\end{subequations}
with $\mathbf{M}_u\left(\boldsymbol{\theta}\right) \in \mathbb{R}^{m \times n}$, $\mathbf{h}_u(\boldsymbol{\theta}, \boldsymbol{\dot{\theta}}) \in \mathbb{R}^m$, $\mathbf{M}_a\left(\boldsymbol{\theta}\right) \in \mathbb{R}^{n-m \times n}$ and $\mathbf{h}_a(\boldsymbol{\theta}, \boldsymbol{\dot{\theta}}) \in \mathbb{R}^{n-m}$.
The constrained forces $\boldsymbol{\lambda}$ can be removed by using equation~\eqref{eq:crank_slider_dynamics_passive}:
\begin{equation}
    \boldsymbol{\lambda} = \mathbf{J}_{l,u}(\boldsymbol{\theta})^{-T} 
    \left( \mathbf{M}_u( \boldsymbol{\theta})\boldsymbol{\ddot{\theta}} + \mathbf{h}_u(\boldsymbol{\theta}, \boldsymbol{\dot{\theta}}) \right).
    \label{eq:crank_slider_constrained_forces}
\end{equation}
Let's now define the quantity:
\begin{equation}
    \boldsymbol{\tau}_a(\boldsymbol{\theta}, \boldsymbol{\dot{\theta}}, \boldsymbol{\ddot{\theta}}, \boldsymbol{\tau}) = \boldsymbol{\tau} - \left( \mathbf{M}_a\left(\boldsymbol{\theta}\right)\boldsymbol{\ddot{\theta}} + \mathbf{h}_a(\boldsymbol{\theta}, \boldsymbol{\dot{\theta}}) \right),
\end{equation}
the actuated torques, and
\begin{equation}
    \boldsymbol{\tau}_u(\boldsymbol{\theta}, \boldsymbol{\dot{\theta}}, \boldsymbol{\ddot{\theta}}) = \mathbf{M}_u\left(\boldsymbol{\theta}\right)\boldsymbol{\ddot{\theta}} + \mathbf{h}_u(\boldsymbol{\theta}, \boldsymbol{\dot{\theta}})
\end{equation}
the \emph{passive} torques.
Equation~\eqref{eq:crank_slider_dynamics_actuated} can be rewritten using $\boldsymbol{\tau}_a$\footnote{Here we drop dependencies of $\boldsymbol{\tau}_a$ and $\boldsymbol{\tau}_u$}:
\begin{equation}
    \boldsymbol{\tau}_a = -\mathbf{J}_{l,a}\left(\boldsymbol{\theta}\right)^T\boldsymbol{\lambda}
\end{equation}
and substituting $\boldsymbol{\lambda}$:
\begin{subequations}
\label{eq:closed_linkage_forward_torque_mapping}
\begin{align}
    \boldsymbol{\tau}_a &= -\mathbf{J}_{l,a}\left(\boldsymbol{\theta}\right)^T\mathbf{J}_{l,u}\left(\boldsymbol{\theta}\right)^{-T}\left( \mathbf{M}_u\left(\boldsymbol{\theta}\right)\boldsymbol{\ddot{\theta}} + \mathbf{h}_u(\boldsymbol{\theta}, \boldsymbol{\dot{\theta}}) \right) = \nonumber \\
    & = -\mathbf{J}_{l,a}\left(\boldsymbol{\theta}\right)^T\mathbf{J}_{l,u}\left(\boldsymbol{\theta}\right)^{-T}\boldsymbol{\tau}_u \tag{\ref{eq:closed_linkage_forward_torque_mapping}} 
     = \mathbf{J}_m\left(\boldsymbol{\theta}\right)^T\boldsymbol{\tau}_u \nonumber
\end{align}
\end{subequations}
with: 
\begin{subequations}
\label{eq:mapping_jacobian_transpose}
\begin{align}    
    \mathbf{J}_m\left(\boldsymbol{\theta}\right)^T &= -\left( {\mathbf{J}_{l,u}\left(\boldsymbol{\theta}\right)^{-1}}\mathbf{J}_{l,a}(\boldsymbol{\theta}) \right )^T = \nonumber \\
     &= - {\mathbf{J}_{l,a}(\boldsymbol{\theta})^T}\mathbf{J}_{l,u}(\boldsymbol{\theta})^{-T}, \tag{\ref{eq:mapping_jacobian_transpose}}
\end{align}
\end{subequations} 
where equation~\eqref{eq:closed_linkage_forward_torque_mapping} permits to map generic torques on the passive DOFs into the actuated ones.
The inverse dynamics can finally be computed from~\eqref{eq:crank_slider_dynamics_actuated}, plugging~\eqref{eq:crank_slider_constrained_forces}:
\begin{subequations}
\label{eq:closed_linkage_inverse_dynamics}
\begin{align}
    \boldsymbol{\tau} &= \mathbf{M}_a(\boldsymbol{\theta})\boldsymbol{\ddot{\theta}} + \mathbf{h}_a(\boldsymbol{\theta}, \boldsymbol{\dot{\theta}}) -\mathbf{J}_{l,a}\left(\boldsymbol{\theta}\right)^T\boldsymbol{\lambda}= \nonumber\\
    &=\mathbf{M}_a(\boldsymbol{\theta})\boldsymbol{\ddot{\theta}} + \mathbf{h}_a(\boldsymbol{\theta}, \boldsymbol{\dot{\theta}}) - \nonumber\\ & \quad \quad -\mathbf{J}_{l,a}\left(\boldsymbol{\theta}\right)^T\mathbf{J}_{l,u}\left(\boldsymbol{\theta}\right)^{-T}\left( \mathbf{M}_u(\boldsymbol{\theta})\boldsymbol{\ddot{\theta}} + \mathbf{h}_u(\boldsymbol{\theta}, \boldsymbol{\dot{\theta}}) \right)= \tag{\ref{eq:closed_linkage_inverse_dynamics}}\\
    &=\mathbf{M}_a(\boldsymbol{\theta})\boldsymbol{\ddot{\theta}} + \mathbf{h}_a(\boldsymbol{\theta}, \boldsymbol{\dot{\theta}}) +  \mathbf{J}_m(\boldsymbol{\theta})^T\left(\mathbf{M}_u(\boldsymbol{\theta})\boldsymbol{\ddot{\theta}} + \mathbf{h}_u(\boldsymbol{\theta}, \boldsymbol{\dot{\theta}}) \right),\nonumber
\end{align}
\end{subequations}
where we have used the~\eqref{eq:mapping_jacobian_transpose}.
\par
\begin{figure}
     \centering
     \begin{subfigure}[b]{1.\columnwidth}
         \centering
         \includegraphics[width=1.\columnwidth, trim={7.5cm 4cm 7.5cm 4cm}, clip=true]{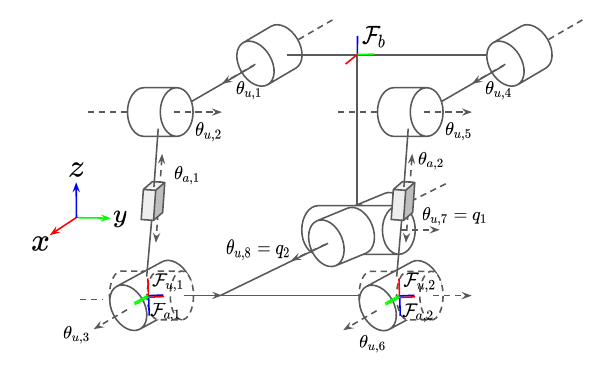}
         \caption{Full mechanism.}
         \label{fig:spatial_differential_linkage}
     \end{subfigure}
     \\
     \vspace{0.5cm}
     \begin{subfigure}[b]{1.\columnwidth}
         \centering
         \includegraphics[width=1.\columnwidth, trim={7.5cm 4cm 7.5cm 4cm}, clip=true]{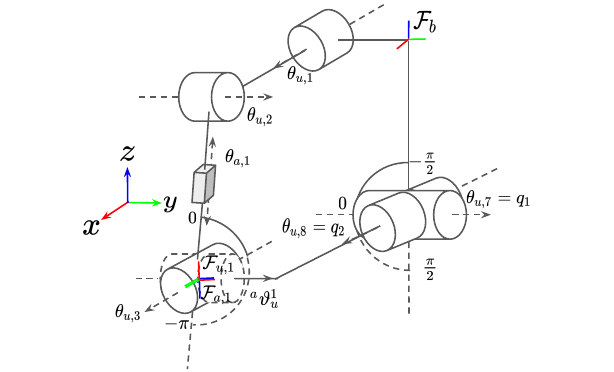}
         \caption{Right brace of the mechanism.}
         \label{fig:half_spatial_differential_linkage}
     \end{subfigure}
     \caption{2DOFs U-2RR\underline{P}U differential mechanism.}
     \label{fig:2DOFsU-2RRPU}
\end{figure}

The constrained approach can be extended to complex mechanisms formed by multiple closed sub-mechanisms by stacking constraint Jacobians associated with each closed sub-mechanism, such that the total number of constraints will be equal to the number of passive DOFs. 
For example, the 2DOFs U-2RR\underline{P}U differential mechanism depicted in Fig.~\ref{fig:spatial_differential_linkage} is modeled introducing a total of $m = 8$ constraints, split in $2$ closed sub-mechanims, each one introducing $4$ constraints:
\begin{equation}
\label{eq:half_spatial_differential_linkage_constraint}
\left\{\begin{matrix}
{^a\mathbf{p}_u^1}\left(\boldsymbol{\theta}\right) = \mathbf{0} & \in \mathbb{R}^3 & \text{for the relative position,}\\
\\ 
{^a\vartheta_u^1}\left(\boldsymbol{\theta}\right) = c & \in \mathbb{R} & \text{for the relative orientation,}
\end{matrix}\right.
\end{equation}
between the frames $\mathcal{F}_{a,1}$ and $\mathcal{F}_{u,1}$ depicted in Fig.~\ref{fig:half_spatial_differential_linkage}, with $\boldsymbol{\theta} \in \mathbb{R}^{10}$, modeling relative rotations using Euler angles $(\phi \ \vartheta \ \varphi)$ for simplicity\footnote{In our implementation and in the experimental results we use a quaternion-based parameterization}.
\par
Finally, given a joint-space selection matrix $\mathbf{P} \in \mathbb{R}^{n-m \times m}$ which selects a subset of the passive DOFs such that:
\begin{equation}
    \mathbf{\dot{q}} = \mathbf{P}\boldsymbol{\dot{\theta}}_u, \quad \quad
    \boldsymbol{\tau}_j = \mathbf{P}\boldsymbol{\tau}_u,
\end{equation}
it is possible to compute the DIK problem w.r.t.~\eqref{eq:closed_linkage_fk}:
\begin{equation}
    \boldsymbol{\dot{\theta}}_a = \left(\mathbf{P}\mathbf{J}_m(\boldsymbol{\theta})\right)^{-1}\mathbf{\dot{q}}_d
    \label{eq:DIK}
\end{equation}
and the inverse torque mapping problem w.r.t.~\eqref{eq:closed_linkage_forward_torque_mapping}:
\begin{equation}
    \boldsymbol{\tau}_j = \left(\mathbf{J}_m(\boldsymbol{\theta})^T\mathbf{P}^T\right)^{-1}\boldsymbol{\tau}.
    \label{eq:ITM}
\end{equation}
Equation~\eqref{eq:DIK} can be used to impose a desired behavior to the passive DOFs, while equation~\eqref{eq:ITM} can be used to compute the equivalent \emph{virtual} torque acting on these passive DOFs.

\subsection{Floating-Base Dynamics including Serial-Parallel Hybrid Chains}
To model a floating-base system presenting hybrid serial-parallel kinematic chains, we augment the generalized coordinates in~\eqref{eq:generalized_coordinates_hsp} with a parameterization in $SE(3)$ of the pose of the floating-base, and we denote with $\boldsymbol{\nu} \in \mathbb{R}^{n+6}$ the generalized velocities:
\begin{equation}
    \mathbf{q} = \begin{bmatrix}
\mathbf{p}\\
\boldsymbol{\rho}\\ 
\boldsymbol{\theta}
\end{bmatrix}, \quad \quad 
\boldsymbol{\nu} = \begin{bmatrix}
\mathbf{\dot{p}}\\
\boldsymbol{\omega}\\ 
\boldsymbol{\dot{\theta}}
\end{bmatrix},
\label{eq:generalized_coordinates}
\end{equation}
with $\mathbf{p} \in \mathbb{R}^3$ and $\boldsymbol{\rho} \in \textbf{S}^3 = \left\{\boldsymbol{\rho} \in \mathbb{R}^4 : \| \boldsymbol{\rho} \| = 1 \right\}$, using quaternions for the orientation.
%
\par
The robot can only exert forces from the contact points $c$, we denote with $\mathbf{F}_i \in \mathbb{R}^3$ the contact force vectors, with $i = 1, ..., c$, $c = 8$ in the case of double support when both feet are on the ground.
The floating-base inverse dynamics model of this system consists of a set of equations:
\begin{subequations}
\begin{align}
&\mathbf{M}(\mathbf{q})\boldsymbol{\dot{\nu}} + \mathbf{h}(\mathbf{q}, \boldsymbol{\nu}) = \mathbf{S}\boldsymbol{\tau} + \mathbf{J}_c(\mathbf{q})^T\mathbf{F} + \mathbf{J}_l(\mathbf{q})^T\boldsymbol{\lambda},
\label{eq:floating_base_inverse_dynamics_full}
\\
&\mathbf{J}_c(\mathbf{q}) \boldsymbol{\dot{\nu}} + \mathbf{\dot{J}}_c(\mathbf{q}, \boldsymbol{\nu}) \boldsymbol{\nu} = \mathbf{0}, 
\label{eq:kinematic_contact_constraint}
\\ 
&\mathbf{J}_l(\mathbf{q}) \boldsymbol{\dot{\nu}} + \mathbf{\dot{J}}_l(\mathbf{q}, \boldsymbol{\nu}) \boldsymbol{\nu} = \mathbf{0}, \label{eq:kinematic_closed_chain_constraint}
\end{align}
\end{subequations}
with $\mathbf{M}(\mathbf{q}) \in \mathbb{R}^{n+6 \times n+6}$ the floating-base inertia matrix, $\mathbf{h}(\mathbf{q}, \boldsymbol{\nu}) \in \mathbb{R}^{n+6}$ the floating-base non-linear terms, $\mathbf{J}_c(\mathbf{q}) \in \mathbb{R}^{3c \times n+6}$ contacts Jacobians, $\mathbf{F} = \left[\mathbf{F}_1^T, \dots, \mathbf{F}_c^T\right]^T$, $\mathbf{J}_l(\mathbf{q}) \in \mathbb{R}^{m \times n+6}$ closed linkages constraint Jacobians and $\boldsymbol{\lambda} \in \mathbb{R}^m$ the associated closed linkages constraint forces, $\mathbf{\dot{J}}_c(\mathbf{q}, \boldsymbol{\nu}) \boldsymbol{\nu} \in \mathbb{R}^{3c}$ and $\mathbf{\dot{J}}_l(\mathbf{q}, \boldsymbol{\nu}) \boldsymbol{\nu} \in \mathbb{R}^m$ the acceleration bias terms for contacts and closed linkages kinematics constraint, respectively, and finally $\mathbf{S} \in \mathbb{R}^{n+6 \times a}$ the selection matrix for the actuated torques.
\par
We can divide the dynamics in~\eqref{eq:floating_base_inverse_dynamics_full} into its actuated, passive, and underactuated, i.e. the floating-base, components:
\begin{subequations}
\begin{align}
&\mathbf{M}_b(\mathbf{q})\boldsymbol{\dot{\nu}} + \mathbf{h}_b(\mathbf{q}, \boldsymbol{\nu}) = \mathbf{J}_{c,b}(\mathbf{q})^T\mathbf{F}, 
\label{eq:inverse_dynamics_floating_base}
\\
&\mathbf{M}_u(\mathbf{q})\boldsymbol{\dot{\nu}} + \mathbf{h}_u(\mathbf{q}, \boldsymbol{\nu}) = \mathbf{J}_{c,u}(\mathbf{q})^T\mathbf{F} + \mathbf{J}_{l,u}(\mathbf{q})^T\boldsymbol{\lambda},
\label{eq:inverse_dynamics_passive_part}
\\
&\mathbf{M}_a(\mathbf{q})\boldsymbol{\dot{\nu}} + \mathbf{h}_a(\mathbf{q}, \boldsymbol{\nu}) = \boldsymbol{\tau} + \mathbf{J}_{c,a}(\mathbf{q})^T\mathbf{F} + \mathbf{J}_{l,a}(\mathbf{q})^T\boldsymbol{\lambda},
\label{eq:inverse_dynamics_actuated_part}
\end{align}
\end{subequations}
with $\mathbf{M}_b(\mathbf{q}) \in \mathbb{R}^{6 \times n + 6}$ and $\mathbf{h}_b(\mathbf{q}, \boldsymbol{\nu}) \in \mathbb{R}^6$, $\mathbf{M}_u(\mathbf{q}) \in \mathbb{R}^{m \times n + 6}$ and $\mathbf{h}_u(\mathbf{q}, \boldsymbol{\nu}) \in \mathbb{R}^m$, $\mathbf{M}_a(\mathbf{q}) \in \mathbb{R}^{n-m \times n + 6}$ and $\mathbf{h}_a(\mathbf{q}, \boldsymbol{\nu}) \in \mathbb{R}^{n-m}$, the inertia matrices and non-linear terms associated to the (underactuated) floating base, the passive part of the closed linkages and the remaining actuated part, respectively;
$\mathbf{J}_{c,b}(\mathbf{q}) \in \mathbb{R}^{3c \times 6}$ and $\mathbf{J}_{c,u}(\mathbf{q}) \in \mathbb{R}^{3c \times m}$ the underactuated and passive parts of the contact Jacobian associated to the floating-base and the closed kinematic chains respectively and, $\mathbf{J}_{c,a}(\mathbf{q}) \in \mathbb{R}^{3c \times n-m}$ its actuated part;
$\mathbf{J}_{l,u}(\mathbf{q}) \in \mathbb{R}^{m \times m}$ the passive part of the constraint Jacobian and $\mathbf{J}_{l,a}(\mathbf{q}) \in \mathbb{R}^{3c \times n}$ its actuated part.
\par
Taking inspiration from~\cite{mansard2012dedicated}, a possible inverse dynamics (ID) scheme can be composed of the following three steps: 
\begin{enumerate}
\item computation of the generalized accelerations $\boldsymbol{\dot{\nu}}$ and contact forces $\mathbf{F}$ using~\eqref{eq:inverse_dynamics_floating_base} and the constraints~\eqref{eq:kinematic_contact_constraint} and~\eqref{eq:kinematic_closed_chain_constraint},
\item computation of Lagrange multipliers $\boldsymbol{\lambda}$ using~\eqref{eq:inverse_dynamics_passive_part},
\item computation of actuation torques $\boldsymbol{\tau}$ using~\eqref{eq:inverse_dynamics_actuated_part}.
\end{enumerate}
The first step can be addressed using a QP, constraining the contact forces inside linearized friction cones\footnote{Here we do not report further constraints which may include torque limits (by including equations~\eqref{eq:whole_body_id_lambda} and~\eqref{eq:whole_body_id_joint}), joint limits,~\cite{DelPrete18}, and self-collision avoidance,~\cite{Khazoom22}, to name a few}:
\begin{subequations}
\label{eq:whole_body_id_full}
\begin{align}
&\min\limits_{\boldsymbol{\dot{\nu}}, \mathbf{F}} \left\| \mathbf{J}_b(\mathbf{q})\boldsymbol{\dot{\nu}} - \mathbf{a}_{b,d} + \mathbf{\dot{J}}_b(\mathbf{q}, \boldsymbol{\nu})\boldsymbol{\nu} \right\| + \epsilon \left\| \boldsymbol{\dot{\nu}} \right\| + \gamma \left\| \mathbf{F} \right\| \nonumber\\ 
&\text{s.t.} 
\quad \eqref{eq:inverse_dynamics_floating_base}, \eqref{eq:kinematic_contact_constraint}, \eqref{eq:kinematic_closed_chain_constraint} \tag{\ref{eq:whole_body_id_full}}\\
&\quad \quad \mathbf{\bar{F}}^t \geq \mathbf{0}, \quad |\mathbf{\bar{F}}^t| \leq \mu\mathbf{\bar{F}}^n, \nonumber
\end{align}
\end{subequations}
with $\mathbf{J}_b(\mathbf{q}) \in \mathbb{R}^{6 \times 6 + n}$ the \emph{task} Jacobian to move the floating base and $\mathbf{a}_{b,d} \in \mathbb{R}^{6}$ a proper Cartesian acceleration reference for the base, $\mathbf{\bar{F}}^t \in \mathbb{R}^{c}$ and $\mathbf{\bar{F}}^n \in \mathbb{R}^{2c}$ the tangential and normal components, respectively, of the contact forces $\mathbf{F}$ rotated in the contact local frames, $\mu$ the friction cone coefficient, $\epsilon$ and $\gamma$ scalar parameters for regularisation. 
Once optimal $\boldsymbol{\dot{\nu}}$ and $\mathbf{F}$ are computed, the second step permits to compute the Lagrange multipliers:
\begin{equation}
    \boldsymbol{\lambda} = \mathbf{J}_{l,u}(\mathbf{q})^{-T}\left( \mathbf{M}_u(\mathbf{q})\boldsymbol{\dot{\nu}} + \mathbf{h}_u(\mathbf{q}, \boldsymbol{\nu}) -\mathbf{J}_{c,u}(\mathbf{q})^T\mathbf{F} \right).
    \label{eq:whole_body_id_lambda}
\end{equation}
Finally, the third step returns the actuated torques:
\begin{equation}
    \boldsymbol{\tau} = \mathbf{M}_a(\mathbf{q})\boldsymbol{\dot{\nu}} + \mathbf{h}_a(\mathbf{q}, \boldsymbol{\nu}) - \mathbf{J}_{c,a}(\mathbf{q})^T\mathbf{F} - \mathbf{J}_{l,a}(\mathbf{q})^T\boldsymbol{\lambda}. 
    \label{eq:whole_body_id_joint}
\end{equation}
\par
Notice that the QP in~\eqref{eq:whole_body_id_full} can be rewritten \emph{projecting} the passive part into the actuated one and therefore considering only in the actuated quantities since the passive ones are dependent variables (for more details see Appendix~\ref{app:B}).

\section{Analysis of Kangaroo Lower Body}\label{sec:Kangaroo}
\begin{figure}[htb!]
    \centering
    \includegraphics[width=1.\columnwidth, trim={6cm 2cm 6cm 2cm}, clip=true]{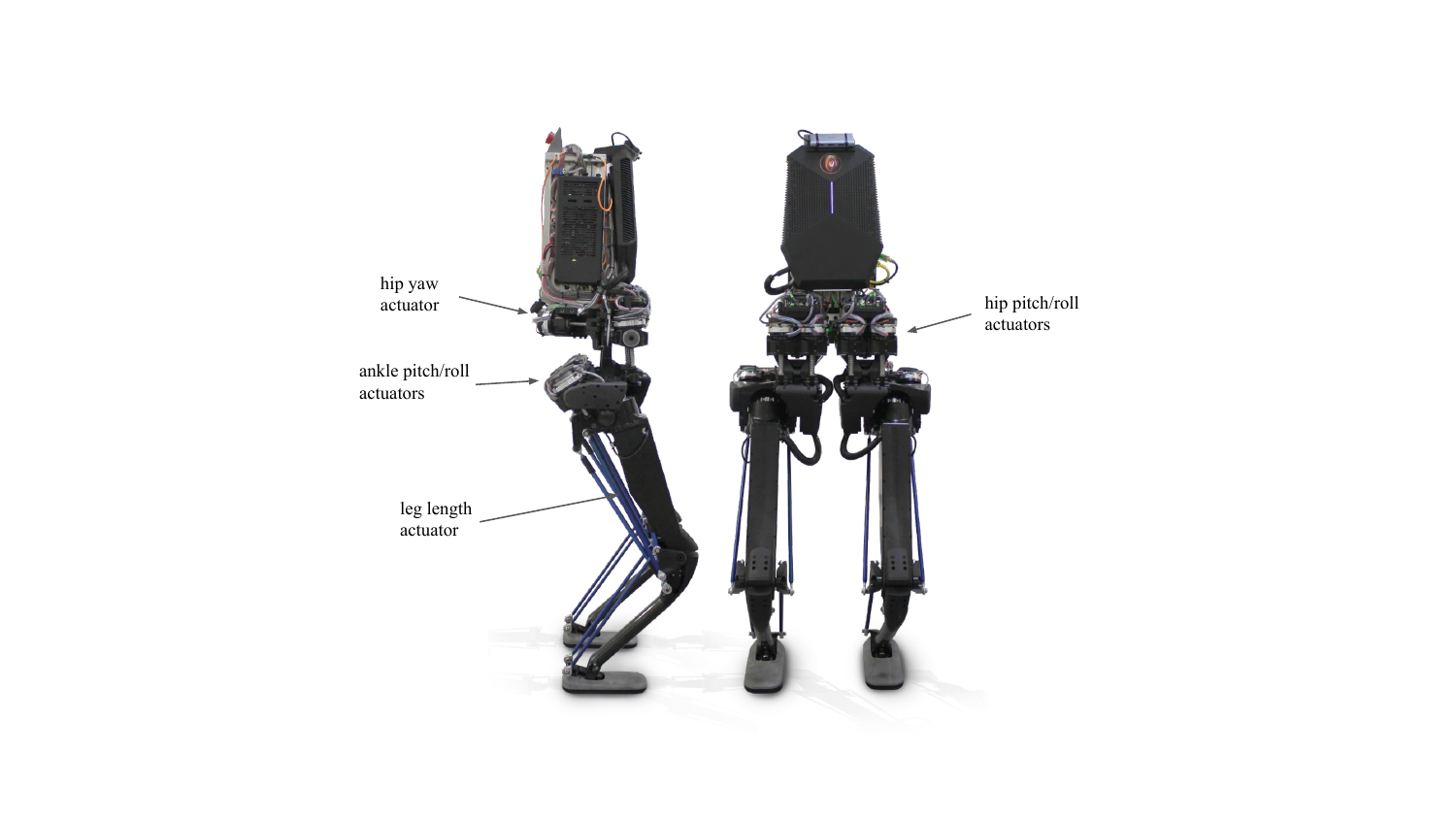}
    \caption{Prototype of the lower body of the Kangaroo robot.}
    \label{fig:kangaroo_hw}
\end{figure}

Fig.~\ref{fig:kangaroo_hw} shows the prototype of the Kangaroo robot highlighting the location of the linear actuators, which are all placed near the pelvis area.
The lower body of the Kangaroo robot consists of two legs, each one modeled with $32$ passive rotational DOFs and $6$ actuated linear DOFs. 
Therefore, the total number of DOFs in the Kangaroo model is $76$, comprising of $64$ passive DOFs and $12$ active DOFs.\footnote{Here, we are not considering the floating-base DOFs}.
The lower body prototype weighs $\approx 42 \ [kg]$ with the protection cages for the torso, not present in Fig.~\ref{fig:kangaroo_hw}, and $\approx 36 \ [kg]$ without them.
All the linear actuators, except for those located at the thighs, are equipped with relative linear encoders, while the thigh actuators are equipped with absolute linear encoders. 
To calibrate the relative linear encoders, an initial calibration is performed using absolute encoders mounted at certain passive DOFs.
\par
\begin{figure}[htb!]
    \centering
    \includegraphics[width=1.\columnwidth, trim={7.5cm 0cm 7.2cm 0cm}, clip=true]{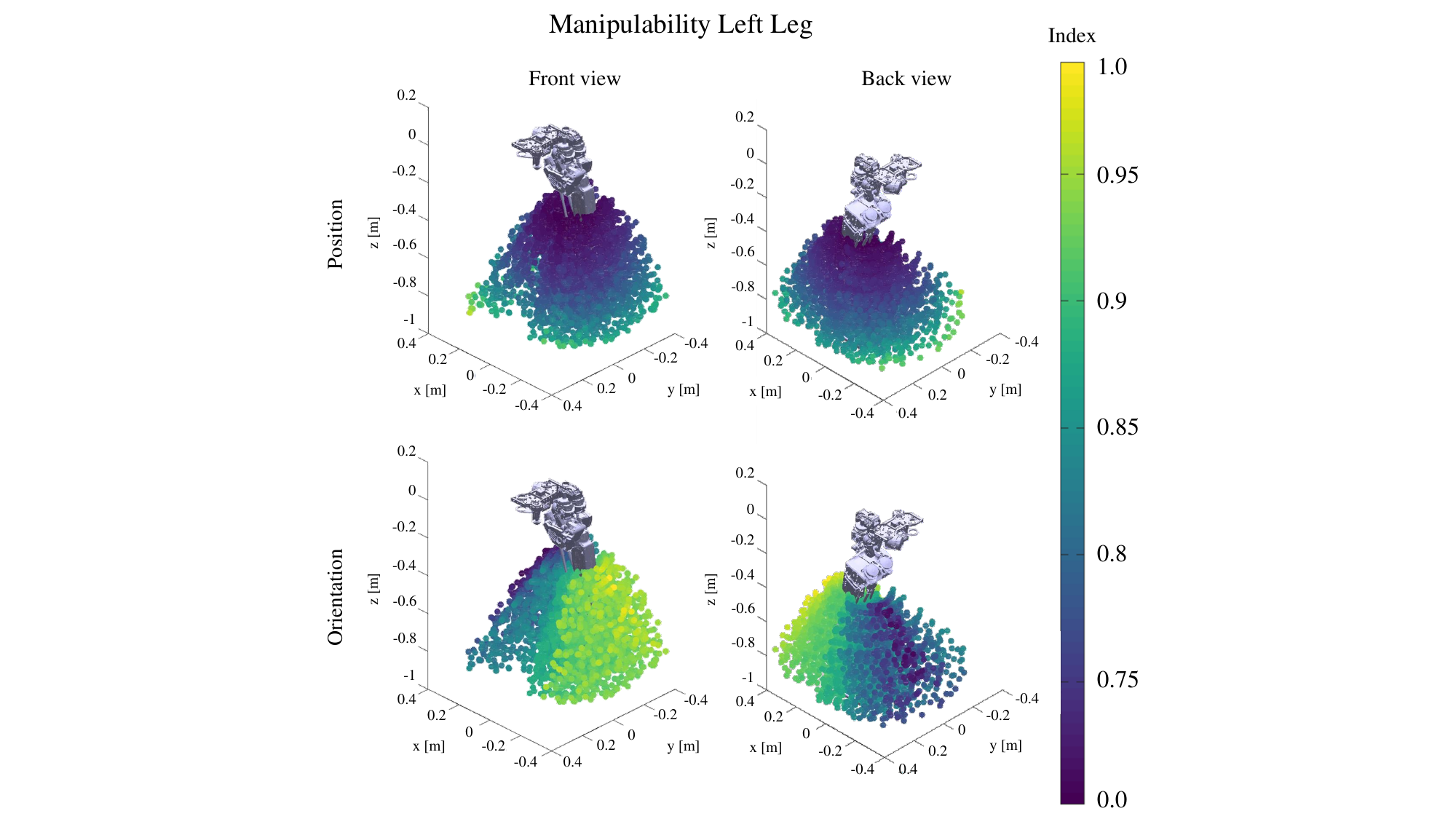}
    \caption{Normalized manipulability of the left leg. Position shown at the top, orientation at the bottom. View from the front on the left, and rear view on the right.}
    \label{fig:manipulability}
\end{figure}

Each leg of Kangaroo is a hybrid serial-parallel kinematic chain formed by $4$ closed sub-mechanisms:
\begin{itemize}
    \item 1DOF RR\underline{P}R \emph{Hip Yaw},
    \item 2DOFs U-2RR\underline{P}U \emph{Hip Pitch/Roll},
    \item 1DOF RRRR-RRR\underline{P}-R \emph{Knee},
    \item 2DOFs U-2(RRR\underline{P}-RRR-UU) \emph{Ankle Pitch/Roll}.
\end{itemize}
We will now discuss each sub-mechanism separately, introducing a brief analysis of the kinematics singularities and the countermeasures taken to avoid them in the design of the platform.
Table~\ref{tab:overview} reports an overview of the closed sub-mechanisms with the number of active and passive DOFs (with the number of total passive DOFs), the number of constraints to model the linkage, the actuators limits, the DOFs limits, and the average human limits at the same DOF (see~\cite{brockett2016biomechanics,lee2014design}).  
\begin{table*}[h]
\caption{Overview of the number of parallel sub-mechanism based modules in Kangaroo lower body} 
\centering 
\begin{tabular}{l c c c c c c c | c c}
Closed sub-mechanism        & \#Active DOFs & \#Passive DOFs & \#Constraints & \multicolumn{2}{c}{Actuator Lims [m]} & \multicolumn{2}{c}{DOFs Lims [deg]}  & \multicolumn{2}{c}{Average Human Lims [deg]} \\ 
                            \cline{5-6}  
                            \cline{7-8}
                            \cline{9-10}
                            &   &  &  & min & max & min & max & min & max \\
\hline
\hline
\makecell[l]{Hip Yaw} & 1 & 2(3) & 2 & -0.02 & 0.02 & -30 & 30 & -30 & 60 \\
\hline
\makecell[l]{Hip Roll \\ Hip Pitch} & 2 & 8(10) & 8 &  -0.04 & 0.04 & \makecell{ -27 \\ -40} & \makecell{ 27 \\ 42} & \makecell{ -30 \\ -135} & \makecell{ 45 \\ 20}\\
\hline
\makecell[l]{Knee} & 1 & 6(8) & 6 & 0 & 0.15 & 0 & 120 & 0 & 130 \\
\hline
\makecell[l]{Ankle Roll \\ Ankle Pitch} & 2 & 16(24) & 16 &  -0.02 & 0.02 & \makecell{ -27 \\ -36} & \makecell{ 27 \\ 34} & \makecell{ -45 \\ -12} & \makecell{ 15 \\ 23}\\
\hline
\end{tabular}
\label{tab:overview}
\end{table*}
In Figure \ref{fig:manipulability}, we present the normalized manipulability index for the left leg. 
This evaluation is conducted through a uniform sampling of the workspace, achieved by varying the linear actuators within the Hip Pitch/Roll and Knee sub-mechanisms.

\subsubsection*{\bf{1DOF Hip Yaw}}\label{subsec:hip_yaw}
\begin{figure}[htb!]
    \centering
    \includegraphics[width=1.\columnwidth, trim={0cm 1cm 0cm 0cm}, clip=true]{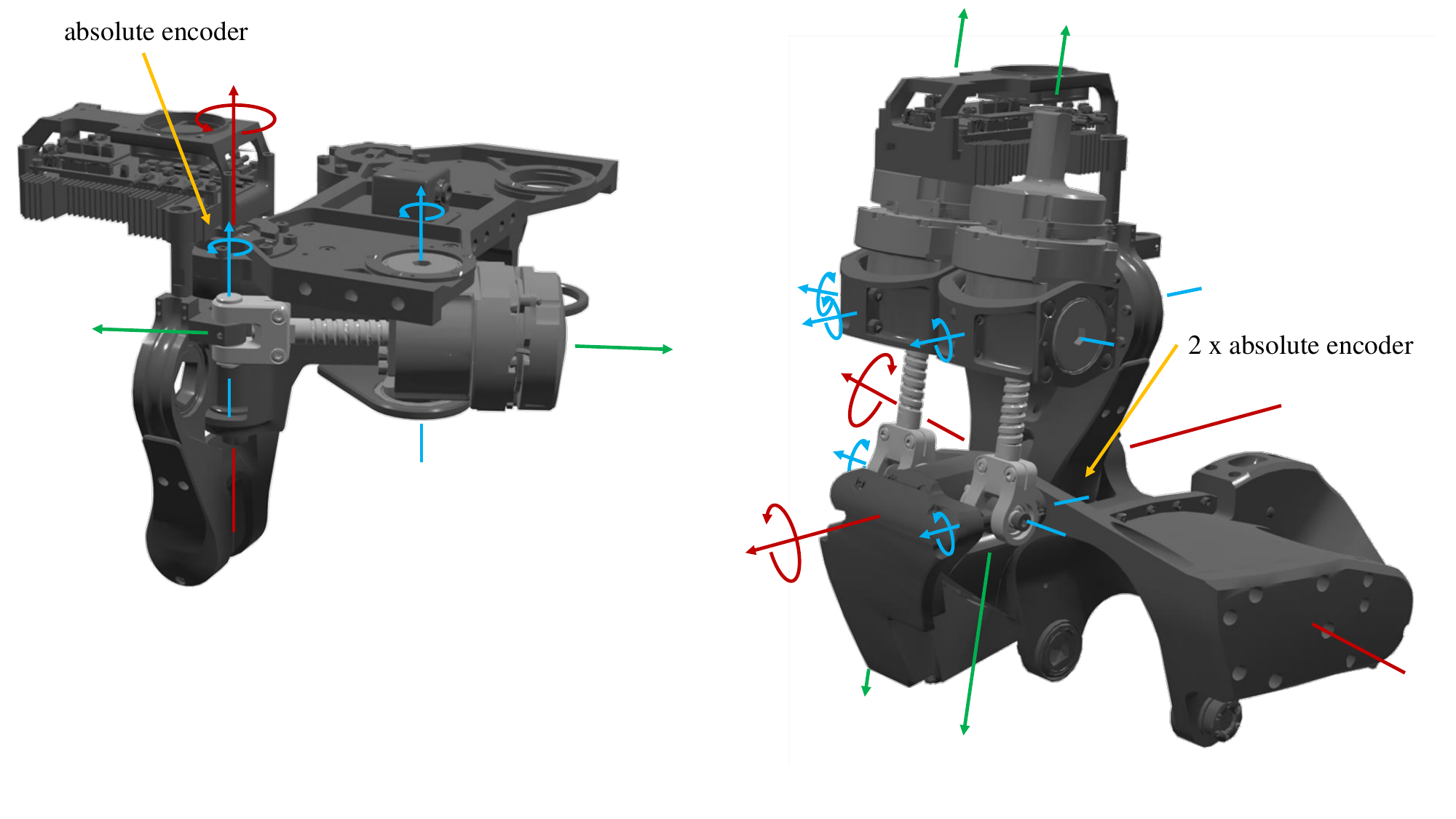}
    \caption{On the left is the Hip Yaw sub-mechanism, and on the right is the Hip Pitch/Roll sub-mechanism. The green arrows show the linear actuators, in blue the passive DOFs, and in red the main passive DOF(s) of the sub-mechanisms. The yellow arrow shows the position of the absolute rotary encoders used for initial calibration.}
    \label{fig:hip_yaw_pitch_roll}
\end{figure}

The Hip Yaw closed sub-mechanism 3D model, whose schematic is depicted in Fig.~\ref{fig:crank_slider}, is shown in the left portion of Fig.~\ref{fig:hip_yaw_pitch_roll}.
This sub-mechanism exhibits two configuration singularities, occurring when the passive DOF $\theta_{u,1}$ (see Fig.~\ref{fig:crank_slider}) is at $0 \ [rad]$ and $-\pi \ [rad]$. 
These singularities correspond to the linear actuator being approximately driven at $0.05 \ [m]$ and $-0.05 \ [m]$ respectively, resulting in the sub-mechanism being reduced to a linear configuration.
Nonetheless, due to the limitations of the linear actuator of $(-0.02, 0.02) \ [m]$, the sub-mechanism is unable to attain the singular configurations.
\subsubsection*{\bf{2DOFs Hip Pitch/Roll}}\label{subsec:hip_pitch_roll}
\begin{figure}
     \centering
     \includegraphics[width=1.\columnwidth, trim={2cm 0cm 2cm 0cm}, clip=true]{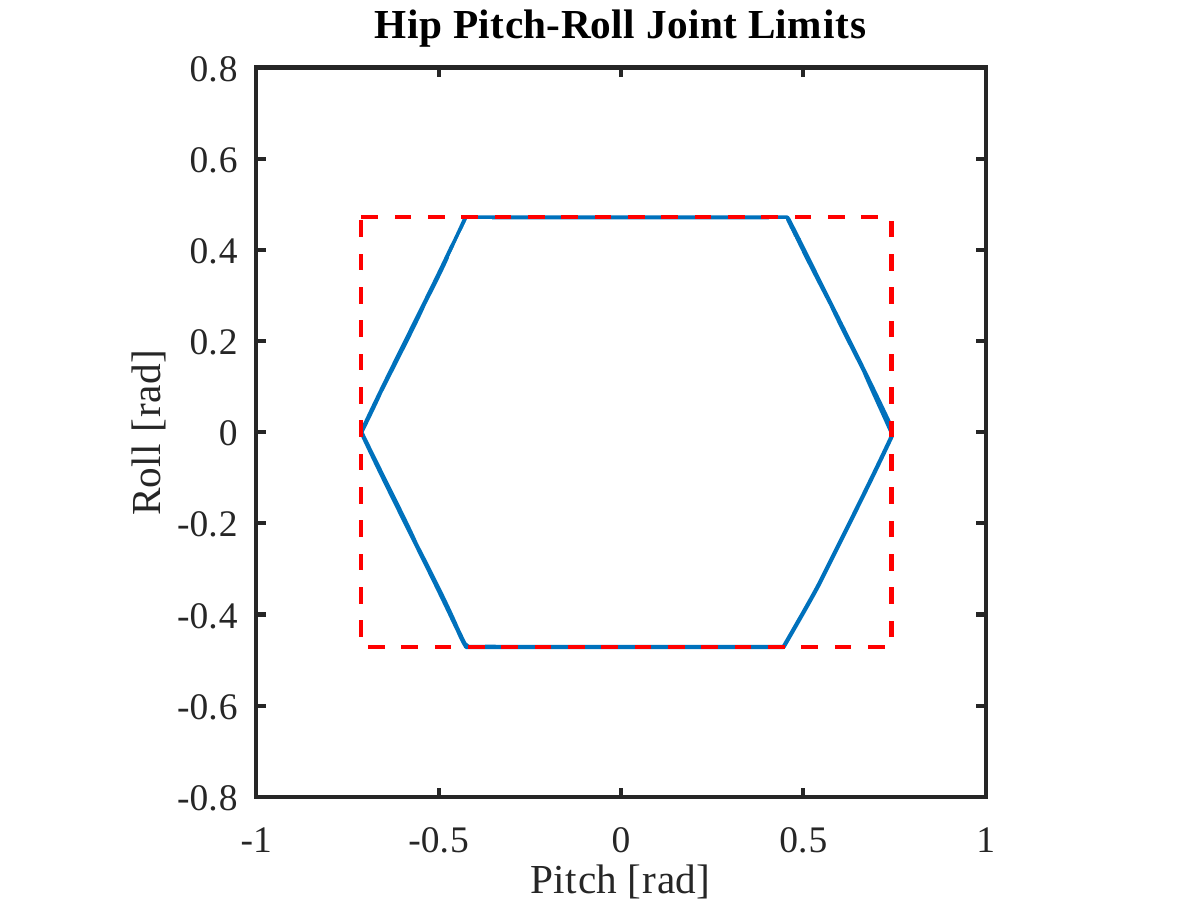}
     \caption{Joint limits in 2DOFs U-2RR\underline{P}U Hip Pitch/Roll. In blue (continuous line) are the real joint limits projected into the differential mechanism, while in red (dashed line) are the joint limits, as box constraints, when not projected.}
     \label{fig:joint_limits_differential_hip}
\end{figure}

The Hip Pitch/Roll, whose 3D model is shown in the right portion of Fig.~\ref{fig:hip_yaw_pitch_roll}, consists of a hybrid serial-parallel sub-mechanism equivalent to the schematic in Fig.~\ref{fig:spatial_differential_linkage}.
This sub-mechanism introduces convex non-linear joint limits that can be computed knowing both the position limits in the linear actuator and passive DOFs, if any. 
In particular, it is possible to cast a QP-based IK, constrained by those limits, to explore the workspace of the sub-mechanism:
\par
\begin{subequations}
\label{eq:qp_joint_limits}
\begin{align}
&\min\limits_{\boldsymbol{\dot{\theta}}} \left\|\boldsymbol{\dot{\theta}}_{a,d} - \boldsymbol{\dot{\theta}}_{a} \right\| \nonumber\\ 
&\text{s.t.}  
\quad \quad {\mathbf{J}_l}\left(\boldsymbol{\theta}\right)\boldsymbol{\dot{\theta}} = \mathbf{0}, \tag{\ref{eq:qp_joint_limits}}\\
& \quad \quad \quad \frac{\boldsymbol{\underline{\theta}} - \boldsymbol{\theta}}{dt} \leq \boldsymbol{\dot{\theta}} \leq \frac{\boldsymbol{\overline{\theta}} - \boldsymbol{\theta}}{dt}, \nonumber
\end{align}
\end{subequations}
with $\left[ \boldsymbol{\underline{\theta}}, \boldsymbol{\overline{\theta}} \right]$ the minimum and maximum DOFs limits, respectively. 
By driving the linear actuators is possible to reconstruct the shape of the interested joints, as shown in Fig.~\ref{fig:joint_limits_differential_hip}.
Recalling Fig.~\ref{fig:2DOFsU-2RRPU}, a first set of singularities appear when the constraint $^a\vartheta_u^1$ and $^a\vartheta_u^2$ reach the $0 \ [rad]$ and $-\pi \ [rad]$ configurations, respectively when the two linear actuators are simultaneously driven to approximately $-0.07 \ [m]$ and $0.05 \ [m]$.
Nevertheless, both singularities lie outside the linear actuator limits, $(-0.04, 0.04) \ [m]$. 
Unfortunately, a singularity arises when the two linear actuators are driven differentially to their extremes, such as $-0.04 \ [m]$ and $0.04 \ [m]$, causing $\theta_{u,8}$ to approach $-\frac{\pi}{2} \ [rad]$ or $\frac{\pi}{2} \ [rad]$. 
To address this issue, a physical constraint has been implemented for the Hip Roll, which can be enforced at the actuator level via the QP formulation in~\eqref{eq:qp_joint_limits}.

\subsubsection*{\bf{1DOF Knee}}\label{subsec:knee}
\begin{figure}[htb!]
    \centering
    \includegraphics[width=1.\columnwidth, trim={6cm 0cm 6cm 0cm}, clip=true]{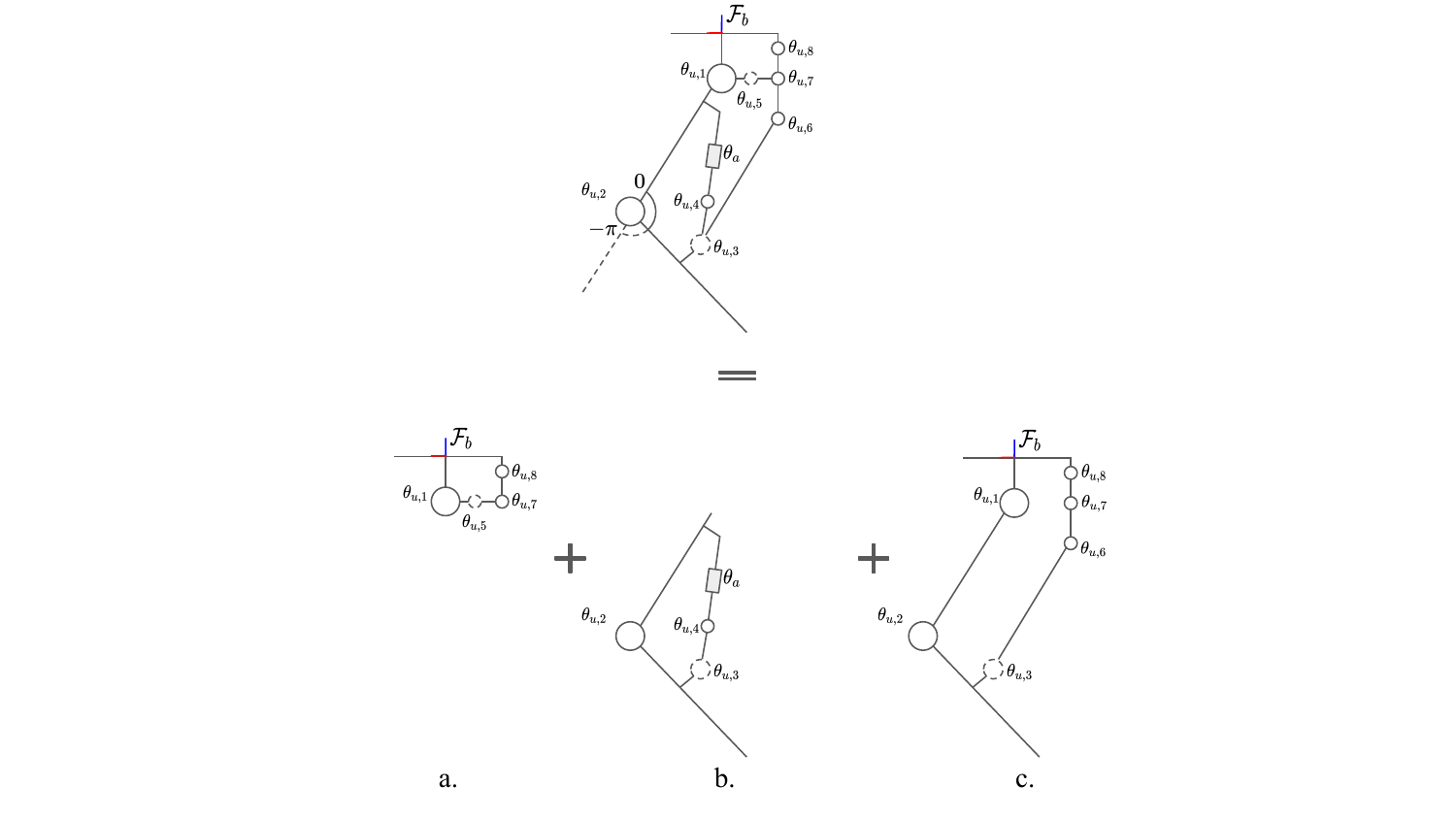}
    \caption{1DOF RRRR-RRR\underline{P}-R Knee simplified sub-mechanisms: a. RRRR introducing $2$ constraints; b. RRR\underline{P}, introducing $2$ constraints; c. RRRRRR introducing $2$ constraints. Some passive DOFs are shared between the linkages.}
    \label{fig:schematic_knee}
\end{figure}

The Knee sub-mechanism is formed by the three closed planar linkages depicted in Fig.~\ref{fig:schematic_knee}.
This is the only mechanism that presents an absolute linear encoder mounted directly on the linear actuator, thus not requiring any initial calibration.
The 3D model of the Knee is presented in the left part of Fig.~\ref{fig:leg_length_ankle}, providing an insight into the thigh.
The knee exhibits two primary singularities when the passive DOF $\theta_{u,2}$ (see Fig.~\ref{fig:schematic_knee}) is in configurations of $0 \ [\text{rad}]$ and $-\pi \ [\text{rad}]$, respectively.
In these instances, the linear actuator situated within the thigh is approximately positioned at $0.095 \ [\text{m}]$ and $-0.095 \ [\text{m}]$, and the sub-mechanism is reduced to a line\footnote{For this analysis we are considering the $0$ of the linear actuator centered at the $-\frac{\pi}{2}$ configuration of the $\theta_{u,2}$ DOF, see Fig.~\ref{fig:schematic_knee}}.
Nonetheless, the linear actuator is confined to a stroke of $0.15 \ \text{m}$, bounded by limits of $(-0.075, 0.075) \ \text{m}$, preventing the Knee from attaining any singular configuration; furthermore the passive Knee motion is additionally mechanically restricted by plastic shells installed in front of the joint, serving the purpose of safeguarding the robot in the event of a fall, as depicted in Fig.~\ref{fig:kangaroo_hw}.
%
%
\subsubsection*{\bf{2DOFs Ankle Pitch/Roll}}\label{subsec:ankle_pitch_roll}
\begin{figure}[htb!]
    \centering
    \includegraphics[width=1.\columnwidth, trim={0cm 0cm 0cm 0cm}, clip=true]{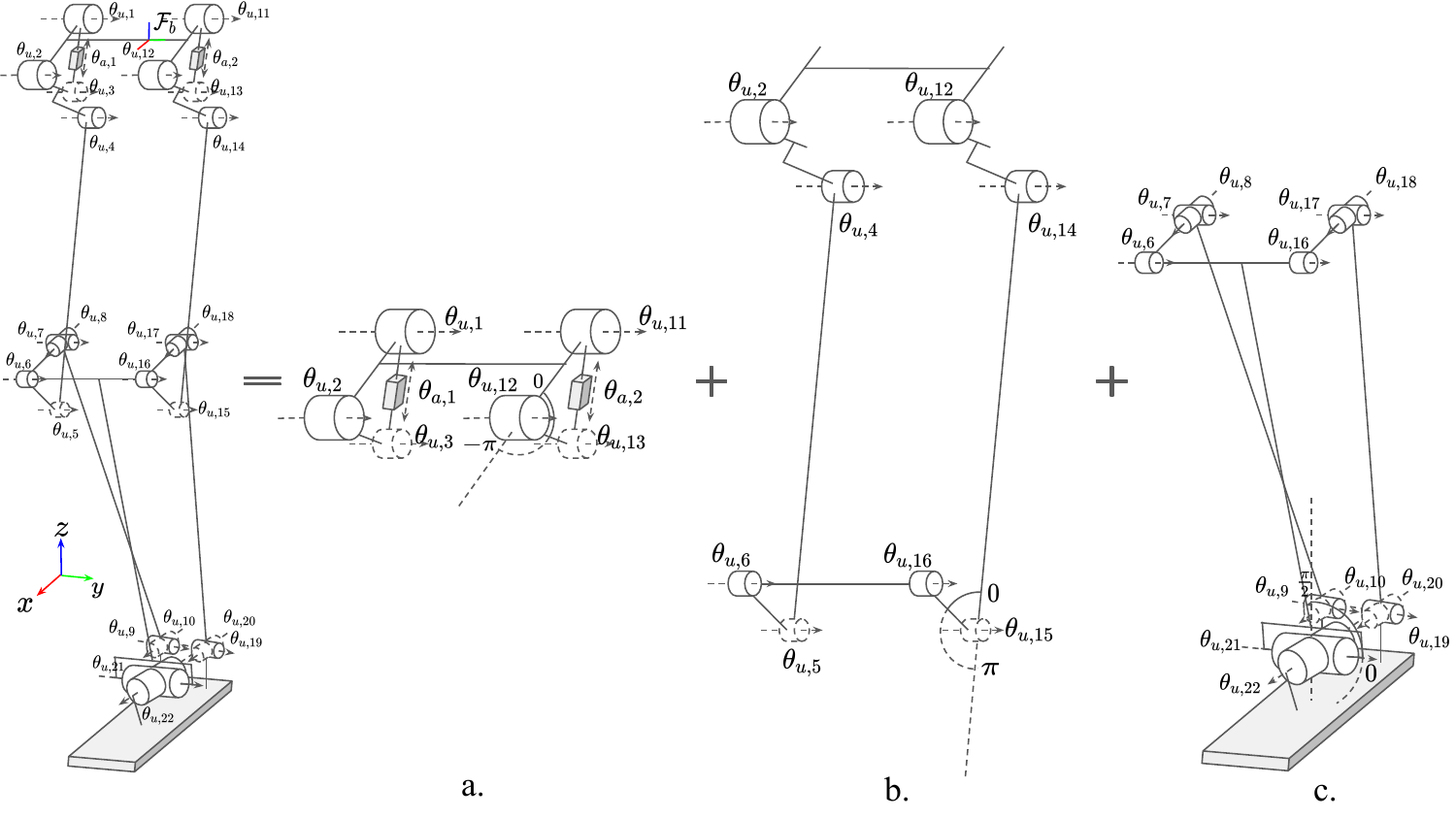}
    \caption{2DOFs U-2(RRR\underline{P}-RRR-UU) Ankle Pitch/Roll simplified sub-mechanisms: a. 2RRR\underline{P} placed at the hip, introducing $4$ constraints; b. 2RRRR  transferring the motion from the hip to the knee introducing $4$ constraints; c. U-2UR introducing $8$ constraints. Some passive DOFs are shared between the linkages.}
    \label{fig:schematic_ankle}
\end{figure}

\begin{figure}[htb!]
    \centering
    \includegraphics[width=1.\columnwidth, trim={9cm 0cm 0cm 0cm}, clip=true]{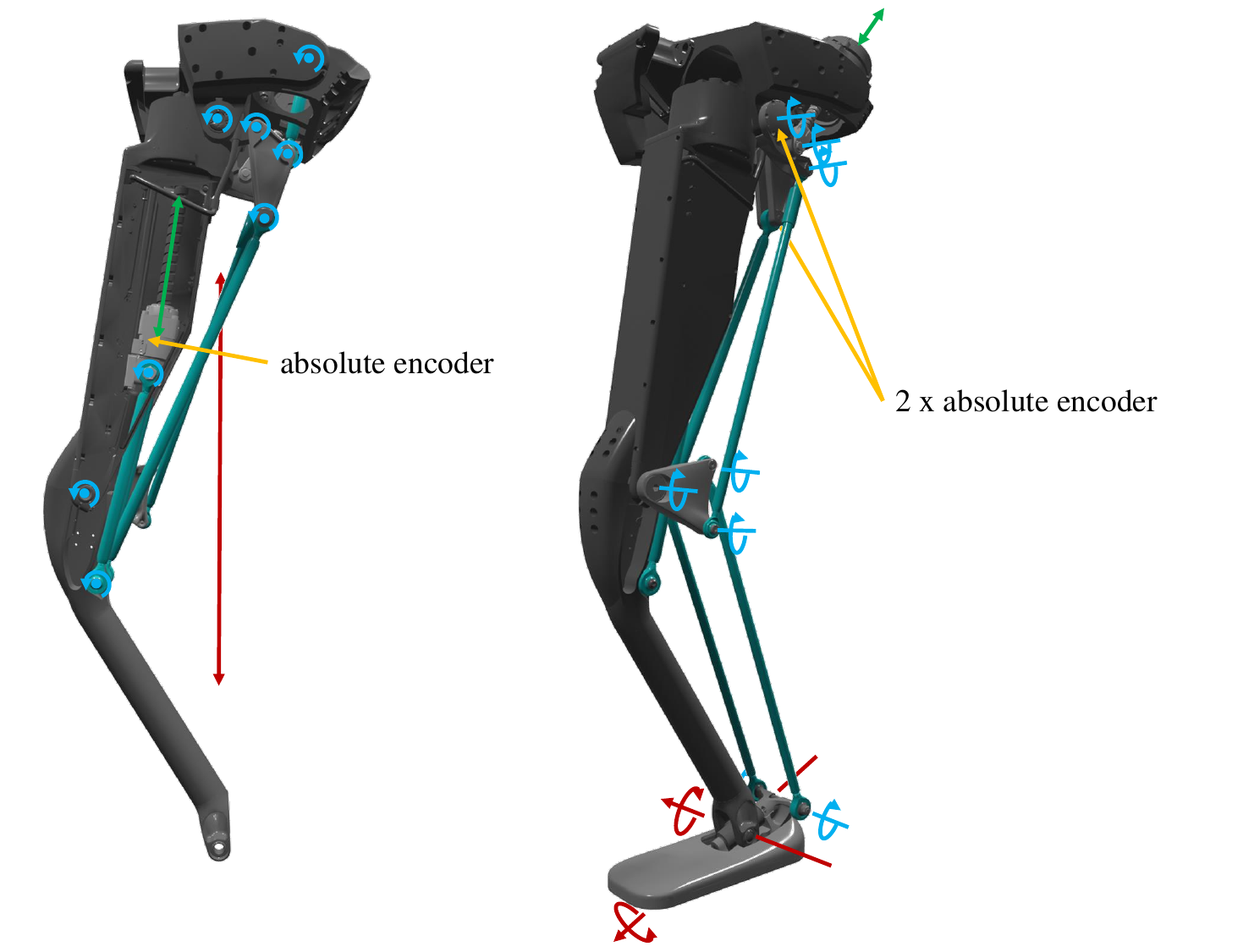}
    \caption{On the left is the Knee sub-mechanism, and on the right is the Ankle Pitch/Roll sub-mechanism. The green arrows show the linear actuators, in blue the passive DOFs, and in red the main passive DOF(s) of the sub-mechanisms. The yellow arrow shows the position of the linear absolute encoder for the Knee and the absolute rotary encoders for the Ankle Pitch/Roll used for initial calibration.}
    \label{fig:leg_length_ankle}
\end{figure}

\begin{figure}
     \centering
     \includegraphics[width=1.\columnwidth, trim={2cm 0cm 2cm 0cm}, clip=true]{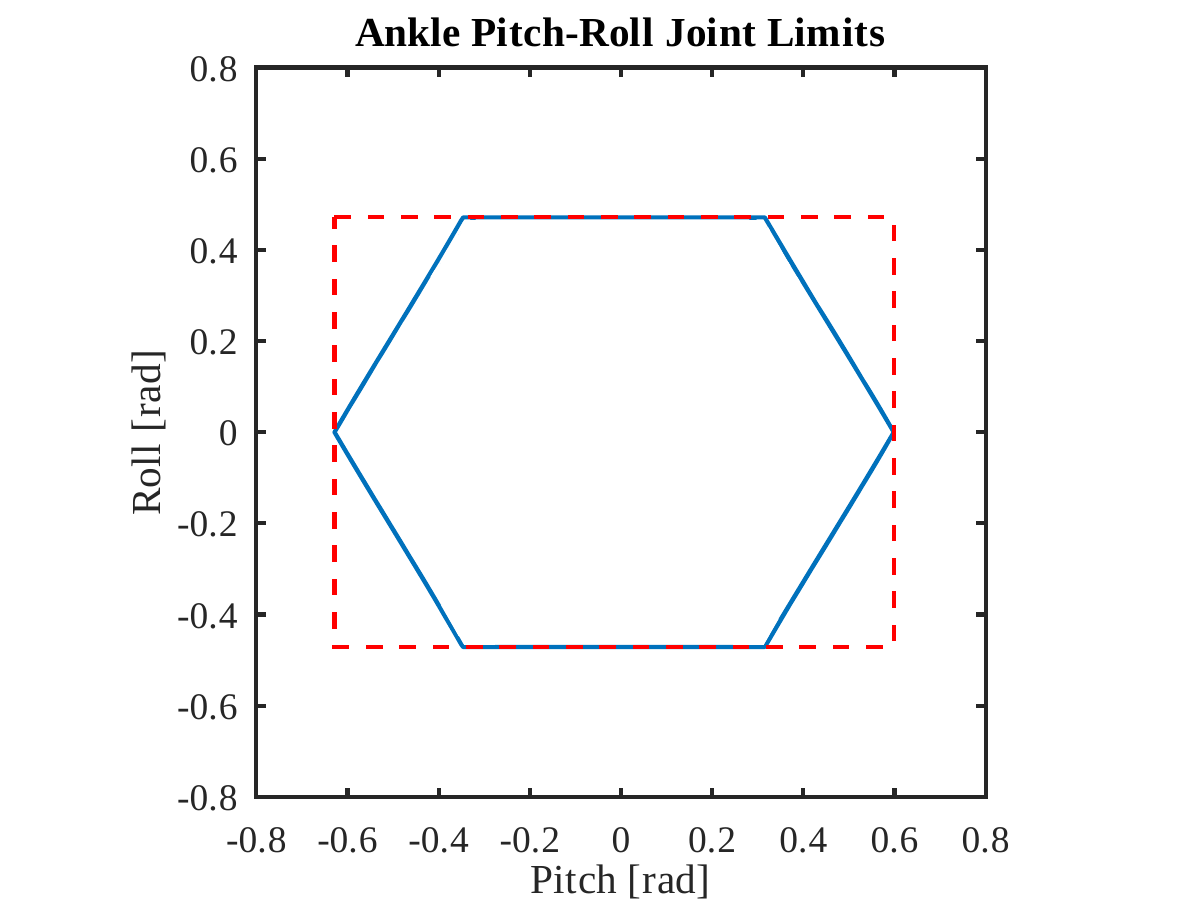}
     \caption{Joint limits in 2DOFs U-2(RRR\underline{P}-RRR-UU) Ankle Pitch/Roll. In blue (continuous line) are the real joint limits projected into the differential mechanism, while in red (dashed line) are the joint limits, as box constraints, when not projected.}
     \label{fig:joint_limits_differential_ankle}
\end{figure}

The Ankle consists of a hybrid serial-parallel sub-mechanism formed by the three sub-mechanisms depicted in Fig.~\ref{fig:schematic_ankle}.
The right model in Fig.~\ref{fig:leg_length_ankle} shows the 3D visualization of the sub-mechanism.
Joint limits are computed using the same methodology as for the 2DOFs Hip Pitch/Roll and reported in Fig.~\ref{fig:joint_limits_differential_ankle}.
The first singularity encountered in the Ankle occurs within sub-mechanism \emph{b.} as depicted in Fig.~\ref{fig:schematic_ankle} when the passive DOFs $\theta_{u,5}$ and $\theta_{u,15}$ reach $0 \ \text{rad}$, corresponding to drive independently the linear actuators to approximately $-0.035 \ \text{m}$.
%
%
A second singularity arises within sub-mechanism \emph{a.}, where the passive DOFs $\theta_{u,2}$ and $\theta_{u,12}$ also assume values of $0 \ \text{rad}$, coinciding with the separate actuation of linear actuators to approximately $0.027 \ \text{m}$.
%
Nevertheless, the linear actuators of the Ankle are limited to a stroke of $0.04 \ [m]$, with $(-0.02, 0.02) \ [m]$ limits, preventing the aforementioned singular configurations and any singularity in common mode.
The differential mode remains free of singularities within the actuator limits.
However, these singularities are linked to the passive degree of freedom (DOF) $\theta_{u,22}$ reaching ${\frac{\pi}{2} \ [\text{rad}]}$ and ${-\frac{\pi}{2} \ [\text{rad}]}$ (see Fig.~\ref{fig:schematic_ankle}), arises when the linear actuators are driven differentially, just slightly surpassing their limitations.

\subsection{Non-Linear Transmission Analysis}
\begin{figure}[htb!]
    \centering
    \includegraphics[width=1.\columnwidth, trim={0.5cm 0cm 1cm 0cm}, clip=true]{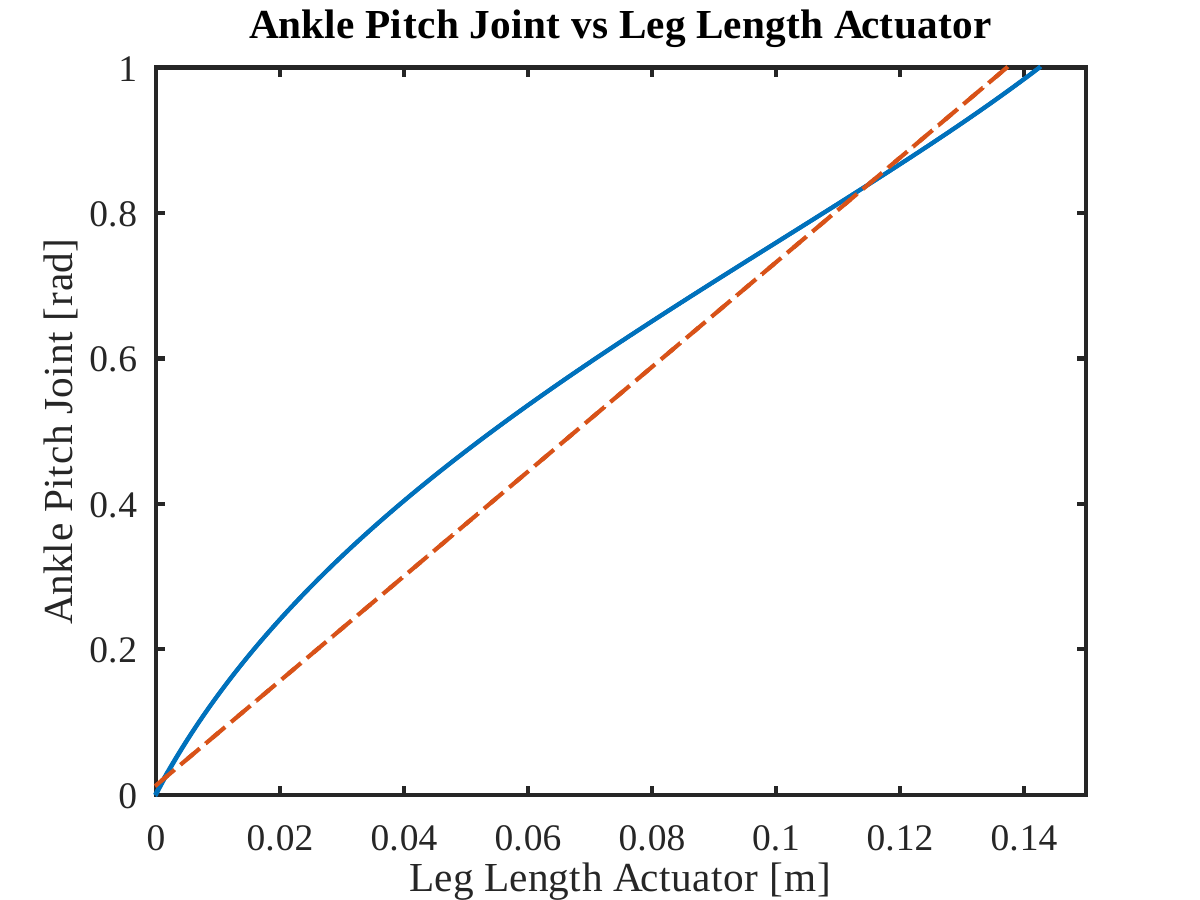}
    \caption{In blue the (non-linear) relation between actuated Leg Length linear actuator and rotational knee pitch joint, retracting and extending the leg. The dashed line shows the line fitting the curve.}
    \label{fig:leg_length_vs_ankle_pitch}
\end{figure}

\begin{figure}[htb!]
    \centering
    \includegraphics[width=1.\columnwidth, trim={0.5cm 0cm 1cm 0cm}, clip=true]{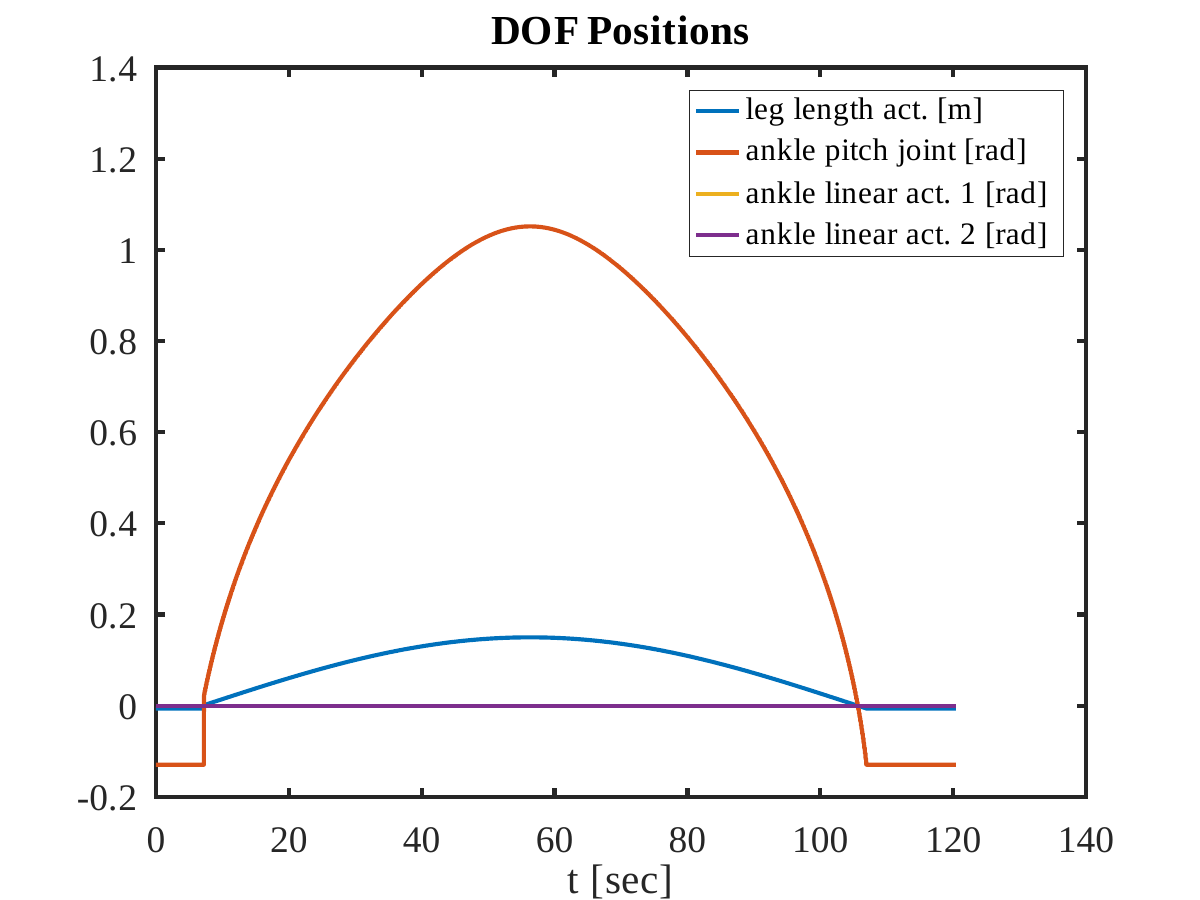}
    \caption{Passive and active DOFs come into play during both the retraction and extension of the leg. The orange line represents the variation in the Ankle pitch joint which preserves the foot's orientation w.r.t. the hip. Notice that this adjustment is achieved with no intervention of the Ankle actuators.}
    \label{fig:dofs_1}
\end{figure}

\begin{figure}[htb!]
    \centering
    \includegraphics[width=1.\columnwidth, trim={0.5cm 0cm 1cm 0cm}, clip=true]{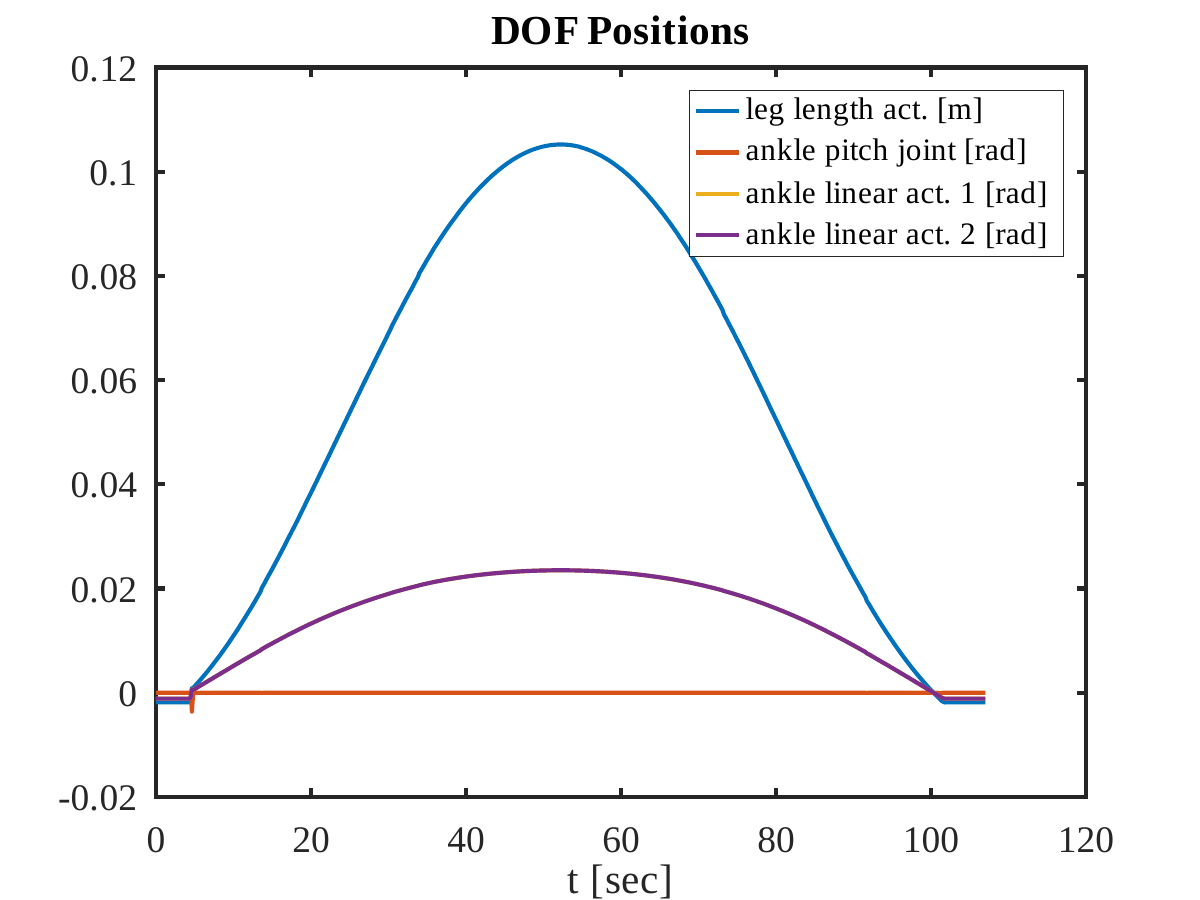}
    \caption{Passive and active DOFs during retraction and extension of the leg while keeping constant the Knee pitch joint. The violet and yellow lines show that to preserve the joint's position (orange line) throughout the motion, adjustments to the actuated Ankle DOFs are also required.}
    \label{fig:dofs_2}
\end{figure}

The Ankle Pitch/Roll together with the Knee, form a non-linear transmission (see Fig.~\ref{fig:leg_length_vs_ankle_pitch}) that permits keeping the orientation of the foot constant w.r.t. the Hip when the Knee linear actuator retracts/extends. 
In contrast with classical leg kinematics, this is achieved without moving the Ankle Pitch/Roll actuators, permitting, for example, in-place jumps by moving only the linear actuator located at the thigh, as shown in Fig.~\ref{fig:dofs_1}.
On the contrary, to keep constant the value of the Ankle Pitch joint when moving retracting/extending the leg, it is necessary to actuate the Ankles linear actuators, as shown in Fig.~\ref{fig:dofs_2}.
\par
\begin{figure}[htb!]
    \centering
    \includegraphics[width=1.\columnwidth, trim={1cm 0.5cm 1.5cm 0.4cm}, clip=true]{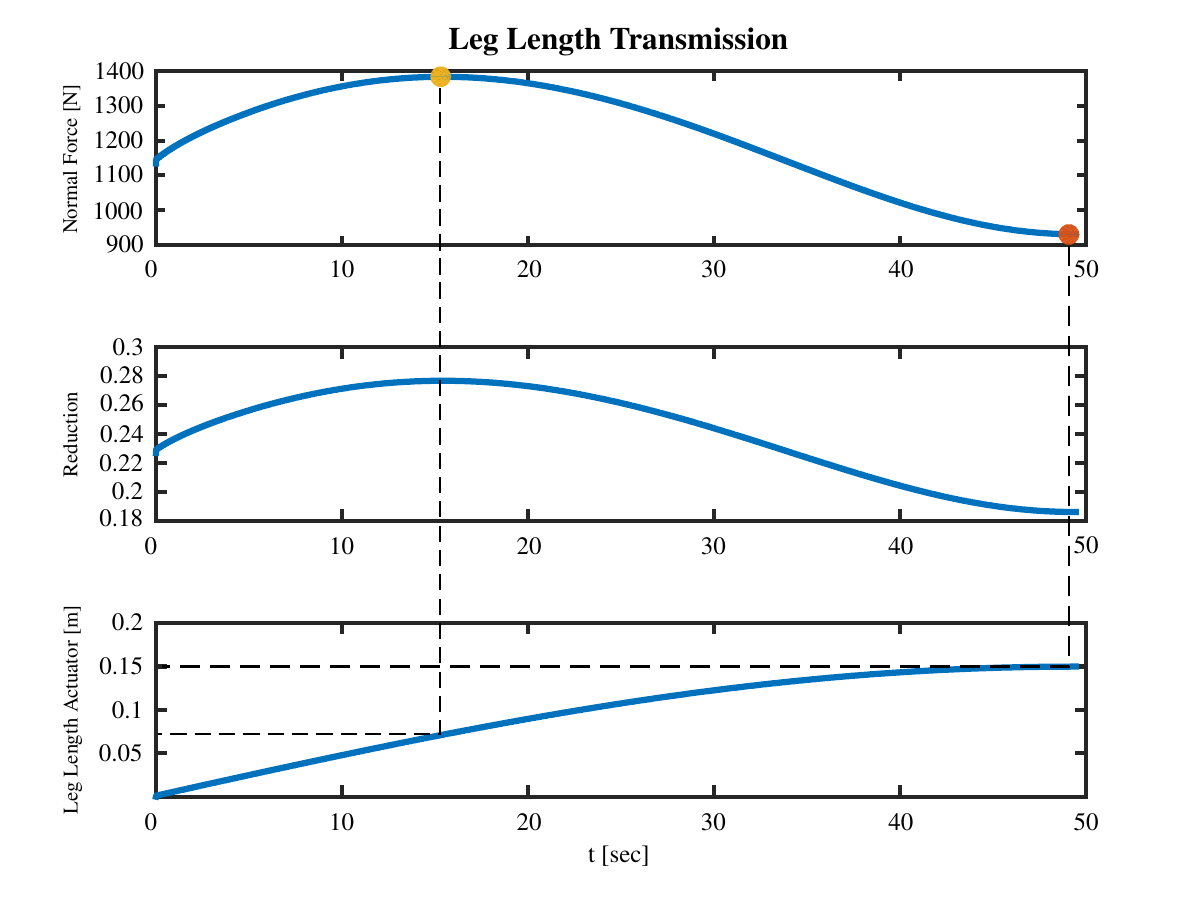}
    \caption{Force Transmission from Knee Linear Actuator to Foot Sole. The topmost plot illustrates the normal force at the sole while performing a squat and applying a maximum linear force ($5000 \ N$) through the actuator. The maximum output force is depicted in yellow, while the minimum is shown in red. The middle plot displays the transformation ratio. The bottom plot depicts the actuator's position during the squat. Observing the graphs, the highest normal foot force is attained when the legs are half-compressed, whereas the lowest force is experienced when fully compressed.}
    \label{fig:force_transmission}
\end{figure}

Finally, Fig.~\ref{fig:force_transmission} reports the non-linear transformation between Leg Length actuator force to foot sole normal force, while performing a squat motion and applying the maximum force at the actuator ($5000 \ N$).
It is worth noticing that at the nominal configuration, the foot's sole normal force is between its minimum and maximum, when completely compressed it reaches its minimum, while at half compression, it reaches its maximum. 
From these plots, we can see that the non-linear transmission has been designed to be able to exert its maximum force when half-squatting, for example during a jump preparation.
Being torque and speed inversely related, the non-linear transmission design was optimized to achieve high force when the robot is still standing at a normal height and high speed when the leg is crouched to accelerate for jumping efficiently.

\section{Comparison with TALOS}\label{sec:comparison}
In this section, we conduct a comparison with another full-size humanoid bipedal system, the TALOS robot, developed by PAL Robotics~\cite{stasse2017talos}. 
TALOS features a more traditional serial kinematics design for its legs, encompassing 6 actuated DOFs per leg, with motors distributed along the entire chain.
\begin{figure}[htb!]
    \centering
    \includegraphics[width=1.\columnwidth, trim={3cm 1.5cm 2cm 1.5cm}, clip=true]{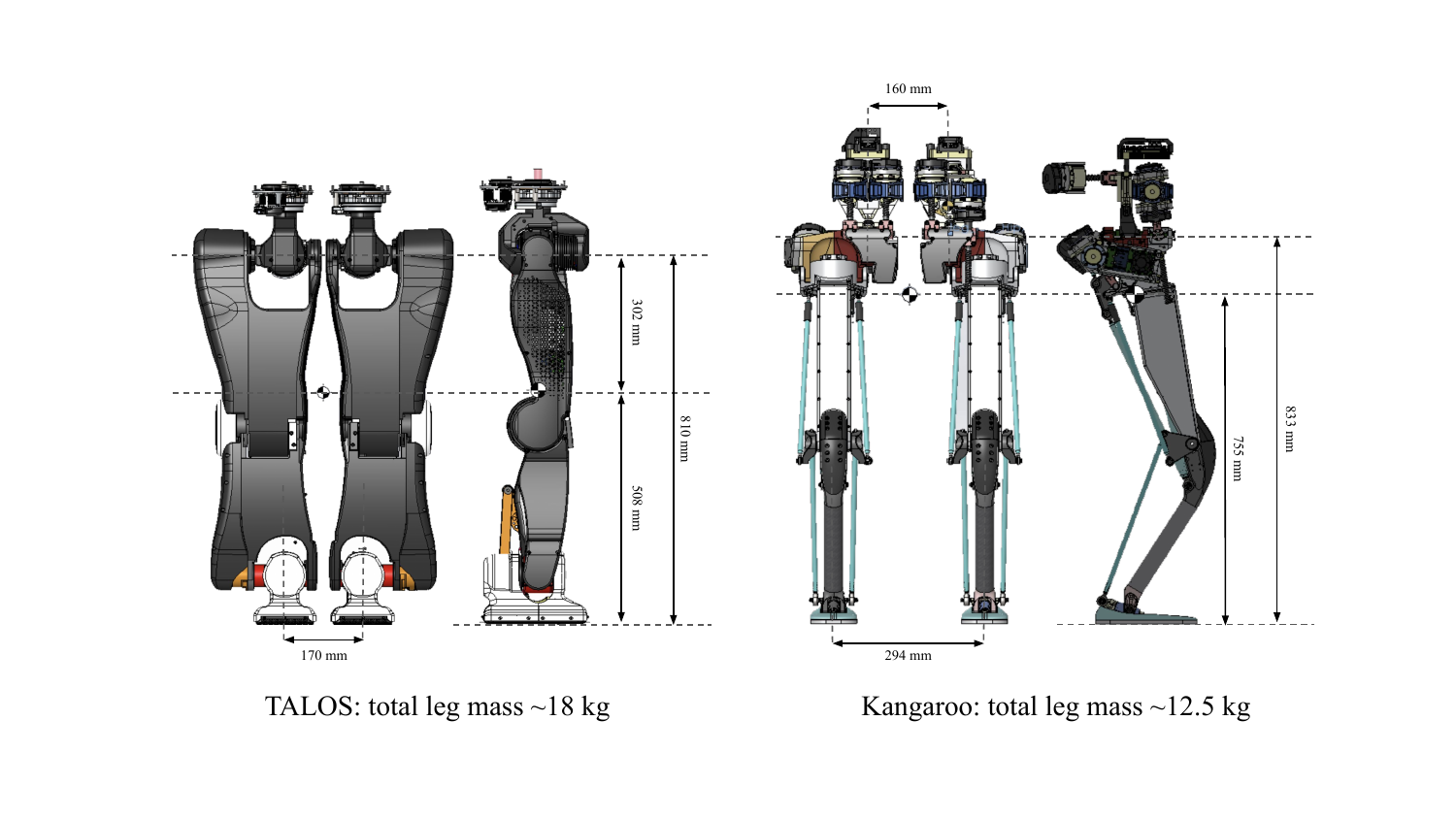}
    \caption{TALOS and Kangaroo lower-body comparison.}
    \label{fig:kangaroo_talos_com}
\end{figure}
Fig.~\ref{fig:kangaroo_talos_com} illustrates a comparison of the lower bodies between TALOS and Kangaroo.
Notably, Kangaroo exhibits a lower total mass compared to TALOS, and its CoM is positioned close to the pelvis. 
In contrast, TALOS has its CoM situated at knee height.
\par
Let's now proceed to compare the equivalent Cartesian inertia matrix at the foott of both the Kangaroo and the TALOS robot.
Considering an open-kinematic chain, the equivalent Cartesian inertia matrix \emph{seen} at a certain frame $\mathcal{F}_l$ w.r.t. the base frame $\mathcal{F}_b$ can be computed as:
\begin{equation}
    \boldsymbol{\Lambda} = \left({\mathbf{J}}\mathbf{M}^{-1}{\mathbf{J}}^T + \epsilon\mathbf{I}\right)^{-1},
    \label{eq:cartesian_inertia_matrix}
\end{equation}
with $\boldsymbol{\Lambda} \in \mathbb{R}^{6 \times 6}$, $\mathbf{J}$ the Jacobian of $\mathcal{F}_l$ in $\mathcal{F}_b$, $\mathbf{M}$ the joint space inertia matrix, $\mathbf{I} \in \mathbb{R}^{6 \times 6}$ identity matrix and $\epsilon$ a regularisation term used when the Jacobian is near singularities.
For serial-parallel hybrid linkages, it is possible to consider the same equation~\eqref{eq:cartesian_inertia_matrix} using the projected versions of the Jacobian and inertia matrices (see Appendix~\ref{app:B}).
\begin{figure}[htb!]
    \centering
    \includegraphics[width=1.\columnwidth, trim={4cm 0cm 4cm 0cm}, clip=true]{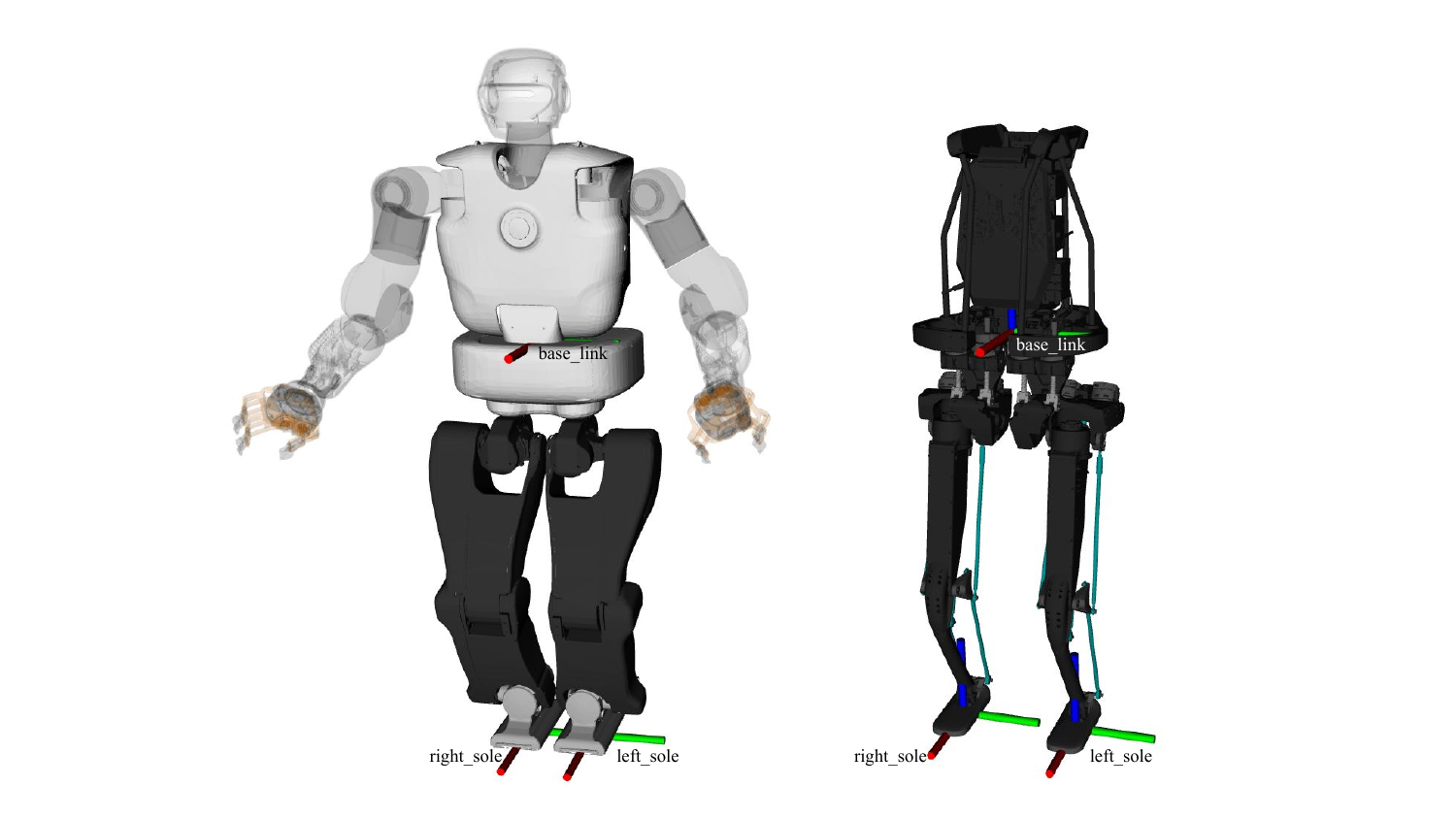}
    \caption{On the left is the TALOS robot. The parts in evidence are the ones considered for the comparison with the Kangaroo robot, on the right.}
    \label{fig:talos_vs_kangaroo}
\end{figure}
The equivalent Cartesian inertia matrix at the \emph{left\_sole} frame, situated at the sole of the left foot, is calculated with respect to the \emph{base\_link} in a standard nominal configuration typically adopted for walking.
In this configuration, the legs are nearly fully extended, as depicted in Fig.~\ref{fig:talos_vs_kangaroo}.
The comparison involves evaluating the norm of the columns of the matrix $\boldsymbol{\Lambda}$.
Let $\boldsymbol{\Lambda}_T$ and $\boldsymbol{\Lambda}_K$ represent the equivalent Cartesian inertia matrices of TALOS and Kangaroo, respectively, computed at the foot. 
We denote the $j$-th columns of these matrices as $\boldsymbol{\lambda}_{T,j}$ and $\boldsymbol{\lambda}_{K,j}$.
The \emph{index} $\chi_j$ is computed as:
\begin{equation}
    \chi_j = \frac{\|\boldsymbol{\lambda}_{T,j}\|}{\|\boldsymbol{\lambda}_{K,j}\|},
    \label{eq:chi_j}
\end{equation}
with $\chi_j > 1$ representing an improvement (reduced) equivalent Cartesian inertia along a certain direction. 
\par
Taking inspiration from the study conducted in~\cite{Ramos22}, we extend our comparison to encompass the \emph{Centroidal Angular Momentum Matrices} (CAMM) of Kangaroo and TALOS.
The CAMM can be computed from the \emph{System Momentum Matrix} (SMM), defined in~\cite{orin2013centroidal} as the product between the \emph{System Inertia} and \emph{System Jacobian}:
\begin{equation}
    \mathbf{A}_S = \mathbf{B}_S\mathbf{J}_S,
    \label{eq:smm}
\end{equation}
with $\mathbf{A}_S \in \mathbb{R}^{6N \times n + 6}$ the System Momentum, $N+1$ is the number of the links of the system, $\mathbf{B}_S \in \mathbb{R}^{6n \times 6n}$ the System Inertia matrix and $\mathbf{J}_S \in \mathbb{R}^{6n \times n + 6}$ the System Jacobian (see~\cite{orin2013centroidal} for details).
The SMM in~\eqref{eq:smm} can be projected at the \emph{centroid} of the system such that $\mathbf{A}_G = \mathbf{X}^T_G\mathbf{A}_S$,
with $\mathbf{X}_G \in \mathbb{R}^{6N \times 6}$ the stacking of adjoint matrices transforming motions from centroidal coordinates to link coordinates, and $\mathbf{A}_G \in \mathbb{R}^{6 \times n + 6}$ defined as the Centroidal Momentum Matrix (CMM), that maps joint velocities into centroidal momentum $\mathbf{h} = \mathbf{A}_G\boldsymbol{\nu} \in \mathbb{R}^6$, with the CAMM the angular part of the CMM.
%
As for the equivalent Cartesian Inertia matrix, it is possible to compute an equivalent CMM projected onto the closed linkage constraints.
\par
We define an index similar to~\eqref{eq:chi_j} computed at the same nominal configuration in Fig.~\ref{fig:talos_vs_kangaroo}:
\begin{equation}
    \gamma_j = \frac{\|\boldsymbol{\alpha}_{T,j}\|}{\|\boldsymbol{\alpha}_{K,j}\|},
\end{equation}
where $\boldsymbol{\alpha}_{\cdot,j}$ is the $j-th$ column of the CAMM with an improvement (if $> 1$) in the mass distribution.
\begin{table}[h]
\caption{Comparison among Kangaroo and Talos} 
\centering 
\begin{tabular}{l c r}
\hline 
$\left.\begin{matrix}
\boldsymbol{\Lambda}_{Linear}
\end{matrix}\right|_{\epsilon = 1e^{-5}}$
 & \makecell{ $\chi_x$ \\ $\chi_y$ \\ $\chi_z$} & \makecell{ 3.6 \\ 4.1 \\ 4.6}\\
\hline
$\left.\begin{matrix}
\boldsymbol{\Lambda}_{Angular}
\end{matrix}\right|_{\epsilon = 1e^{-5}}$ & \makecell{ $\chi_x$ \\ $\chi_y$ \\ $\chi_z$} & \makecell{ 11.4 \\ 4.9 \\ 3.6}\\
\hline
\hline
CAMM & \makecell{ $\gamma_x$ \\ $\gamma_y$ \\ $\gamma_z$} & \makecell{ 2.6 \\ 2.7 \\ 2.0}\\
\hline
\end{tabular}
\label{tab:cartesian_inertia}
\end{table}
\par
A succinct overview of the findings from this comparison is presented in Table~\ref{tab:cartesian_inertia}. 
The results highlight a substantial enhancement in the Kangaroo platform in comparison to Talos. 
This improvement pertains to both the equivalent Cartesian inertia observed at the feet and the Centroidal Angular Momentum Matrix (CAMM).

\section{Preliminary Experiments with Full-Model}\label{sec:experiments}
This section reports preliminary experiments on the real platform employing the full-model and applying algorithms derived from the constrained modeling presented in Section~\ref{sec:modeling}.
The Kangaroo full-model is constructed following the Unified Robotics Description Format (URDF) guidelines. 
However, the default URDF parser\footnote{\url{https://github.com/ros/urdfdom}} disallows joints with identical parent or child links, preventing the direct modeling of closed linkages. 
In contrast, the Simulation Description Format (SDF)\footnote{\url{http://sdformat.org/}} permits such joints, enabling the simulation of closed linkages via the Gazebo Simulator~\cite{koenig2004design}.
For this reason, we designed a configuration YAML file and a novel parser that permits the use of the URDF of Kangaroo with control libraries (e.g. the CartesI/O framework, by~\cite{Laurenzi19}) where all the necessary closed-linkage constraints are listed. 
Finally, we developed a C++ library for the kinematics and dynamics modeling of series-parallel hybrid chains named \emph{Closed Linkage Library} (CLL) based on RBDL~\cite{felis2017rbdl} and Eigen~\cite{eigenweb}.
The following experiments have been carried out using the CLL library on an Intel Core i7 CPU 2.30GHz × 16 cores with 32 GiB RAM, and are presented in the video accompanying this paper.

\subsection{Computation of Passive Kinematics Quantities and Initial Calibration}\label{sec:kinematic_estimation}
In general, when modeling serial-parallel hybrid linkages as constrained multi-body systems, it is necessary to have knowledge of the complete state vector denoted by $\boldsymbol{\theta}$ and $\boldsymbol{\dot{\theta}}$. 
However, this information may not be entirely accessible due to some passive quantities being either fully or partially unmeasurable. 
On the other hand, the actuated quantities typically remain observable.
Furthermore, the linear actuators in Kangaroo are based on ball-screws, equipped with absolute encoders at the motor side, permitting a relative measurement of the linear displacement of the screw, hence working as linear relative encoders.
For this reason, it is necessary to properly offset the initial value of the linear actuators to have a linear absolute reading, at every robot initialization. 
This calibration is made according to the arrangement of the closed kinematics and the measurements from the absolute encoders mounted on some of the passive DOFs.
\par
In this section, we formulate an algorithm to estimate the passive state, which is fundamental to reconstructing the full state $\boldsymbol{\theta}$ and $\boldsymbol{\dot{\theta}}$. 
This algorithm can serve as well as a systematic procedure for the initial calibration.
We consider having an initial estimation of the closed linkages DOF positions $\boldsymbol{\theta}_k$ and velocities $\boldsymbol{\dot{\theta}}_k$, and measured actuated positions $\boldsymbol{\bar{\theta}}_a$ and velocities $\boldsymbol{\bar{\dot{\theta}}}_a$, at instant $k$.
A new estimation for the passive velocities $\boldsymbol{\dot{\theta}}_{u, k+1}$ can be obtained using:
\begin{equation}
    \boldsymbol{\dot{\theta}}_{u, k+1} = \mathbf{J}_m\left(\boldsymbol{\theta}_k\right)\boldsymbol{\bar{\dot{\theta}}}_a,    
\end{equation}
that leads to the estimation of the new DOF velocities:
\begin{equation}
    \boldsymbol{\dot{\theta}}_{k+1} = \begin{bmatrix}
    \boldsymbol{\dot{\theta}}_{u, k+1} \\ \boldsymbol{\bar{\dot{\theta}}}_a
    \end{bmatrix}.
\end{equation}
\par
Considering equations~\eqref{eq:crank_slider_constraint} and~\eqref{eq:crank_slider_constraint_differential}, is possible to consider as well a properly computed error ${\mathbf{e}_l(\boldsymbol{\theta})} \in \mathbb{R}^m$ associated to the closed linkage: 
\begin{equation}
    {\mathbf{J}_l}\left(\boldsymbol{\theta}\right)\boldsymbol{\dot{\theta}} = \alpha{\mathbf{e}_l(\boldsymbol{\theta})},
    \label{eq:crank_slider_constraint_differential_with_error}
\end{equation}
which define the error dynamics $\mathbf{\dot{e}}_l + \alpha{\mathbf{e}_l(\boldsymbol{\theta})} = \mathbf{0}$ converging to zero exponentially.
%
Using~\eqref{eq:separate_jacobians_crank_slider} in~\eqref{eq:crank_slider_constraint_differential_with_error} leads to the estimator:
\begin{equation}
    \boldsymbol{\dot{\theta}}_{u,k+1}^* = \mathbf{J}_m\left(\boldsymbol{\theta}_k\right)\boldsymbol{\dot{\theta}}_{a,k} + \alpha{\mathbf{J}_{l,u}}\left(\boldsymbol{\theta}_k\right)^{-1}{\mathbf{e}_l(\boldsymbol{\theta}_k)},
    \label{eq:passive_estimation}
\end{equation}
with $\boldsymbol{\dot{\theta}}_{a,k} = \boldsymbol{\bar{\dot{\theta}}}_a + \beta\left(\boldsymbol{\bar{\theta}}_a - \boldsymbol{\theta}_{a, k}\right)$, and $\alpha$ and $\beta$ two scalars, positive, tunable gains.
\par
A new estimation for the positions $\boldsymbol{\theta}_{k+1}$ can be obtained using one-step Euler integration:
\begin{equation}
    \boldsymbol{\theta}_{k+1} = \boldsymbol{\theta}_{k} + dt \boldsymbol{\dot{\theta}}_{k+1}^*,
    \label{eq:position_estimation}
\end{equation}
given the time step $dt$ and 
\begin{equation}
    \boldsymbol{\dot{\theta}}_{k+1}^* = \begin{bmatrix}
        \boldsymbol{\dot{\theta}}_{u,k+1}^*\\
        \boldsymbol{\dot{\theta}}_{a,k}
    \end{bmatrix}.
\end{equation}
\par
The two terms in~\eqref{eq:passive_estimation} consist of a correction for the measurement and a correction for the constraint error. 
Both these terms are used to mitigate the drift phenomena induced by the discrete integration in~\eqref{eq:position_estimation}, therefore the values of $\alpha$ and $\beta$ are bounded by the rate of the estimation loop.
\par
\begin{figure}[htb!]
    \centering
    \includegraphics[width=1.\columnwidth, trim={1.5cm 0cm 1cm 0cm}, clip=true]{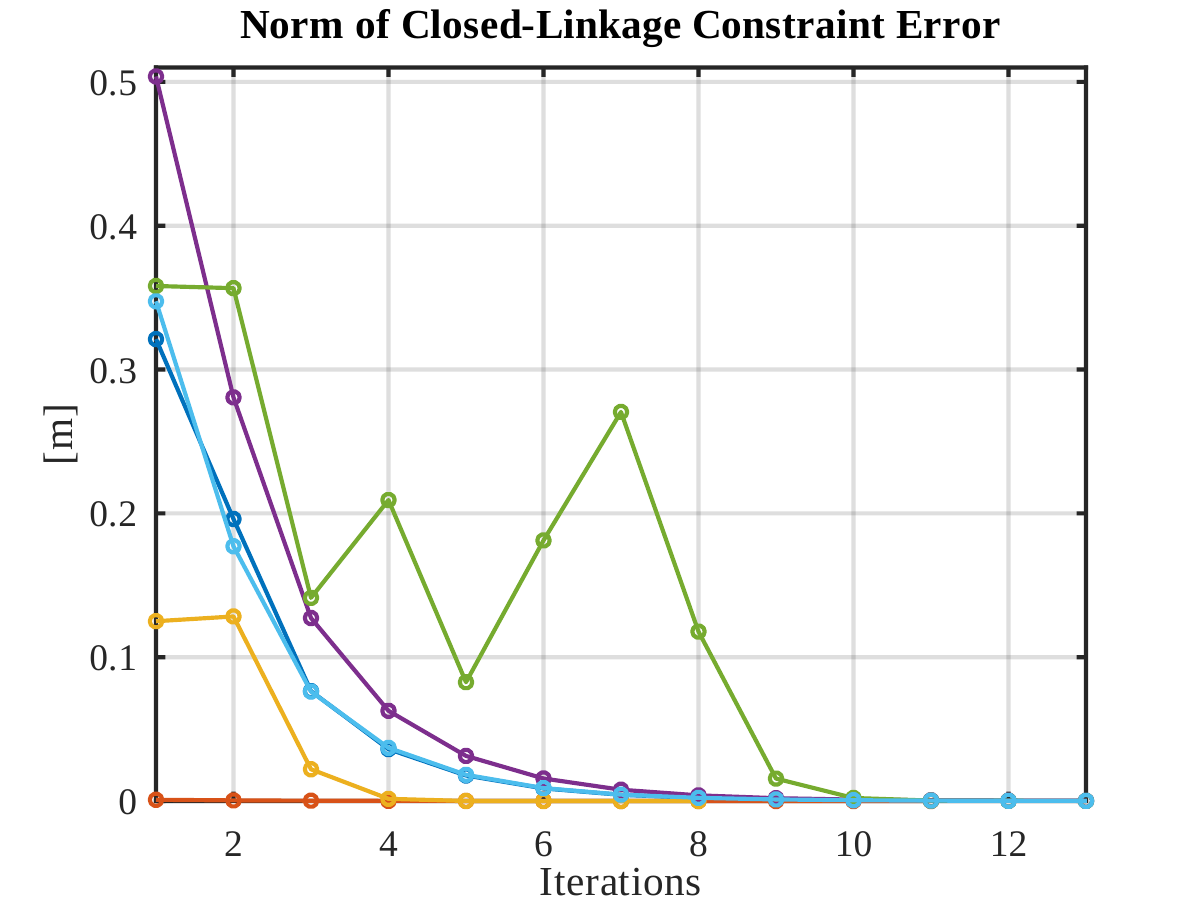}
    \caption{Calibration procedure validation. Notice that the constraint error contains both linear and angular parts.}
    \label{fig:calibration_ik_exp}
\end{figure}

%
%
%
It is possible to define a simple calibration procedure based on the previous algorithm.
We consider augmenting the constraint Jacobian with the measurement matrix $\mathbf{E}$:
\begin{equation}
\mathbf{J}\left(\boldsymbol{\theta}\right) = \begin{bmatrix}
{\mathbf{J}_l}\left(\boldsymbol{\theta}\right)\\ 
\mathbf{E}
\end{bmatrix},
\end{equation}
and the constraint error with the measurement error $\mathbf{e}_m$:
\begin{equation}
\mathbf{e}\left(\boldsymbol{\theta}\right) = \begin{bmatrix}
{\mathbf{e}_l(\boldsymbol{\theta})}\\ 
\mathbf{e}_m
\end{bmatrix}.
\end{equation}
In the case of non-redundant measurements, as for the Kangaroo robot, we will have $\mathbf{J}\left(\boldsymbol{\theta}\right) \in \mathbb{R}^{n \times n}$ with $\mathbf{E} \in \mathbb{R}^{l \times n}$, such as $l + m = n$, and $\mathbf{e}\left(\boldsymbol{\theta}\right) \in \mathbb{R}^n$.
Each row in the measurement matrix $\mathbf{E}$ is a zero row with a single $``1"$ placed at the measured (passive) DOF.
%
The measured error consists of the error associated with each measurement $e_i = \theta_{j,m} - \theta_j$, with $\mathbf{e}_m = \left[e_0, e_1, \dots, e_l \right]^T$.
Finally, it is worth noticing that the reference for the closed linkage error ${\mathbf{e}_l(\boldsymbol{\theta})}$ is computed at a configuration where all the open linkages are properly closed.
The calibration procedure is based on the iterative resolution of the linear system $\mathbf{J}\left(\boldsymbol{\theta}\right)\boldsymbol{\dot{\theta}} = \alpha\mathbf{e}\left(\boldsymbol{\theta}\right)$, equivalent to~\eqref{eq:crank_slider_constraint_differential_with_error}, until the norm of the error is above a certain threshold $\epsilon$: $\left \| \mathbf{e}\left(\boldsymbol{\theta}\right) \right \| < \epsilon$.
%
%
In Fig.~\ref{fig:calibration_ik_exp} is reported the norm of the closed-linkage error, for 6 different initial measurements. 
It is possible to see that the initialization procedure converges below $1e^{-4}$ global error norm in a few iterations.
%
\par
\begin{figure}[htb!]
    \centering
    \includegraphics[width=1\columnwidth, trim={0cm 0cm 1cm 0cm}, clip=true]{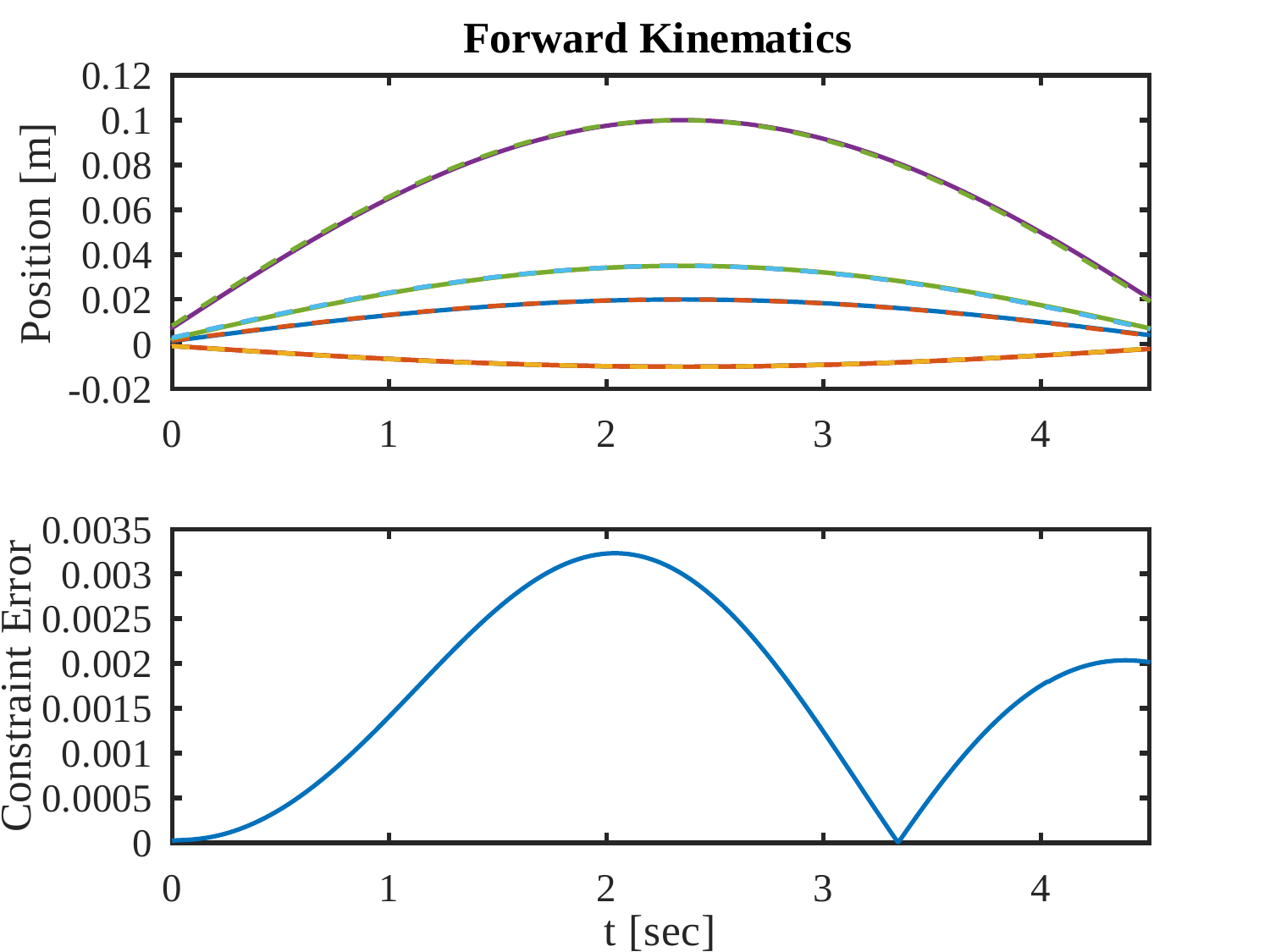}
    \caption{The top plot reports the tracking of measured (dashed) versus estimated (continuous) position and velocity, respectively, of the actuated DOFs. In the bottom plot the norm of the closed linkage constraint error (linear and angular).}
    \label{fig:fk_exp}
\end{figure}
\begin{figure}[htb!]
    \centering
    \includegraphics[width=1.\columnwidth, trim={2cm 0.5cm 2cm 0.5cm}, clip=true]{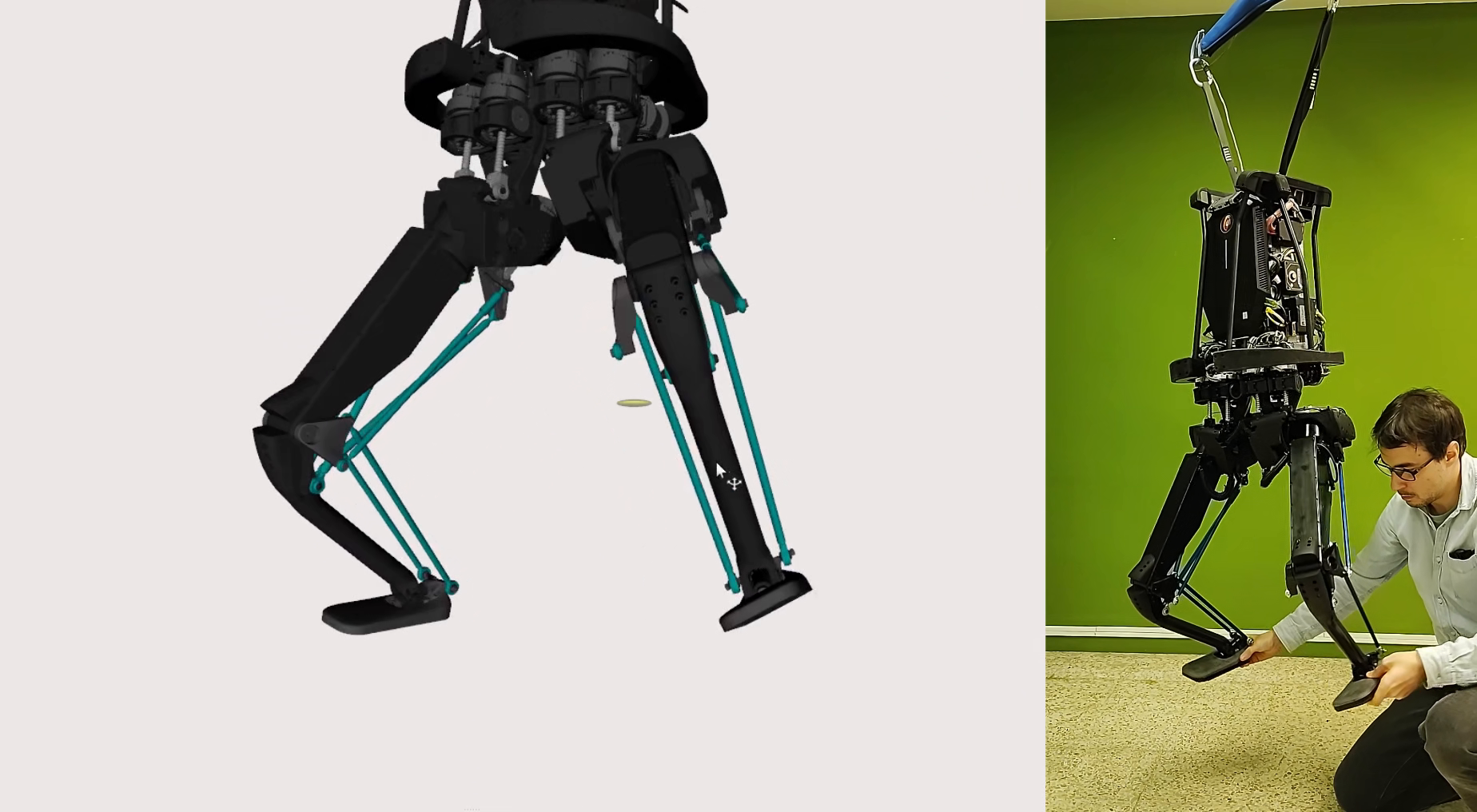}
    \caption{The kinematic estimation running on the real Kangaroo platform.}
    \label{fig:kangaroo_fk_real}
\end{figure}

Concerning the computation of passive kinematics quantities, it is worth noticing that it can be seen as a tracking problem of the measured actuated positions, constrained by the closed linkages. 
The validation is therefore performed using sinusoidal position and velocity reference (measured) trajectories for the actuated DOFs, at $1 \ [kHz]$, that is tracked by the estimator algorithm, running as well at $1 \ [kHz]$, with parameters $\alpha = 1$ and $\beta = 1000$.
In Fig.~\ref{fig:fk_exp} are reported the position and velocity \emph{tracking} between the estimated and measured actuated quantities, and the norm of the error of the closed linkage constraint.   
The average computation time for the algorithm, together with the update of the full-model, and computation of inverse dynamics torques using equations~\eqref{eq:whole_body_id_lambda} and~\eqref{eq:whole_body_id_joint} without contact forces, is $\approx0.38 \ [ms]$, using the \texttt{partialPivLu}\footnote{\url{https://eigen.tuxfamily.org/dox/classEigen_1_1PartialPivLU.html}} method from the Eigen software library, and dense matrices representation.
The motion performed in this numerical validation, together with results achieved on the real Kangaroo prototype is reported in the accompanying video.  
Fig.~\ref{fig:kangaroo_fk_real} shows the kinematic estimation algorithm running on the real robot.

\subsection{Quasi-Static Contact Wrench Estimation} \label{sec:contact_force_estimation}
As of the current writing of this paper, the Kangaroo robot prototype does not incorporate any direct force/torque sensor measurements at its feet.
By not housing electronics from the feet to the thigh, the robot becomes less vulnerable to impacts.
However, in light of this design approach, there arises a need to estimate contact wrenches based on the force measurements of the actuators, all while considering the serial-parallel hybrid kinematics.
\par
Assuming $\boldsymbol{\nu} = \mathbf{0}$ and $\boldsymbol{\dot{\nu}} = \mathbf{0}$, equations~\eqref{eq:whole_body_id_lambda} and~\eqref{eq:whole_body_id_joint} become respectively:
\begin{subequations}
\begin{align}
    \boldsymbol{\lambda} &= \mathbf{J}_{l,u}(\mathbf{q})^{-T}\left( \mathbf{g}_u(\mathbf{q}) -\mathbf{J}_{c,u}(\mathbf{q})^T\mathbf{F} \right), 
    \label{eq:whole_body_statics_lambda} \\
    \boldsymbol{\tau} &= \mathbf{g}_a(\mathbf{q}) - \mathbf{J}_{c,a}(\mathbf{q})^T\mathbf{F} - \mathbf{J}_{l,a}(\mathbf{q})^T\boldsymbol{\lambda}.
    \label{eq:whole_body_statics_joints}
    \end{align}
\end{subequations}
%
Substituting~\eqref{eq:whole_body_statics_lambda} in~\eqref{eq:whole_body_statics_joints}, leads to:
\begin{equation}
    \boldsymbol{\tau} = \mathbf{g}_a(\mathbf{q}) - \mathbf{J}_{c,a}(\mathbf{q})^T\mathbf{F} + \mathbf{J}_m\left(\mathbf{q}\right)^T\left( \mathbf{g}_u(\mathbf{q}) -\mathbf{J}_{c,u}(\mathbf{q})^T\mathbf{F} \right),
\end{equation}
where we have used~\eqref{eq:mapping_jacobian_transpose}.
Reorganizing leads to:
\begin{subequations}
\label{eq:force_estimation_1}
\begin{align}
    \boldsymbol{\tau} - \mathbf{g}_a(\mathbf{q}) - \mathbf{J}_m\left(\mathbf{q}\right)^T\mathbf{g}_u(\mathbf{q}) =  \nonumber\\
    =- \mathbf{J}_{c,a}(\mathbf{q})^T\mathbf{F} - \mathbf{J}_m\left(\mathbf{q}\right)^T\mathbf{J}_{c,u}(\mathbf{q})^T\mathbf{F} = \nonumber\\
    =-\left(\mathbf{J}_{c,a}(\mathbf{q})^T + \mathbf{J}_m\left(\mathbf{q}\right)^T\mathbf{J}_{c,u}(\mathbf{q})^T\right)\mathbf{F} = \nonumber\\
    =\mathbf{J}_{c,m}(\mathbf{q})^T\mathbf{F}, \tag{\ref{eq:force_estimation_1}}
\end{align}
\end{subequations}
that can be used to statically estimate contact forces given measured force/torques $\boldsymbol{\bar{\tau}}$:
\begin{equation}
    \left(\mathbf{J}_{c,m}(\mathbf{q})^T\right)^{\dagger}\left( \boldsymbol{\bar{\tau}} - \mathbf{g}_a(\mathbf{q}) - \mathbf{J}_m\left(\mathbf{q}\right)^T\mathbf{g}_u(\mathbf{q}) \right) = \mathbf{F},
\end{equation}
with $\mathbf{J}_{c,m}(\mathbf{q}) \in \mathbb{R}^{3c \times a}$ and $(\cdot)^{\dagger}$ a properly computed pseudo-inverse.
Notice that velocity measurements could be included to consider the effect of Coriolis/centrifugal terms.
%
%
From the wrench estimation, we compute the Zero Moment Point (ZMP) being a fundamental quantity used in bipedal locomotion for stabilization, see~\cite{Alex12, kajita2014introduction}.
%
\par
\begin{figure}[hbt!]
    \centering
    \includegraphics[width=1.\columnwidth, trim={7cm 0cm 7cm 0cm}, clip=true]{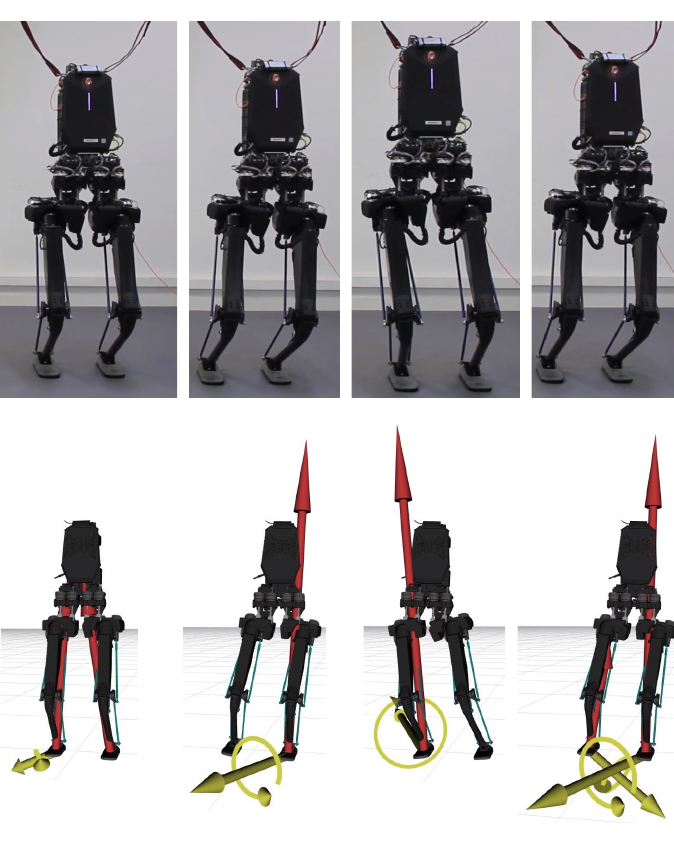}
    \caption{In the top picture, the Kangaroo robot demonstrates lateral swing motions. In the bottom picture, RVIZ visualizes the estimated full configuration and contact wrenches.}
    \label{fig:kangaroo_swing}
\end{figure}

We conclude by showcasing an experiment involving the robot executing lateral swinging motions (see Fig.~\ref{fig:kangaroo_swing}).
In this experiment, although the motion of each sub-mechanism is individually controlled through geometric calculations, we utilize the full model for estimating the contact wrenches. 
%
%
%
The estimation of the axial force $\bar{\tau}$ in each linear actuator is computed from the measured motor torque $\tau_m$ as:
\begin{equation}
    \bar{\tau} = \frac{\tau_m \cdot 2\pi \cdot \eta}{L},
\end{equation}
with $\eta = 0.95$ the efficiency of the ball screw provided by the constructor, and $L$ the screw lead, that for the leg length actuator is $0.1 \ [m]$ and for the hip and ankle actuators is $0.05 \ [m]$.
\begin{figure}
     \centering
     \begin{subfigure}[b]{1.\columnwidth}
         \centering
         \includegraphics[width=1.\columnwidth, trim={0cm 0cm 1cm 0cm}, clip=true]{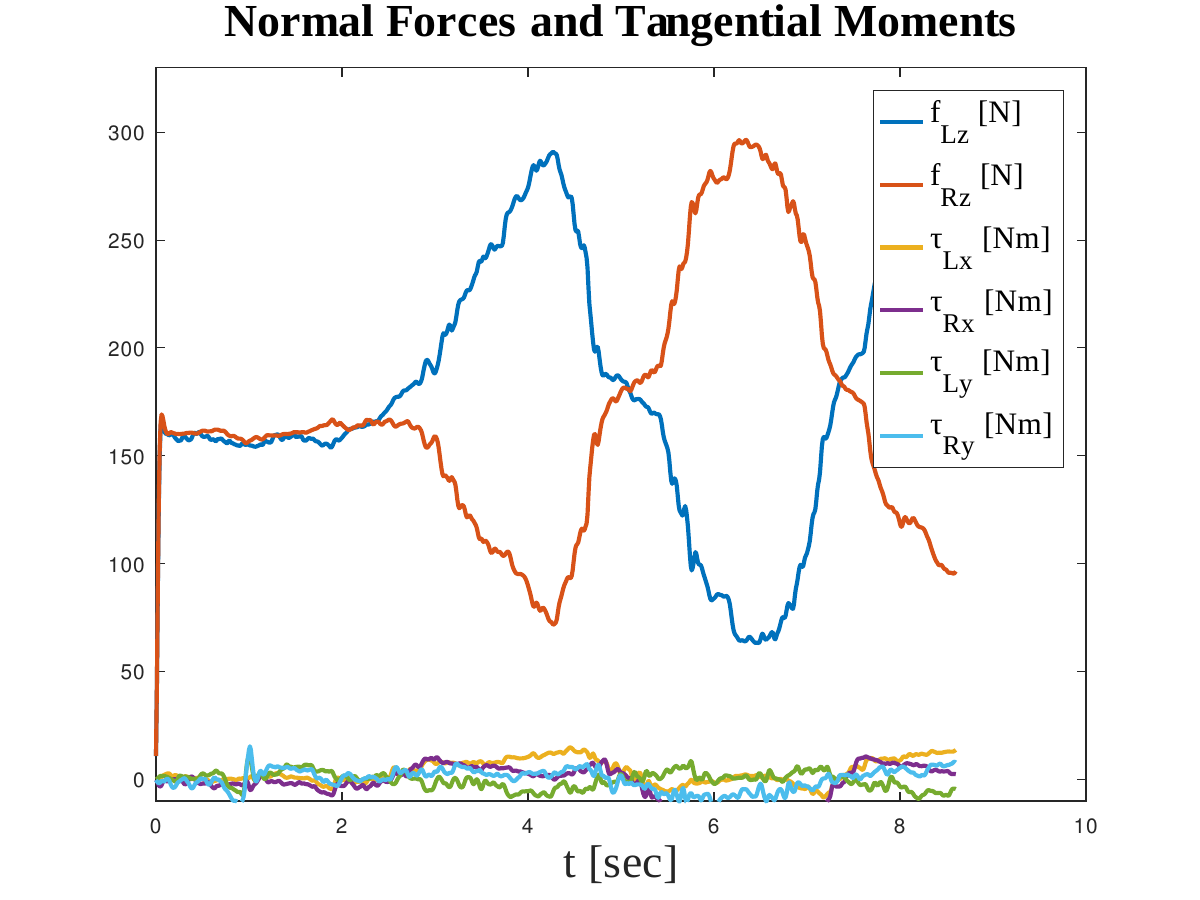}
         \caption{}
         \label{fig:real_contact_wrenches_fb}
     \end{subfigure}
     \\
     \begin{subfigure}[b]{1.\columnwidth}
         \centering
         \includegraphics[width=1.\columnwidth, trim={0cm 0cm 1cm 0cm}, clip=true]{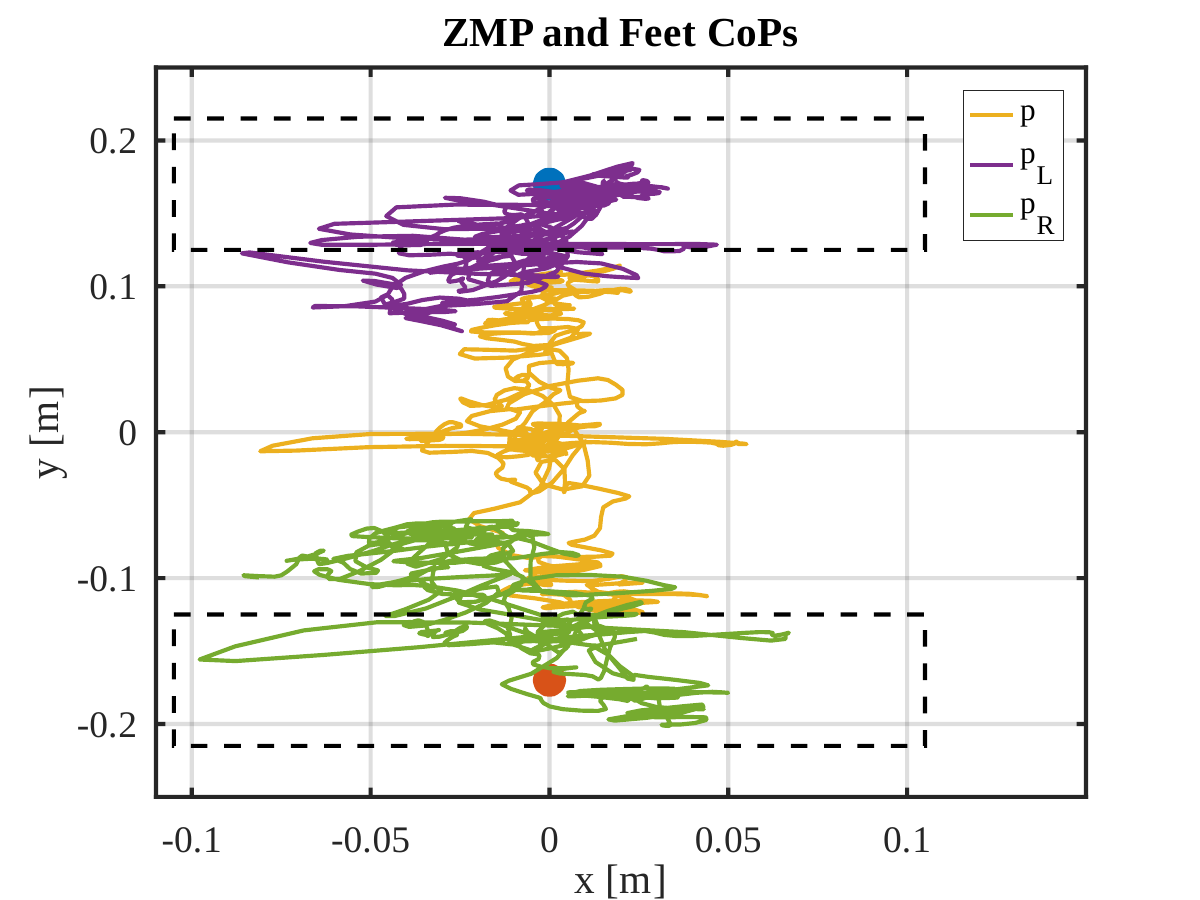}
         \caption{}
         \label{fig:real_zmp}
     \end{subfigure}
     \caption{Results of the estimated contact wrenches using data from lateral swing experiment with the Kangaroo hardware.}
\end{figure}
Fig.~\ref{fig:real_contact_wrenches_fb} reports the reconstructed contact wrench filtered using a low-pass second-order Butterworth filter with cuttoff frequency $5 \ [Hz]$. 
It is possible to see that the estimated tangential forces correctly sum up at the weight of the robot without the cage, approximately $36 \ [kg]$. 
In Fig.~\ref{fig:real_zmp} are reported the computed ZMP and feet Centers of Pressure (COPs) using the estimated contact wrenches.  
The contact wrench algorithms estimation time is $\approx0.03 \ [ms]$, using the \texttt{fullPivLu}\footnote{\url{https://eigen.tuxfamily.org/dox/classEigen_1_1FullPivLU.html}} method from the Eigen library, and dense matrices representation, allowing for fast real-time implementation.

\section{Conclusion and Future Works}\label{sec:conclusion}
In this paper, we presented the kinematic and dynamic modeling, and study of the lower body of Kangaroo, a novel humanoid bipedal robot designed and manufactured by PAL Robotics. 
Distinguished by its incorporation of series-parallel hybrid chains, Kangaroo's unique design prompted us to delve into constraint-based modeling techniques for serial-parallel hybrid systems
We conducted an in-depth analysis of Kangaroo's kinematics, such as the non-linear transmission mechanisms within the knee and differential parallel components governing hip and ankle movements.
Furthermore, we computed the equivalent Cartesian inertia at the end-effector and the Centroidal Momentum matrix in the context of serial-parallel hybrid chains. 
This involved a comparative analysis with the Talos robot, a preceding humanoid creation from PAL Robotics. 
Our examination underscored the advancements achieved in Kangaroo's design.
To facilitate our research, we introduced a suite of software tools designed to model, analyze, and control both fixed and floating-base robots housing serial-parallel hybrid chains as constrained multi-body systems. 
An example of our software contributions is the \emph{Closed Linkage Library}, that facilitated the analyses and preliminary real-world experiments undertaken with the Kangaroo platform.
\par
Future works will regard the deployment of Whole-Body Inverse Dynamics on the Kangaroo platform together with model predictive control algorithms to perform highly agile and dynamic motions, such as running and jumping. 
In particular, we envision the employment of a simplified, serial model of the robot for the model predictive control, i.e. considering only significative passive DOFs, and mapping the control input onto the full model using the IK and ID presented in this paper. 
We are also investigating the extension of passivity-based approaches, see~\cite{cic}, for Operational Space Control to series-parallel hybrid chains.
Finally, the tools developed in this work will be the basis for the design and development of the upper body for the Kangaroo robot, to evaluate different kinematic structures for the torso and the arms.



\bibliographystyle{IEEEtran}	
\bibliography{biblio}

\begin{thebibliography}{10}
\providecommand{\url}[1]{#1}
\csname url@rmstyle\endcsname
\providecommand{\newblock}{\relax}
\providecommand{\bibinfo}[2]{#2}
\providecommand\BIBentrySTDinterwordspacing{\spaceskip=0pt\relax}
\providecommand\BIBentryALTinterwordstretchfactor{4}
\providecommand\BIBentryALTinterwordspacing{\spaceskip=\fontdimen2\font plus
\BIBentryALTinterwordstretchfactor\fontdimen3\font minus
  \fontdimen4\font\relax}
\providecommand\BIBforeignlanguage[2]{{%
\expandafter\ifx\csname l@#1\endcsname\relax
\typeout{** WARNING: IEEEtran.bst: No hyphenation pattern has been}%
\typeout{** loaded for the language `#1'. Using the pattern for}%
\typeout{** the default language instead.}%
\else
\language=\csname l@#1\endcsname
\fi
#2}}

\bibitem{kaneko2002design}
K.~Kaneko, F.~Kanehiro, S.~Kajita, K.~Yokoyama, K.~Akachi, T.~Kawasaki, S.~Ota,
  and T.~Isozumi, ``Design of prototype humanoid robotics platform for hrp,''
  in \emph{IEEE/RSJ International Conference on Intelligent Robots and
  Systems}, vol.~3, 2002, pp. 2431--2436.

\bibitem{kaneko2008humanoid}
K.~Kaneko, K.~Harada, F.~Kanehiro, G.~Miyamori, and K.~Akachi, ``Humanoid robot
  hrp-3,'' in \emph{IEEE/RSJ International Conference on Intelligent Robots and
  Systems}, 2008, pp. 2471--2478.

\bibitem{hirose2007honda}
M.~Hirose and K.~Ogawa, ``Honda humanoid robots development,''
  \emph{Philosophical Transactions of the Royal Society A: Mathematical,
  Physical and Engineering Sciences}, vol. 365, no. 1850, pp. 11--19, 2007.

\bibitem{tellez2008reem}
R.~Tellez, F.~Ferro, S.~Garcia, E.~Gomez, E.~Jorge, D.~Mora, D.~Pinyol,
  J.~Oliver, O.~Torres, J.~Velazquez, \emph{et~al.}, ``Reem-b: An autonomous
  lightweight human-size humanoid robot,'' in \emph{IEEE/RAS International
  Conference on Humanoid Robots}, 2008, pp. 462--468.

\bibitem{park2007mechanical}
I.-W. Park, J.-Y. Kim, J.~Lee, and J.-H. Oh, ``Mechanical design of the
  humanoid robot platform, hubo,'' \emph{Advanced Robotics}, vol.~21, no.~11,
  pp. 1305--1322, 2007.

\bibitem{lohmeier2009humanoid}
S.~Lohmeier, T.~Buschmann, and H.~Ulbrich, ``Humanoid robot lola,'' in
  \emph{IEEE International Conference on Robotics and Automation}, 2009, pp.
  775--780.

\bibitem{hobbelen2008}
D.~Hobbelen, T.~De~Boer, and M.~Wisse, ``System overview of bipedal robots
  flame and tulip: Tailor-made for limit cycle walking,'' in \emph{IEEE/RSJ
  international conference on intelligent robots and systems}, 2008, pp.
  2486--2491.

\bibitem{spenko2018darpa}
M.~Spenko, S.~Buerger, and K.~Iagnemma, \emph{The DARPA robotics challenge
  finals: humanoid robots to the rescue}.\hskip 1em plus 0.5em minus
  0.4em\relax Springer, 2018, vol. 121.

\bibitem{Tsagarakis:2017}
N.~G. Tsagarakis, D.~G. Caldwell, F.~Negrello, W.~Choi, L.~Baccelliere, V.~G.
  Loc, J.~H. Noorden, L.~Muratore, A.~Margan, A.~Cardellino, L.~Natale,
  E.~Mingo~Hoffman, H.~Dallali, N.~Kashiri, J.~Malzahn, J.~Lee, P.~Kryczka,
  D.~Kanoulas, M.~Garabini, M.~G. Catalano, M.~Ferrati, V.~Varricchio,
  L.~Pallottino, C.~Pavan, A.~Bicchi, A.~Settimi, A.~Rocchi, and A.~Arash,
  ``Walk-man: A high-performance humanoid platform for realistic
  environments,'' \emph{Journal of Field Robotics}, vol.~34, no.~7, pp.
  1225--1259, 2017.

\bibitem{radford2015valkyrie}
N.~A. Radford, P.~Strawser, K.~Hambuchen, J.~S. Mehling, W.~K. Verdeyen, A.~S.
  Donnan, J.~Holley, J.~Sanchez, V.~Nguyen, L.~Bridgwater, \emph{et~al.},
  ``Valkyrie: Nasa's first bipedal humanoid robot,'' \emph{Journal of Field
  Robotics}, vol.~32, no.~3, pp. 397--419, 2015.

\bibitem{kojima2015development}
K.~Kojima, T.~Karasawa, T.~Kozuki, E.~Kuroiwa, S.~Yukizaki, S.~Iwaishi,
  T.~Ishikawa, R.~Koyama, S.~Noda, F.~Sugai, \emph{et~al.}, ``Development of
  life-sized high-power humanoid robot jaxon for real-world use,'' in
  \emph{IEEE/RAS International Conference on Humanoid Robots}, 2015, pp.
  838--843.

\bibitem{stasse2017talos}
O.~Stasse, T.~Flayols, R.~Budhiraja, K.~Giraud-Esclasse, J.~Carpentier,
  J.~Mirabel, A.~Del~Prete, P.~Souères, N.~Mansard, F.~Lamiraux, J.-P.
  Laumond, L.~Marchionni, H.~Tome, and F.~Ferro, ``Talos: A new humanoid
  research platform targeted for industrial applications,'' in \emph{IEEE/RAS
  International Conference on Humanoid Robots}, 2017, pp. 689--695.

\bibitem{guizzo2019leaps}
E.~Guizzo, ``By leaps and bounds: An exclusive look at how boston dynamics is
  redefining robot agility,'' \emph{IEEE Spectrum}, vol.~56, no.~12, pp.
  34--39, 2019.

\bibitem{hurst2019walk}
J.~Hurst, ``Walk this way: To be useful around people, robots need to learn how
  to move like we do,'' \emph{IEEE Spectrum}, vol.~56, no.~3, pp. 30--51, 2019.

\bibitem{zhu2023design}
T.~Zhu, \emph{Design of a Highly Dynamic Humanoid Robot}.\hskip 1em plus 0.5em
  minus 0.4em\relax University of California, Los Angeles, 2023.

\bibitem{kumar2020survey}
S.~Kumar, H.~W{\"o}hrle, J.~de~Gea~Fern{\'a}ndez, A.~M{\"u}ller, and
  F.~Kirchner, ``A survey on modularity and distributivity in series-parallel
  hybrid robots,'' \emph{Mechatronics}, vol.~68, p. 102367, 2020.

\bibitem{stasse2019overview}
O.~Stasse and T.~Flayols, ``An overview of humanoid robots technologies,''
  \emph{Biomechanics of Anthropomorphic Systems}, pp. 281--310, 2019.

\bibitem{Ruscelli18}
F.~Ruscelli, A.~Laurenzi, E.~Mingo~Hoffman, and N.~G. Tsagarakis, ``A fail-safe
  semi-centralized impedance controller: Validation on a parallel kinematics
  ankle,'' in \emph{IEEE/RSJ International Conference on Intelligent Robots and
  Systems}, 2018, pp. 1--9.

\bibitem{Kaminaga16}
H.~Kaminaga, T.~Ko, R.~Masumura, M.~Komagata, S.~Sato, S.~Yorita, and
  Y.~Nakamura, ``Mechanism and control of whole-body electro-hydrostatic
  actuator driven humanoid robot hydra,'' in \emph{International Symposium on
  Experimental Robotics}.\hskip 1em plus 0.5em minus 0.4em\relax Springer
  International Publishing, 2017, pp. 656--665.

\bibitem{lahr2016biologically}
D.~F. Lahr, H.~Yi, and D.~W. Hong, ``Biologically inspired design of a parallel
  actuated humanoid robot,'' \emph{Advanced Robotics}, vol.~30, no.~2, pp.
  109--118, 2016.

\bibitem{schutz2017carl}
S.~Sch{\"u}tz, A.~Nejadfard, K.~Mianowski, P.~Vonwirth, and K.~Berns,
  ``Carl—a compliant robotic leg featuring mono-and biarticular actuation,''
  in \emph{IEEE-RAS International Conference on Humanoid Robots}, 2017, pp.
  289--296.

\bibitem{gim2018design}
K.~G. Gim, J.~Kim, and K.~Yamane, ``Design and fabrication of a bipedal robot
  using serial-parallel hybrid leg mechanism,'' in \emph{IEEE/RSJ International
  Conference on Intelligent Robots and Systems}, 2018, pp. 5095--5100.

\bibitem{chitta2017ros_control}
S.~Chitta, E.~Marder-Eppstein, W.~Meeussen, V.~Pradeep, A.~R. Tsouroukdissian,
  J.~Bohren, D.~Coleman, B.~Magyar, G.~Raiola, M.~L{\"u}dtke, \emph{et~al.},
  ``ros\_control: A generic and simple control framework for ros,'' \emph{The
  Journal of Open Source Software}, vol.~2, no.~20, pp. 456--456, 2017.

\bibitem{mronga2021whole}
D.~Mronga, S.~Kumar, and F.~Kirchner, ``Whole-body control of series-parallel
  hybrid robots,'' in \emph{International Conference on Robotics and
  Automation}, 2022, pp. 228--234.

\bibitem{Kumar20}
S.~Kumar, K.~A.~v. Szadkowski, A.~Mueller, and F.~Kirchner, ``An analytical and
  modular software workbench for solving kinematics and dynamics of
  series-parallel hybrid robots,'' \emph{Journal of Mechanisms and Robotics},
  vol.~12, no.~2, 02 2020.

\bibitem{ebetaer2021design}
J.~E$\beta$er, S.~Kumar, H.~Peters, V.~Bargsten, J.~de~Gea~Fernandez,
  C.~Mastalli, O.~Stasse, and F.~Kirchner, ``Design, analysis and control of
  the series-parallel hybrid rh5 humanoid robot,'' in \emph{IEEE/RAS
  International Conference on Humanoid Robots}, 2021, pp. 400--407.

\bibitem{grimes2012design}
J.~A. Grimes and J.~W. Hurst, ``The design of atrias 1.0 a unique monopod,
  hopping robot,'' in \emph{Adaptive Mobile Robotics}.\hskip 1em plus 0.5em
  minus 0.4em\relax World Scientific, 2012, pp. 548--554.

\bibitem{reher2019dynamic}
J.~Reher, W.-L. Ma, and A.~D. Ames, ``Dynamic walking with compliance on a
  cassie bipedal robot,'' in \emph{IEEE European Control Conference}, 2019, pp.
  2589--2595.

\bibitem{apgar2018fast}
T.~Apgar, P.~Clary, K.~Green, A.~Fern, and J.~W. Hurst, ``Fast online
  trajectory optimization for the bipedal robot cassie.'' in \emph{Robotics:
  Science and Systems}, vol. 101, 2018, p.~14.

\bibitem{schumacher2021versatile}
C.~Schumacher, E.~Knoop, and M.~B{\"a}cher, ``A versatile inverse kinematics
  formulation for retargeting motions onto robots with kinematic loops,''
  \emph{IEEE Robotics and Automation Letters}, vol.~6, no.~2, pp. 943--950,
  2021.

\bibitem{lemburg2011aila}
J.~Lemburg, J.~de~Gea~Fern{\'a}ndez, M.~Eich, D.~Mronga, P.~Kampmann, A.~Vogt,
  A.~Aggarwal, Y.~Shi, and F.~Kirchner, ``Aila-design of an autonomous mobile
  dual-arm robot,'' in \emph{IEEE International Conference on Robotics and
  Automation}, 2011, pp. 5147--5153.

\bibitem{englsberger2014overview}
J.~Englsberger, A.~Werner, C.~Ott, B.~Henze, M.~A. Roa, G.~Garofalo, R.~Burger,
  A.~Beyer, O.~Eiberger, K.~Schmid, \emph{et~al.}, ``Overview of the
  torque-controlled humanoid robot toro,'' in \emph{IEEE-RAS International
  Conference on Humanoid Robots}, 2014, pp. 916--923.

\bibitem{lee2014thesis}
B.~K. T.-S. Lee, ``Design of a humanoid robot for disaster response,'' Ph.D.
  dissertation, Virginia Tech, 2014.

\bibitem{lahr2014thesis}
D.~F. Lahr, ``Design and control of a humanoid robot, saffir,'' Ph.D.
  dissertation, Virginia Tech, 2014.

\bibitem{cafolla2016larmbot}
D.~Cafolla, M.~Wang, G.~Carbone, and M.~Ceccarelli, ``Larmbot: a new humanoid
  robot with parallel mechanisms,'' in \emph{Symposium on Robot Design,
  Dynamics and Control}.\hskip 1em plus 0.5em minus 0.4em\relax Springer, 2016,
  pp. 275--283.

\bibitem{kumar2019kinematic}
S.~Kumar, A.~Nayak, H.~Peters, C.~Schulz, A.~M{\"u}ller, and F.~Kirchner,
  ``Kinematic analysis of a novel parallel 2sprr+1u ankle mechanism in humanoid
  robot,'' in \emph{Advances in Robot Kinematics 2018 16}.\hskip 1em plus 0.5em
  minus 0.4em\relax Springer, 2019, pp. 431--439.

\bibitem{laulusa2008review}
A.~Laulusa and O.~A. Bauchau, ``Review of classical approaches for constraint
  enforcement in multibody systems,'' \emph{Journal of computational and
  nonlinear dynamics}, vol.~3, no.~1, 2008.

\bibitem{carpentier_hal-03271811}
J.~Carpentier, R.~Budhiraja, and N.~Mansard, ``{Proximal and Sparse Resolution
  of Constrained Dynamic Equations},'' in \emph{{Robotics: Science and
  Systems}}, July 2021.

\bibitem{mansard2012dedicated}
N.~Mansard, ``A dedicated solver for fast operational-space inverse dynamics,''
  in \emph{IEEE International Conference on Robotics and Automation}.\hskip 1em
  plus 0.5em minus 0.4em\relax IEEE, 2012, pp. 4943--4949.

\bibitem{DelPrete18}
A.~D. Prete, ``Joint position and velocity bounds in discrete-time
  acceleration/torque control of robot manipulators,'' \emph{IEEE Robotics and
  Automation Letters}, vol.~3, no.~1, pp. 281--288, 2018.

\bibitem{Khazoom22}
C.~Khazoom, D.~Gonzalez-Diaz, Y.~Ding, and S.~Kim, ``Humanoid self-collision
  avoidance using whole-body control with control barrier functions,'' in
  \emph{IEEE-RAS International Conference on Humanoid Robots}, 2022.

\bibitem{brockett2016biomechanics}
C.~L. Brockett and G.~J. Chapman, ``Biomechanics of the ankle,''
  \emph{Orthopaedics and trauma}, vol.~30, no.~3, pp. 232--238, 2016.

\bibitem{lee2014design}
B.~Lee, C.~Knabe, V.~Orekhov, and D.~Hong, ``Design of a human-like range of
  motion hip joint for humanoid robots,'' in \emph{International Design
  Engineering Technical Conferences and Computers and Information in
  Engineering Conference}, vol.~5B.\hskip 1em plus 0.5em minus 0.4em\relax
  American Society of Mechanical Engineers, 2014.

\bibitem{Ramos22}
Y.~Sim and J.~Ramos, ``Tello leg: The study of design principles and metrics
  for dynamic humanoid robots,'' \emph{IEEE Robotics and Automation Letters},
  vol.~7, no.~4, pp. 9318--9325, 2022.

\bibitem{orin2013centroidal}
D.~E. Orin, A.~Goswami, and S.-H. Lee, ``Centroidal dynamics of a humanoid
  robot,'' \emph{Autonomous robots}, vol.~35, no.~2, pp. 161--176, 2013.

\bibitem{koenig2004design}
N.~Koenig and A.~Howard, ``Design and use paradigms for gazebo, an open-source
  multi-robot simulator,'' in \emph{IEEE/RSJ International Conference on
  Intelligent Robots and Systems}, vol.~3, 2004, pp. 2149--2154.

\bibitem{Laurenzi19}
A.~Laurenzi, E.~M. Hoffman, L.~Muratore, and N.~G. Tsagarakis, ``Cartes{I}/{O}:
  {A} {ROS} {B}ased {R}eal-{T}ime {C}apable {C}artesian {C}ontrol
  {F}ramework,'' in \emph{IEEE International Conference on Robotics and
  Automation}, 2019, pp. 591--596.

\bibitem{felis2017rbdl}
M.~L. Felis, ``Rbdl: an efficient rigid-body dynamics library using recursive
  algorithms,'' \emph{Autonomous Robots}, vol.~41, no.~2, pp. 495--511, 2017.

\bibitem{eigenweb}
G.~Guennebaud, B.~Jacob, \emph{et~al.}, ``Eigen v3,''
  http://eigen.tuxfamily.org, 2010.

\bibitem{Alex12}
Z.~Li, B.~Vanderborght, N.~G. Tsagarakis, L.~Colasanto, and D.~G. Caldwell,
  ``Stabilization for the compliant humanoid robot coman exploiting intrinsic
  and controlled compliance,'' in \emph{IEEE International Conference on
  Robotics and Automation}, 2012, pp. 2000--2006.

\bibitem{kajita2014introduction}
S.~Kajita, H.~Hirukawa, K.~Harada, and K.~Yokoi, \emph{Introduction to humanoid
  robotics}.\hskip 1em plus 0.5em minus 0.4em\relax Springer, 2014, vol. 101.

\bibitem{cic}
A.~Albu-Schaffer, C.~Ott, U.~Frese, and G.~Hirzinger, ``Cartesian impedance
  control of redundant robots: recent results with the dlr-light-weight-arms,''
  in \emph{IEEE International Conference on Robotics and Automation}, vol.~3,
  2003, pp. 3704--3709.

\end{thebibliography}

\begin{appendices}
\section{Computation of Relative Kinematics}\label{app:A}
\begin{figure}[htb!]
    \centering
    \includegraphics[width=1.\columnwidth, trim={5cm 2cm 5cm 2cm}, clip=true]{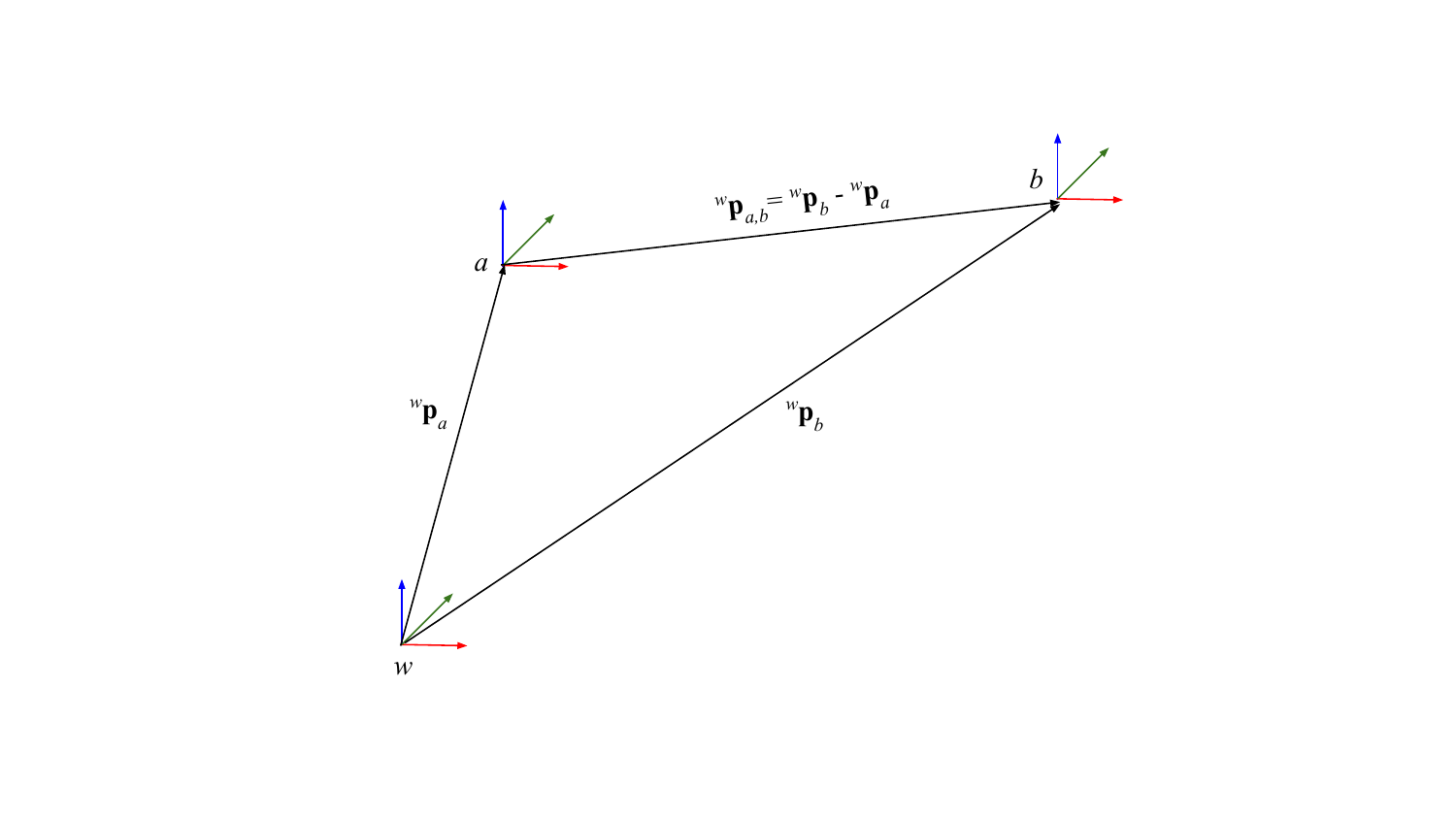}
    \caption{Frames and notation.}
    \label{fig:frames}
\end{figure}

Considering three frames, denoted as $w$, $a$, and $b$ as depicted in Fig.~\ref{fig:frames}, our interest is in calculating relative kinematic values associated with frame $b$ w.r.t. frame $a$, expressed in frame $a$, and using kinematics quantities of $a$ and $b$ expressed in frame $w$.
\par
According to Fig.~\ref{fig:frames}, we denote with ${^w\mathbf{p}_a}$ and ${^w\mathbf{p}_b}$ the position of the frames $a$ and $b$, respectively, w.r.t. the frame $w$. 
The position of frame $a$ w.r.t. frame $b$ expressed in frame $w$ is denoted with ${^w\mathbf{p}_{a,b}} = {^w\mathbf{p}_b} - {^w\mathbf{p}_a}$.
As previously stated, we consider known the following kinematics quantities relative to the frames $a$ and $b$: ${^w\mathbf{v}_{\cdot}}$, ${^w\boldsymbol{\omega}_{\cdot}}$, ${^w\mathbf{J}_{\cdot}}$, ${^w\mathbf{a}_{\cdot}}$, ${^w\boldsymbol{\dot{\omega}}_{\cdot}}$, and ${^w\mathbf{\dot{J}}_{\cdot}}\mathbf{\dot{q}}$, respectively, the linear and angular velocities, Jacobian, linear and angular accelerations, and Cartesian acceleration bias, expressed in frame $w$. 
In the Jacobians, we denote with the superscript ``$\rightarrow$'' the \emph{linear} part while with ``$\angle$'' the \emph{angular} part.

\subsection{Relative Velocity and Jacobian}\label{app:A1}
We first compute the relative linear and angular velocities of the frame $b$ w.r.t. the frame $a$ expressed in frame $a$, denoted respectively with ${^a\mathbf{v}_{a,b}}$ and ${^a\boldsymbol{\omega}_{a,b}}$:
\begin{equation}
    \begin{bmatrix}
{^a\mathbf{v}_{a,b}}\\ 
{^a\boldsymbol{\omega}_{a,b}}
\end{bmatrix}
=
\begin{bmatrix}
{^a\mathbf{R}_w} & \mathbf{0}\\ 
\mathbf{0} & {^a\mathbf{R}_w}
\end{bmatrix}
\begin{bmatrix}
{^w\mathbf{v}_b}\\ 
{^w\boldsymbol{\omega}_b}
\end{bmatrix} -
\mathbf{A}
\begin{bmatrix}
{^w\mathbf{v}_a}\\ 
{^w\boldsymbol{\omega}_a}
\end{bmatrix},
\label{eq:relative_twist}
\end{equation}
with ${^a\mathbf{R}_w}$ the rotation matrix from $a$ to $w$ such that ${^a\mathbf{R}_w}{^w\mathbf{v}_a} = {^a\mathbf{v}_a}$ and ${^w\mathbf{p}_{{a,b}_{[\times]}}}$ the skew-symmetric matrix from $^w\mathbf{p}_{a,b}$ and: 
\begin{equation}
     \mathbf{A} = \begin{bmatrix}
{^a\mathbf{R}_w} & -{^a\mathbf{R}_w}{^w\mathbf{p}_{{a,b}_{[\times]}}}\\ 
\mathbf{0} & {^a\mathbf{R}_w}
\end{bmatrix}.
\end{equation}
\par
We can further expand~\eqref{eq:relative_twist} in its linear and angular parts:
\begin{subequations}
\label{eq:relative_twist_2}
\begin{align}
{^a\mathbf{v}_{a,b}} &= {^a\mathbf{R}_w}{^w\mathbf{v}_b} - {^a\mathbf{R}_w}{^w\mathbf{v}_a} + {^a\mathbf{R}_w}\left({^w\mathbf{p}_{a,b}} \times {^w\boldsymbol{\omega}_a} \right ) = \nonumber\\
&= {^a\mathbf{R}_w}\left({^w\mathbf{v}_b} - {^w\mathbf{v}_a} - {^w\boldsymbol{\omega}_a} \times {^w\mathbf{p}_{a,b}} \right ), \label{eq:relative_twist_2_a}
\\
&\nonumber
\\
{^a\boldsymbol{\omega}_{a,b}} &= {^a\mathbf{R}_w}{^w\boldsymbol{\omega}_b} - {^a\mathbf{R}_w}{^w\boldsymbol{\omega}_a} =  {^a\mathbf{R}_w}\left({^w\boldsymbol{\omega}_b} - {^w\boldsymbol{\omega}_a} \right ) = \nonumber \\
&= {^a\mathbf{R}_w}{^w\boldsymbol{\omega}_{a,b}}. \label{eq:relative_twist_2_b}
\end{align}
\end{subequations}
\par
From equation~\eqref{eq:relative_twist} is finally possible to extract the relative Jacobian ${^a\mathbf{J}_{a,b}}$:
\begin{equation}
    {^a\mathbf{J}_{a,b}}
=
\begin{bmatrix}
{^a\mathbf{R}_w} & \mathbf{0}\\ 
\mathbf{0} & {^a\mathbf{R}_w}
\end{bmatrix}
{^w\mathbf{J}_b}
-
\begin{bmatrix}
{^a\mathbf{R}_w} & -{^a\mathbf{R}_w}{^w\mathbf{p}_{{a,b}_{[\times]}}}\\ 
\mathbf{0} & {^a\mathbf{R}_w}
\end{bmatrix}
{^w\mathbf{J}_a}.
\label{eq:relative_jacobian}
\end{equation}

\subsection{Relative Acceleration and Acceleration Bias}\label{app:A2}
We want now to compute the relative linear and angular accelerations of the frame $b$ w.r.t. the frame $a$ expressed in frame $a$, denoted respectively with ${^a\mathbf{a}_{a,b}}$ and ${^a\boldsymbol{\dot{\omega}}_{a,b}}$.
 \par
Let begin with the relative angular acceleration obtained deriving equation~\eqref{eq:relative_twist_2_b} w.r.t. time:
\begin{equation}
    {^a\boldsymbol{\dot{\omega}}_{a,b}} = {^a\mathbf{\dot{R}}_w}{^w\boldsymbol{\omega}_{a,b}} + {^a\mathbf{R}_w}{^w\boldsymbol{\dot{\omega}}_{a,b}}.
    \label{eq:relative_angular_acceleration}
\end{equation}
Notice that the derivative of the rotation matrix can be written as: 
\begin{align}
{^a\mathbf{\dot{R}}_w} = -^a\mathbf{R}_w{^w\boldsymbol{\omega}_a}_{[\times]}, 
\end{align}
that substituted in~\eqref{eq:relative_angular_acceleration} gives the expression of the relative angular acceleration ${^a\boldsymbol{\dot{\omega}}_{a,b}}$:
\begin{align}
    {^a\boldsymbol{\dot{\omega}}_{a,b}} &= {^a\mathbf{\dot{R}}_w}{^w\boldsymbol{\omega}_{a,b}} + {^a\mathbf{R}_w}{^w\boldsymbol{\dot{\omega}}_{a,b}}=
\nonumber\\
&=-{^a\mathbf{R}_w}{^w\boldsymbol{\omega}_a} \times{^w\boldsymbol{\omega}_{a,b}} + {^a\mathbf{R}_w}{^w\boldsymbol{\dot{\omega}}_{a,b}}=
\nonumber\\
&=  {^a\mathbf{R}_w}\left( {^w\boldsymbol{\dot{\omega}}_{a,b}} - {^w\boldsymbol{\omega}_a} \times{^w\boldsymbol{\omega}_{a,b}} \right ) =
\nonumber\\
&={^a\mathbf{R}_w}\left( {^w\boldsymbol{\dot{\omega}}_{a,b}} - {^w\boldsymbol{\omega}_a} \times{^w\boldsymbol{\omega}_b} + \cancel{{^w\boldsymbol{\omega}_a} \times{^w\boldsymbol{\omega}_a}} \right ) =
\nonumber\\
&=
{^a\mathbf{R}_w}\left( {^w\boldsymbol{\dot{\omega}}_b} - {^w\boldsymbol{\dot{\omega}}_a} - {^w\boldsymbol{\omega}_a} \times{^w\boldsymbol{\omega}_b} \right ).
\label{eq:relative_angular_acceleration_2}
\end{align}     
From the expression~\eqref{eq:relative_angular_acceleration_2} we can compute the angular part of the relative Cartesian acceleration bias ${^a{\mathbf{\dot{J}}}_{a,b}^{\angle}}\mathbf{\dot{q}}$, reported in equation~\eqref{eq:angular_jdotqdot}.
\par
Concerning the relative linear acceleration, its expression is reported in equation~\eqref{eq:relative_linear_acceleration_2}, obtained deriving equation~\eqref{eq:relative_twist_2_a} w.r.t. time and using ${^w\mathbf{a}_{a,b}} = {^w\mathbf{a}_b} - {^w\mathbf{a}_a}$ and ${^w\mathbf{v}_{a,b}} = {^w\mathbf{v}_b} - {^w\mathbf{v}_a}$.
Finally, from expression~\eqref{eq:relative_linear_acceleration_2} we can compute the linear part of the relative Cartesian acceleration bias ${^a{\mathbf{\dot{J}}}_{a,b}^{\rightarrow}}\mathbf{\dot{q}}$, reported in equation~\eqref{eq:linear_jdotqdot} with $\mathbf{b} = -2{^w\boldsymbol{\omega}_a} \times {^w\mathbf{v}_{a,b}} +{^w\boldsymbol{\omega}_a} \times {^w\boldsymbol{\omega}_a} \times {^w\mathbf{p}_{a,b}}$.

\begin{figure*}[!htb]
    \begin{align}
{^a\boldsymbol{\dot{\omega}}_{a,b}} &=
{^a\mathbf{R}_w}\left( {^w\boldsymbol{\dot{\omega}}_b} - {^w\boldsymbol{\dot{\omega}}_a} - {^w\boldsymbol{\omega}_a} \times{^w\boldsymbol{\omega}_b} \right ) = 
\nonumber\\
&=
{^a\mathbf{R}_w}\left( {^w\mathbf{J}_b^{\angle}}\mathbf{\ddot{q}} + {^w\mathbf{\dot{J}}_b^{\angle}}\mathbf{\dot{q}} - {^w\mathbf{J}_a^{\angle}}\mathbf{\ddot{q}} - {^w\mathbf{\dot{J}}_a^{\angle}}\mathbf{\dot{q}} - {^w\boldsymbol{\omega}_a} \times{^w\boldsymbol{\omega}_b} \right ) =
\nonumber\\
&=
{^a\mathbf{R}_w}\left( {^w\mathbf{J}_b^{\angle}} - {^w\mathbf{J}_a^{\angle}}\right )\mathbf{\ddot{q}} + {^a\mathbf{R}_w}\left({^w\mathbf{\dot{J}}_b^{\angle}}\mathbf{\dot{q}} - {^w\mathbf{\dot{J}}_a^{\angle}}\mathbf{\dot{q}} - {^w\boldsymbol{\omega}_a} \times{^w\boldsymbol{\omega}_b}\right ) = 
\nonumber\\
&=
{^a{\mathbf{J}}_{a,b}^{\angle}}\mathbf{\ddot{q}} + {^a{\mathbf{\dot{J}}}_{a,b}^{\angle}}\mathbf{\dot{q}},
\label{eq:angular_jdotqdot}
\\
\nonumber\\    
{^a\mathbf{a}_{a,b}} &= {^a\mathbf{\dot{R}}_w}{^w\mathbf{v}_{a,b}} + {^a\mathbf{R}_w}{^w\mathbf{a}_{a,b}} - {^a\mathbf{\dot{R}}_w}\left({^w\boldsymbol{\omega}_a} \times {^w\mathbf{p}_{a,b}} \right ) - {^a\mathbf{R}_w}\left({^w\boldsymbol{\dot{\omega}}_a} \times {^w\mathbf{p}_{a,b}} \right ) - {^a\mathbf{R}_w}\left({^w\boldsymbol{\omega}_a} \times {^w\mathbf{v}_{a,b}} \right ) =
\nonumber \\ 
& = -{^a\mathbf{R}_w}{^w\boldsymbol{\omega}_a}\times{^w\mathbf{v}_{a,b}} + {^a\mathbf{R}_w}{^w\mathbf{a}_{a,b}} + {^a\mathbf{R}_w}{^w\boldsymbol{\omega}_a}\times\left({^w\boldsymbol{\omega}_a} \times {^w\mathbf{p}_{a,b}} \right ) - {^a\mathbf{R}_w}\left({^w\boldsymbol{\dot{\omega}}_a} \times {^w\mathbf{p}_{a,b}} \right ) - {^a\mathbf{R}_w}\left({^w\boldsymbol{\omega}_a} \times {^w\mathbf{v}_{a,b}} \right ) =
\nonumber \\
& = {^a\mathbf{R}_w}\left[-{^w\boldsymbol{\omega}_a}\times{^w\mathbf{v}_{a,b}} + {^w\mathbf{a}_{a,b}} + {^w\boldsymbol{\omega}_a}\times\left({^w\boldsymbol{\omega}_a} \times {^w\mathbf{p}_{a,b}} \right ) - \left({^w\boldsymbol{\dot{\omega}}_a} \times {^w\mathbf{p}_{a,b}} \right ) - \left({^w\boldsymbol{\omega}_a} \times {^w\mathbf{v}_{a,b}} \right )\right] =
\nonumber \\
& = {^a\mathbf{R}_w}\left[{^w\mathbf{a}_{a,b}}- {^w\boldsymbol{\dot{\omega}}_a} \times {^w\mathbf{p}_{a,b}} -2{^w\boldsymbol{\omega}_a}\times{^w\mathbf{v}_{a,b}}+ {^w\boldsymbol{\omega}_a}\times\left({^w\boldsymbol{\omega}_a} \times {^w\mathbf{p}_{a,b}} \right ) \right],
 \label{eq:relative_linear_acceleration_2}
 \\
 \nonumber \\
{^a\mathbf{a}_{a,b}} &= {^a\mathbf{R}_w} \left({^w\mathbf{a}_b} - {^w\mathbf{a}_a} - {^w\boldsymbol{\dot{\omega}}_a} \times {^w\mathbf{p}_{a,b}} + \mathbf{b}\right ) = 
\nonumber\\
&= {^a\mathbf{R}_w} \left[{^w\mathbf{J}_b^{\rightarrow}}\mathbf{\ddot{q}} + {^w\mathbf{\dot{J}}_b^{\rightarrow}}\mathbf{\dot{q}} - {^w\mathbf{J}_a^{\rightarrow}}\mathbf{\ddot{q}} - {^w\mathbf{\dot{J}}_a^{\rightarrow}}\mathbf{\dot{q}} - \left( {^w\mathbf{J}_a^{\angle}}\mathbf{\ddot{q}} + {^w\mathbf{\dot{J}}_a^{\angle}}\mathbf{\dot{q}} \right) \times {^w\mathbf{p}_{a,b}} + \mathbf{b}\right ] = 
\nonumber\\
&= {^a\mathbf{R}_w}\left({^w\mathbf{J}_b^{\rightarrow}}-{^w\mathbf{J}_a^{\rightarrow}}+{^w\mathbf{p}_{a,b}} \times {^w\mathbf{J}_a^{\angle}} \right)\mathbf{\ddot{q}} 
+ {^a\mathbf{R}_w}\left({^w\mathbf{\dot{J}}_b^{\rightarrow}}\mathbf{\dot{q}}-{^w\mathbf{\dot{J}}_a^{\rightarrow}}\mathbf{\dot{q}} - {^w\mathbf{\dot{J}}_a^{\angle}}\mathbf{\dot{q}}\times {^w\mathbf{p}_{a,b}} + \mathbf{b}\right) = 
\nonumber\\
&= {^a\mathbf{J}_{a,b}^{\rightarrow}}\mathbf{\ddot{q}} + {^a\mathbf{\dot{J}}_{a,b}^{\rightarrow}}\mathbf{\dot{q}}.
\label{eq:linear_jdotqdot}
\end{align}
\end{figure*}

\section{Projected Floating-Base Dynamics}\label{app:B}
This appendix presents the derivation of the floating-base dynamics projected into the closed linkage constraints.
In particular, the final expression will linearly depend only on the actuated accelerations $\boldsymbol{\theta}_a$, base accelerations $\left[ \mathbf{p}^T, \boldsymbol{\rho}^T \right]^T$, and contact forces $\mathbf{F}$.
\par
We first introduce the quantity $\mathbf{q}_a = \left[\mathbf{p}^T, \boldsymbol{\rho}^T, \boldsymbol{\theta}_a^T\right]^T$ and corresponding derivatives computed as for~\eqref{eq:generalized_coordinates}.
%
Furthermore, we consider the quantities $\mathbf{q}$ and $\boldsymbol{\nu}$ as known. 
We want to compute the equations of motions in the actuated accelerations $\boldsymbol{\dot{\nu}}_a$.
With these concerns, we can rewrite the floating-base part of the floating-base dynamics~\eqref{eq:inverse_dynamics_floating_base} as:
\begin{align}
    \mathbf{M}_{b,b}(\mathbf{q})\begin{bmatrix} 
\mathbf{\ddot{p}}\\ 
\boldsymbol{\dot{\omega}}
\end{bmatrix}
 + \mathbf{M}_{b,u}(\mathbf{q})\boldsymbol{\ddot{\theta}}_u + \mathbf{M}_{b,a}(\mathbf{q})\boldsymbol{\ddot{\theta}}_a
 + \mathbf{h}_b(\mathbf{q}, \boldsymbol{\nu}) &= \nonumber \\ 
 \mathbf{J}_{c,b}(\mathbf{q})^T\mathbf{F}
 \end{align}
with $\mathbf{M}_{b,b}(\mathbf{q}) \in \mathbb{R}^{6 \times 6}$, $\mathbf{M}_{b,u}(\mathbf{q}) \in \mathbb{R}^{6 \times m}$ and $\mathbf{M}_{b,a}(\mathbf{q}) \in \mathbb{R}^{6 \times n-m}$.
Substituting equation~\eqref{eq:passive_accelerations_crank_slider} and reorganizing leads to the floating-base part of the projected dynamics:
\begin{equation}
\mathbf{M}_{b,m}(\mathbf{q})\boldsymbol{\dot{\nu}}_a + \mathbf{h}_{b,m}(\mathbf{q}, \boldsymbol{\nu}) = \mathbf{J}_{c,b}(\mathbf{q})^T\mathbf{F},
\label{eq:floating_base_dynamics_actuated}
\end{equation}
with:
\begin{subequations}
\begin{align}
    \mathbf{M}_{b,m}(\mathbf{q}) &= 
    \begin{bmatrix}  
    \mathbf{M}_{b,b}(\mathbf{q}) & \mathbf{M}_{b,u}(\mathbf{q})\mathbf{J}_m(\mathbf{q}) +  \mathbf{M}_{b,a}(\mathbf{q})
    \end{bmatrix},\\
    \mathbf{h}_{b,m}(\mathbf{q}, \boldsymbol{\nu}) &= \mathbf{h}_b(\mathbf{q}, \boldsymbol{\nu}) - \mathbf{M}_{b,u}(\mathbf{q}){{\mathbf{J}_{l,u}}(\mathbf{q})^{-1}}
    {\mathbf{\dot{J}}_l}(\mathbf{q},\boldsymbol{\nu})\boldsymbol{\nu},
\end{align}
\end{subequations}
where $\mathbf{M}_{b,m}(\mathbf{q}) \in \mathbb{R}^{6 \times n-m+6}$ and $\mathbf{h}_{b,m}(\mathbf{q}, \boldsymbol{\nu}) \in \mathbb{R}^{6}$.
\par
Given a generic task in operational space at the acceleration level:
\begin{equation}
    \mathbf{J}_t(\mathbf{q})\boldsymbol{\dot{\nu}} + \mathbf{\dot{J}}_t(\mathbf{q}, \boldsymbol{\nu})\boldsymbol{\nu} = \mathbf{a}_d, 
    \label{eq:generic_jacobian}
\end{equation}
also, this can be expressed only using the actuated acceleration:
\begin{equation}
    \mathbf{J}_{t,m}(\mathbf{q})\boldsymbol{\dot{\nu}}_a + \mathbf{\dot{J}}_{t,m}(\mathbf{q}, \boldsymbol{\nu})\boldsymbol{\nu} = \mathbf{a}_d,
    \label{eq:task_actuated}
\end{equation}
with:
\begin{subequations}
\begin{align}
    \mathbf{J}_{t,m}(\mathbf{q}) &= \begin{bmatrix}
    \mathbf{J}_{t,b}(\mathbf{q}) & \mathbf{J}_{t,a}(\mathbf{q}) + \mathbf{J}_{t,u}(\mathbf{q})\mathbf{J}_m(\mathbf{q})   
    \end{bmatrix}, \label{eq:projected_jacobian}\\
    \mathbf{\dot{J}}_{t,m}(\mathbf{q}, \boldsymbol{\nu})\boldsymbol{\nu} &= \mathbf{\dot{J}}_t(\mathbf{q}, \boldsymbol{\nu})\boldsymbol{\nu} - \mathbf{J}_{t,u}{{\mathbf{J}_{l,u}}(\mathbf{q})^{-1}}{\mathbf{\dot{J}}_l}(\mathbf{q},\boldsymbol{\nu})\boldsymbol{\nu}
\end{align}
\end{subequations}
\par
Equations~\eqref{eq:floating_base_dynamics_actuated} and~\eqref{eq:task_actuated} permits to rewrite the QP problem in~\eqref{eq:whole_body_id_full} only in the actuated and base accelerations, and contact forces.
\par
In the same way, it is possible to compute the Lagrange multipliers $\boldsymbol{\lambda}$ only in the actuated accelerations:
\begin{equation}
    \boldsymbol{\lambda} = \mathbf{J}_{l,u}(\mathbf{q})^{-T}\left( \mathbf{M}_{u,m}(\mathbf{q})\boldsymbol{\dot{\nu}}_a + \mathbf{h}_{u,m}(\mathbf{q}, \boldsymbol{\nu}) -\mathbf{J}_{c,u}(\mathbf{q})^T\mathbf{F} \right),
    \label{eq:whole_body_id_lambda_actuated}
\end{equation}
with:
\begin{subequations}
\begin{align}
    \mathbf{M}_{u,m}(\mathbf{q}) &= \begin{bmatrix}
        \mathbf{M}_{u,b}(\mathbf{q}) & \mathbf{M}_{u,u}(\mathbf{q})\mathbf{J}_m(\mathbf{q}) + \mathbf{M}_{u,a}(\mathbf{q})
    \end{bmatrix},\\
    \mathbf{h}_{u,m}(\mathbf{q}, \boldsymbol{\nu}) &= \mathbf{h}_u(\mathbf{q}, \boldsymbol{\nu}) - \mathbf{M}_{u,u}(\mathbf{q}){{\mathbf{J}_{l,u}}(\mathbf{q})^{-1}}{\mathbf{\dot{J}}_l}(\mathbf{q},\boldsymbol{\nu})\boldsymbol{\nu},
\end{align}
\end{subequations}
where $\mathbf{M}_{u,b}(\mathbf{q}) \in \mathbb{R}^{m \times 6}$, $\mathbf{M}_{u,u}(\mathbf{q}) \in \mathbb{R}^{m \times m}$, $\mathbf{M}_{u,a}(\mathbf{q}) \in \mathbb{R}^{m \times n-m}$,
and finally the actuated torques $\boldsymbol{\tau}$:\begin{equation}
    \boldsymbol{\tau} = \mathbf{M}_{a,m}(\mathbf{q})\boldsymbol{\dot{\nu}}_a + \mathbf{h}_{a,m}(\mathbf{q}, \boldsymbol{\nu}) - \mathbf{J}_{c,a}(\mathbf{q})^T\mathbf{F} - \mathbf{J}_{l,a}(\mathbf{q})^T\boldsymbol{\lambda}. 
    \label{eq:whole_body_id_joint_actuated}
\end{equation}
with:
\begin{subequations}
\begin{align}
    \mathbf{M}_{a,m}(\mathbf{q}) &= \begin{bmatrix}
        \mathbf{M}_{a,b}(\mathbf{q}) & \mathbf{M}_{a,u}(\mathbf{q})\mathbf{J}_m(\mathbf{q}) + \mathbf{M}_{a,a}(\mathbf{q})
    \end{bmatrix},\\
    \mathbf{h}_{a,m}(\mathbf{q}, \boldsymbol{\nu}) &= \mathbf{h}_a(\mathbf{q}, \boldsymbol{\nu}) - \mathbf{M}_{a,u}(\mathbf{q}){{\mathbf{J}_{l,u}}(\mathbf{q})^{-1}}{\mathbf{\dot{J}}_l}(\mathbf{q},\boldsymbol{\nu})\boldsymbol{\nu},
\end{align}
\end{subequations}
where $\mathbf{M}_{a,b}(\mathbf{q}) \in \mathbb{R}^{n-m \times 6}$, $\mathbf{M}_{a,a}(\mathbf{q}) \in \mathbb{R}^{n-m \times n-m}$, $\mathbf{M}_{a,u}(\mathbf{q}) = \mathbf{M}_{u,a}(\mathbf{q})^T \in \mathbb{R}^{n-m \times m}$.
\par
The joint space part of the projected dynamics can be rewritten as:
\begin{equation}
\mathbf{M}_{j,m}(\mathbf{q})\boldsymbol{\dot{\nu}}_a + \mathbf{h}_{j,m}(\mathbf{q}, \boldsymbol{\nu}) = \boldsymbol{\tau} + \mathbf{J}_{c,m,j}(\mathbf{q})^T\mathbf{F},
\label{eq:joint_part_dynammics_actuated}
\end{equation}
with $\mathbf{M}_{j,m}(\mathbf{q}) \in \mathbb{R}^{n-m \times n-m}$:
\begin{subequations}
\begin{align}
    \mathbf{M}_{j,m}(\mathbf{q}) & = 
    \begin{bmatrix}
    \mathbf{M}_{j,m,b}(\mathbf{q}) & \mathbf{M}_{j,m,j}(\mathbf{q})
    \end{bmatrix}, \\
    \mathbf{M}_{j,m,b}(\mathbf{q}) & = \mathbf{M}_{a,b}(\mathbf{q}) + \mathbf{J}_m^T(\mathbf{q})\mathbf{M}_{u,b}(\mathbf{q}), \\
    \mathbf{M}_{j,m,j}(\mathbf{q}) & = \mathbf{M}_{a,a}(\mathbf{q}) + \mathbf{J}_m^T(\mathbf{q})\mathbf{M}_{u,u}(\mathbf{q})\mathbf{J}_m(\mathbf{q}) + \nonumber\\ 
    &\mathbf{M}_{a,u}(\mathbf{q})\mathbf{J}_m(\mathbf{q}) + \mathbf{J}_m(\mathbf{q})^T\mathbf{M}_{u,a}(\mathbf{q}),
\end{align}
\end{subequations}
and $\mathbf{h}_{j,m}(\mathbf{q}, \boldsymbol{\nu}) \in \mathbb{R}^{n-m}$:
\begin{align}
\mathbf{h}_{j,m}(\mathbf{q}, \boldsymbol{\nu}) &= \mathbf{h}_a(\mathbf{q}, \boldsymbol{\nu}) + \mathbf{J}_m(\mathbf{q})^T\mathbf{h}_u(\mathbf{q}, \boldsymbol{\nu})- \nonumber\\ &-\mathbf{M}_{a,u}(\mathbf{q}){{\mathbf{J}_{l,u}}(\mathbf{q})^{-1}}{\mathbf{\dot{J}}_l}(\mathbf{q},\boldsymbol{\nu})\boldsymbol{\nu}- \nonumber\\
&-\mathbf{J}_m(\mathbf{q})^T\mathbf{M}_{u,u}(\mathbf{q}){{\mathbf{J}_{l,u}}(\mathbf{q})^{-1}}{\mathbf{\dot{J}}_l}(\mathbf{q},\boldsymbol{\nu})\boldsymbol{\nu},
\end{align}
and $\mathbf{J}_{c,m,j}(\mathbf{q}) = \mathbf{J}_{c,a}(\mathbf{q}) + \mathbf{J}_{c,u}(\mathbf{q})\mathbf{J}_m(\mathbf{q})$, the joint part of the contact Jacobian projected as in~\eqref{eq:projected_jacobian}.

\end{appendices}

\end{document}